\documentclass[preprint,12pt]{elsarticle}

\usepackage{amssymb}
\usepackage{amsmath}
\usepackage{algorithm}  
\usepackage{algorithmicx} 
\usepackage{algpseudocode}  
\usepackage{subfigure}
\usepackage[section]{placeins}
\usepackage{tikz}
\usetikzlibrary{backgrounds,mindmap}
\usepackage{xcolor}
\usetikzlibrary{calc,positioning,intersections}
\usepackage{pgfplots}
\usepackage{listings}
\usepackage[left= 2cm,right=2cm,top=2.5cm,bottom=2.5cm]{geometry}

\usepackage{graphicx}
\usepackage{caption}
\journal{Swarm and Evolutionary Computation}

\begin{document}

\begin{frontmatter}



\title{Reinforcement learning based parameters adaption method for particle swarm optimization}


\author{Yin Shiyuan}

\affiliation{organization={},%Department and Organization
            addressline={}, 
            city={},
            postcode={}, 
            state={},
            country={China}}

\begin{abstract}
Particle swarm optimization (PSO) is a well-known optimization algorithm that shows good performance in solving different optimization problems. However, PSO usually suffers from slow convergence. 
In this article, a reinforcement learning-based online parameters adaption method(RLAM) is developed to enhance PSO in convergence by designing a network to control the coefficients of PSO.
Moreover, based on RLAM, a new RLPSO is designed.

In order to investigate the performance of RLAM and RLPSO, experiments on 28 CEC 2013 benchmark functions are carried out when comparing with other online adaption method and PSO variants. The reported computational results show that the proposed RLAM is efficient and effictive and that the the proposed RLPSO is more superior compared with several state-of-the-art PSO variants.

\end{abstract}

\begin{graphicalabstract}
graphicalabstract
\end{graphicalabstract}

\begin{highlights}
\item Research highlight 1
\item Research highlight 2
\end{highlights}


\end{frontmatter}


\section{INTRODUCTION}

In recent years, the research community has witnessed an explosion of literature in the area of swarm and evolutionary computation\cite{DELSER2019220}. Hundreds of novel optimization algorithms have been reported along the years and applied successfully in many applications, e.g., system reliability optimization \cite{XU2019100562}, DNA sequence compression \cite{6031913}, system of boundary value problem \cite{arqub2014numerical}, solving mathematical equations \cite{abu2017adaptation,abu2018solutions}, object-level video advertising \cite{zhang2016object}, wireless networks \cite{milner2012nature}.

Particle swarm optimization (PSO), which was first
proposed in 1995 by Kennedy and Eberhart \cite{kennedy1995particle}[9], is one of the most famous swarm-based optimization algorithms.

Due to its simplicity and high performance, a multitude
of enhancements have been presented on PSO during the last few decades, which can be simply categorized into three types: parameter selection, topology, and hybridization with other algorithms\cite{Xu2020}. 

When solving different optimization problems, appropriate parameters need to be configured for PSO and its variants. The performance of PSO heavily depends on the parameters settings, which shows the importance of parameters adaption. However, it is very hard for humans to conduct a laborious preprocessing phase for the detection of promising parameter values and to design the strategy to monitor the running state of PSO and then adjust the running parameters\cite{Liu2019}. So, this papar focuses on parameters adaption.

The current adaptation algorithms are mainly divided into two categories, offline adaptation and online adaptation. 
Offline adaptation is relatively simple to implement, but because all its designs are determined before the optimization algorithm runs, it loses some adaptability to the problem during running.
In contrast, online adaptation will monitor the operation of the optimization algorithm to control the operation of the algorithm.
Common online adaptation methods include control based on history, control based on test experiments, control based on fuzzy logic and control based on reinforcement learning and so on.
In the existing online adaptations, almost all control rules need to be learned during algorithm running. Although fuzzy logic has pre-configured rules, it also needs to be manually configured item by item, which makes the current adaptation algorithm inefficient.
In addition, these adaptation algorithms are usually designed for a certain optimization algorithm and cannot be applied to various optimization algorithms.
But now in the image domain \cite{liu2021swin} and NLP domain \cite{Devlin2019}, the pre-training model is very mature, which greatly improves the performance of subsequent tasks. This inspired us to use reinforcement learning to improve particle swarm performance through pre-training. 

In this paper, we propose a Reinforcement-Learning-based parameter adaption method(RLAM) by embedding Deep deterministic policy gradient(DDPG)\cite{Lillicrap2016} into the process of PSO. In the proposed method, there are two neural network: actor network and action-value network. Actor network is trained to help the particles in PSO choose their best parameters according to their states. Actor-value network is trained to evaluate the performance of the action network and provide gradients for the training of the actor network. The inputs of actor network includes three parts: the percentage of iterations, no-imporving iterations and the diversity of the swarm. All particle will be divided into several groups and each group has their own action generated by the actor network. Typically the action controls w, c1 and c2 in PSO, but it has the ability to control any parameter if needed. Reward function is needed to train the two networks. The design of it is very simple and targeting at encouraging the PSO to get better solution in every iteration. The whole process of training and utilizing the two networks will be described in Sect \ref{DDPG}.

To evaluate the performance of the method proposed in this paper, three sets of experiments are designed in this paper:
1. Six particle swarm algorithm variants were selected, and their performances combined with RLAM were compared with their original performances.
2. Combine RLAM with the original PSO and compare it with other online adaption methods.
3. A new RLPSO will be designed based on this method, and its performance will be compared with five particle swarm variants and several advanced algorithms proposed in recent years.
These experiments verify the effectiveness of RLAM.

\subsection{The main contribution of this paper}

1. The reinforcement learning algorithm DDPG is introduced into self adaption, which greatly improves the parameter adaptation ability.

2. Introduce the concept of pre-training into self adaption, so that the particle swarm algorithm can not only adapt parameters through the current situation, but also adapt parameters through past experience, which improves the intelligence level of the algorithm.

3. Based on the above improvements, a self-adaptive particle swarm optimization algorithm based on reinforcement learning(RLPSO) is proposed, which greatly improves the performance of particle swarm optimization.
\subsection{The structure of the paper}
The rest of this paper is organized as follows.
related work about self-adaption is introduced in Sect. \ref{sec:RELATED WROK}.
the definition of PSO and DDPG is in Sect. \ref{sec:BACKGROUND INFORMATION}.
the implementation of proposed RLAM and RLPSO algorithms is described is in Sect. \ref{sec:PROPOSED ALGORITHM}.
Experimental studies are presented in Sect. \ref{sec:EXPERIMENTS}.
A conclusion summarizing the contributions of this paper is in Sect. \ref{sec:CONCLUTION}.

\section{RELATED WROK}\label{sec:RELATED WROK}

The original particle swarm is designed to imitate the behavior of birds and fish, but in fact the behavior of birds and fish is far more complicated than that of particle swarm algorithm, and many adaptations will be made when the environment changes.
Moreover, the performance of the particle swarm variant algorithm largely depends on its parameter configuration.
The characteristics of natural organisms and the dependence of particle swarms on parameters have prompted more and more researchers to study how to make particle swarms more intelligent, that is, more adaptive, to meet various optimization problems.
The essence of this method can be understood as an optimizer for an optimizer.

\begin{figure}
  \centering
  \includegraphics[width=\textwidth]{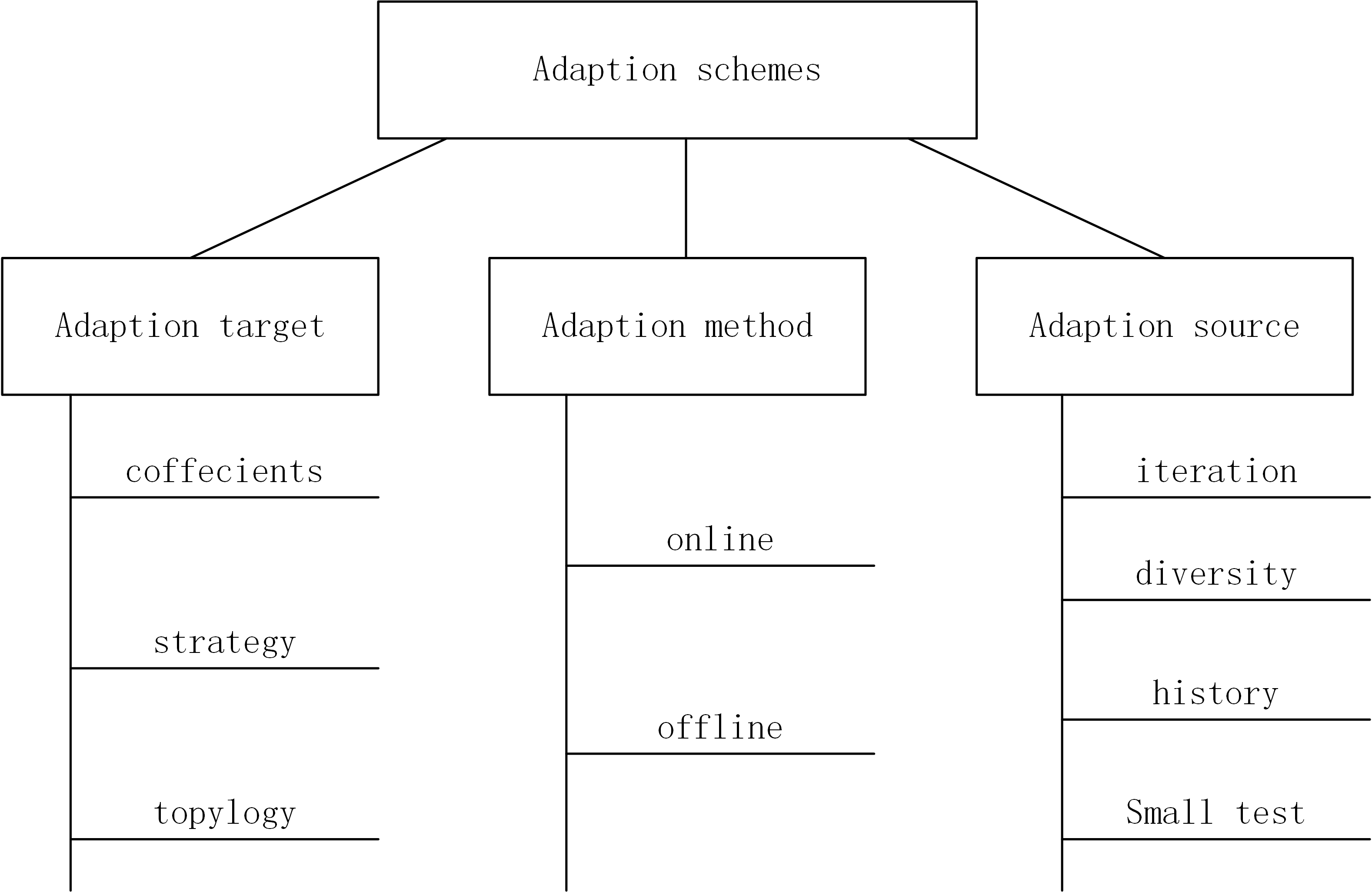}
  \caption{ADAPTION-SCHEMES}
  \label{ADAPTION-SCHEMES}
\end{figure}

Each adaptive algorithm consists of three parts, adaption target, adaption source and adaption method.
The adaptive algorithm receives information from the adaptation source and finally guides the adaptation target to change through the adaptation method.
Its classification is shown in Figure \ref{ADAPTION-SCHEMES}.

This article will focus on the adaptive methods part of the adaptive algorithm.

Adaptive method are distinguished into the two categories, online tuning and offline tuning, depending
on whether they tune the parameters prior or during the algorithm’s execution.

\subsection{offline adaption}

Offline adaption refers to those adaptive methods whose parameters are determined before the actual optimization problem is solved, and can be divided into three categories:
fine tuning, based on the optimization progress of the optimization problem, the parameter linear change and nonlinear change scheme.

\subsubsection{fine tuning}

There are many parameters that need to be configured in the particle swarm algorithm, and these parameters are related to the quality of the final optimization result.
Therefore, the parameters can be tested in groups, the parameter groups can be tested on all test problems, and the parameter group with the best effect can be selected.
Design of Experiments \cite{bartz2007experimental}[2], F-Race \cite{birattari2009tuning}[3], and ParamILS \cite{hoos2011automated}[10] are part of the state-of-the-art in this field.

\subsubsection{Linear strategies}

Initially, all parameters in the particle swarm algorithm were set to a fixed value \cite{kennedy1995particle}, but researchers found that this method was not efficient, and it was difficult to balance the relationship between exploration and discovery.
Therefore, linear strategy is proposed.
Linear strategy means that some algorithms pre-specify a linear expression during the running process, and determine some current running parameters according to some running states.

In the paper \cite{eberhart2000comparing}, the researchers propose a linearly reduced $w$ parameter, which improves the fine tune capability of PSO in the later stage. 
In this paper, the value of $w$ will decrease linearly from an initial value wmax to wmin. The specific formula for this change is as follows:

$$\omega(t)=\frac{t_{\max }-t}{t_{\max }}\left(\omega_{\max }-\omega_{\min }\right)+\omega_{\min }$$
% hpso tvac
Then in the paper \cite{Ratnaweera2004}, the researchers proposed a linear change operation for three parameters, in addition to the linear decrease of w, c1 will increase linearly, and c2 will decrease linearly.
These parameters are preset with their maximum and minimum values. This method makes the PSO algorithm more inclined to search near the optimal value found by itself in the early stage, which improves the particle's exploration ability, and improves its fine tune ability in the later stage, which can quickly converge to the global optimal value . The formula for this change of $c1$ and $c2$ is shown in Eq.\ref{RELATED-WORK-HPSO-TVAC}, and the change of w is consistent with the above formula:

\begin {equation}\label{RELATED-WORK-HPSO-TVAC}
\begin{aligned}
c_{1}=\left(c_{1 f}-c_{1 i}\right) \frac{\text { iter }}{\text { MAXITR }}+c_{1 i} \\
c_{2}=\left(c_{2 f}-c_{2 i}\right) \frac{\text { iter }}{\text { MAXITR }}+c_{2 i}
\end{aligned}
\end {equation}

In the paper \cite{Zheng20031802}, the researchers constructed a linearly growing w, which can also achieve better performance in some of the problems given in the paper. The equation is as follows:

$$\omega(t)=0.5 \times \frac{t}{t_{\max }}+0.4$$

Considering that both the linear growth and decline of w have advantages in some problems, in the paper \cite{Cui200789}, the researchers constructed a linearly increasing w first and then linearly decreasing. The specific changes are as follows:

$$
\omega(t)=\left\{\begin{array}{cc}
1 \times \frac{t}{t_{\max }}+0.4, & 0 \leq \frac{t}{t_{\max }} \leq 0.5 \\
-1 \times \frac{t}{t_{\max }}+1.4, & 0.5<\frac{t}{t_{\max }} \leq 1
\end{array}\right.
$$

In addition to the above-mentioned linear adjustment according to the running progress of the algorithm, there are also papers that linearly adjust the parameters according to some other parameters.

For example, in the paper \cite{Yu20051286}, the researchers adjust the w value according to the average distance between particles. The specific adjustment method is as follows:

$$
\omega(t)=\frac{\omega_{\max }-\omega_{\min }}{D_{\max }-D_{\min }} \times D(t)+\frac{D_{\max } \omega_{\min }-D_{\min } \omega_{\max }}{D_{\max }-D_{\min }}
$$
where $D(t)$ denotes the average distance amongst particles and $\omega_{\max }$ and $\omega_{\min }$ are predefined. 

\subsubsection{Nonlinear strategies}

Inspired by many linear strategies, many researchers have turned their attention to nonlinear strategies. These methods make parameter changes more flexible and closer to the needs of the problem.

In the paper \cite{chen2006natural}, the author proposes to use the E index to update w, which improves the convergence speed of the algorithm on some problems.

In the paper \cite{guimin2006study}, the author uses a quadratic function to update the parameters. Test results show that this method outperforms the linear transformation algorithm in most continuous optimization problems.

In the paper \cite{malik2007new}, the author combines the sigmoid function into the linear transformation, so that the algorithm can quickly converge in the search process.

In the papers \cite{feng2007comparing} and \cite{chen2018chaotic}, the author applies the chaotic model (Logistic map) to the parameter transformation, which makes the algorithm have stronger search ability.

Similarly, in the paper \cite{eberhart2001tracking}, the author sets the parameters to random numbers, and also obtains good results.

\subsection{online adaption}

online tuners are techniques for the dynamic adaptation of the algorithm’s parameters
during its execution. They are typically based on performance feedback from the algorithm on the considered problem instance.

\subsubsection{History based}

Some papers divide the whole running process into many small processes during the running process. Use different parameters or strategies in each small process, and judge whether the parameters or strategies are good enough according to the performance of the algorithm at the end of the small stage. The strategies or parameters that are good enough will be chosen more in subsequent runs. 
In the paper \cite{tanabe2013success}, the author builds a parameter memory.
In each run, all running particles are assigned different parameter groups, and the parameters used by some particles with better performance after the small-stage run is completed will be saved in the parameter memory.
The parameters chosen by subsequent particles will tend to be close to the average of the parameter memory.
In the paper \cite{Lynn2017}, the author designs 5 particle swarm operation strategies based on some excellent particle swarm variant algorithms in the past, and keeps a record of success rate for each strategy.
In the initial state, all success rates are set to 50\%, and then in each small process, a strategy will be randomly selected for execution based on the weight of the past success rate. How many particles have been promoted, based on which the memory of the success rate is updated.
In the paper \cite{liu2020multipopulation}, the author improved the original EPSO, updated the designed strategy, divided the particle swarm into multiple subgroups, and evaluated the strategy separately, which further improved the performance of the algorithm.

\subsubsection{Small Test period based}

This method divides the entire optimization process into multiple sub-processes, and divides each sub-process into two parts, one part is used to test the performance of parameters or strategies, and the number of evaluations is generally less than 10\%, and the other part is the normal optimization process.
In the paper \cite{tatsis2017grid}, the author divides the parameters into many parameter groups according to the grid.
In each performance test process, the population is divided into multiple subgroups, and the parameter groups around the grid of the previous process are tested. After running several iterations, the parameter group with the best effect is selected as the parameter group to be used.
In the paper cite{tatsis2019dynamic}, the author divides the test sub-process into two processes. In the first process, the adjacent points of the current selection point will be evaluated, and the probability of success will be calculated,
Then in the second process, take the direction with the highest probability of success as the exploration direction, explore multiple steps in this direction, and finally select the best parameter group in the second step as the parameter configuration for the next operation.This method further improves performance.

\subsubsection{Fuzzy rules}

In the paper \cite{Olivas2016}, the author presented a new method for dynamic
parameter adaptation in PSO, where the authors proposed
an improvement to the convergence and diversity of the
swarm in PSO using interval type-2 fuzzy logic. The experimental results
were compared with the original PSO, achieving that the
proposed approach improves the performance of PSO.
In the paper \cite{MELIN20133196}, the author  presented a work to improve the
convergence and diversity of the swarm in PSO using type-1 fuzzy logic applied to classification problems.

\subsubsection{Reinforcement learning based}

At present, most optimization algorithms based on reinforcement learning are based on Q-learning for adaptive control parameters and strategies.
In the paper \cite{xu2020reinforcement}, The author combines a variety of topological structures, using the diversity of the particle swarm and the topological structure of the previous step as the state, and selects the topological structure with the largest Q value from the Qtable as the topology to be used in the next optimization step.
In the paper \cite{liu2019adaptive}, the author uses the Qlearning algorithm. Enter the state with how far from the optimal value and the ranking of the particle evaluation value among all particles. Use different parameter groups as adaptation targets. The reward is determined based on whether the entire optimization problem grows, the current optimization progress, and the currently selected action.
In the paper \cite{samma2016new}, the author uses the Qlearning algorithm. 
The strategy selected in the previous step is the state input. 
Output actions control different strategies such as jump or finetune. 
Finally, the reward is determined according to whether the whole optimization problem grows.
In the paper \cite{lu2021reinforcement}, the author uses the Qlearning algorithm,
take the particle position as the state input. 
The output action controls the prediction speed of different particle strategies. Finally, the reward is determined according to the increase or decrease of the particle evaluation value.
In the paper \cite{hsieh2016q}, the author uses the Qlearning algorithm, using the serial number of each particle, the particle position as the state input, and the output action to control two different strategies, and finally determine the reward according to whether the entire optimization problem grows.
In the paper \cite{wu2022employing}, the author adopts the algorithm of policy gradient, which takes the particle position pbest as the state input, and the output action controls C1 and C2, and finally determines the reward according to the growth rate of the entire optimization problem.

A summary of these papers is placed in the table \ref{RELATED-WORK-RL-PSO}, which describes the method used in the paper, state input, action output, reward input and whether to control each particle separately.

\begin{table}[]
\caption{reinforcement learning based self-adaption methods.}\label{RELATED-WORK-RL-PSO}
\resizebox{\textwidth}{!}{
\begin{tabular}{llllll}
paper    & input(adaption source)                 & output(adaption target)             & reward                  & single particle &algorithm \\
\cite{xu2020reinforcement} & topological structure              & next topological structure & diversity, gbest           & N   & Q  \\
\cite{liu2019adaptive} & distance with the best/rank & c1, c2, w                & gbest & Y   & Q  \\
\cite{samma2016new} & pervious strategy              & strategy & gbest                    & Y   & Q  \\
\cite{lu2021reinforcement} & particle postion                  & velocity           & evaluate fitness                  & Y   & Q  \\
\cite{hsieh2016q} & particle index/postion          & strategy               & gbest                    & Y   & Q  \\
\cite{wu2022employing} & postion, pbest            & c1, c2            & gbest                    & Y   & PG
\end{tabular}
}
\end{table}

%\subsection{summary}

\section{BACKGROUND INFORMATION}\label{sec:BACKGROUND INFORMATION}

\subsection{particle swarm optimization}

Particle swarm optimization is the representation of swarm behaviors in some ecological system, such as birds flying and bees foraging [A reinforcement learning-based communication topology in particle swarm optimization:29]\cite{WANG2018162}.In classic PSO, the movement of a particle in PSO is influenced by its own previous best postion and the best postion of the best particle in the swarm. To describe the state of particles, velocity V and postion X is defined as follows:

$$V_{i}=\left(V_{i}^{1}, V_{i}^{2}, \ldots, V_{i}^{D}\right), i=1,2, \ldots, N$$\label{PSO_V_DEFINE}

$$X_{i}=\left(X_{i}^{1}, X_{i}^{2}, \ldots, X_{i}^{D}\right), i=1,2, \ldots, N$$\label{PSO_X_DEFINE}

Here, $D$ represents the dimension of the search space. $N$ represents the number of particles.
As the search progresses, the two movement vectors are updated as follows:

$$
v_{i}(t+1)=w v_{i}(t)+c_{1} r_{1}(t)\left(g{ Best }_{i}(t)-x_{i}(t)\right)+c_{2} r_{2}(t)\left(p{ Best }(t)-x_{i}(t)\right)$$

$$
x_{i}(t+1)=x_{i}(t)+v_{i}(t+1)
$$

Here, $w$ is the inertia weight. $c_{1}$ is the cognitive acceleration coefficient. $c_{2}$ is the social acceleration coefficient. $r_1$ and $r_2$ are uniformly distributed random numbers within [0,1], and $v_{i}(t+1)$ denotes the velocity of the $i^{th}$ particle in the $t^{th}$ generation.
$p{ Best }_{i}$ is the personal best position for the $i^{th}$ particle, and $g{ Best }$ is the best position in this swarms.

\subsection{Reinforcement learning and deep deterministic policy gradient}

this subsection will introduce DDPG and RL briefly. For more detail you can see \cite{Lillicrap2016}.

\subsubsection{Reinforcement learning}

Reinforcement learning(RL) is a kind of machine learning. Its purpose is to guide the agent to perform optimal actions in the environment to maximize the cumulative reward.
The agent will interact with the environment in discrete timesteps.
At each timestep $t$, the agent receives an observation $s_t \in S$, takes an action $a_t \in A$ and receives a scalar reward $r_t(s_t, a_t)$. Here $S$ is the state space, $A$ is the action space. 
The agent`s behavior is controlled by a policy $\pi:S \rightarrow A$ , which maps each observation to an action. 

\subsubsection{Deep deterministic policy gradient}\label{DDPG}

In DDPG, there are 4 neural network designed to get the best policy $\pi$: actor network $\mu(s_t|\theta^{\mu})$, target actor network $\mu^{\prime}(s_t|\theta^{\mu^{\prime}})$, action-value network $Q(s_t,a_t|\theta^{Q})$ and target action-value network $Q^{\prime}(s_t,a_t|\theta^{Q^{\prime}})$. $\mu$ and $\mu^{\prime}$ are used to choose action according to state, $Q$ and $Q^{\prime}$ are used to evaluate the action choosed by the actor network. $\theta^{\mu}$, $\theta^{\mu^{\prime}}$, $\theta^{Q}$ and $\theta^{Q^{\prime}}$ are the neural network weights of above neural networks. At the beginning, $\theta^{\mu^{\prime}}$ is a copy of $\theta^{\mu}$ and $\theta^{Q^{\prime}}$ is a copy of $\theta^{Q}$. During training, the weights of these target networks are updated as follows:

\begin {equation} 
\label{TARGET-UPDATE}
\begin{aligned} 
\theta^{Q^{\prime}} \leftarrow  \tau \theta^{Q} + (1-\tau)\theta^{Q^{\prime}} \\
\theta^{\mu^{\prime}} \leftarrow  \tau \theta^{\mu} + (1-\tau)\theta^{\mu^{\prime}}
\end{aligned}
\end{equation}

Here, $\tau \ll 1$.

To train the actor-value network, we need to minimize the loss function:

\begin {equation} 
\label{THETA-Q-UPDATE}
L\left(\theta^{Q}\right)=\left(r\left(s_{t}, a_{t}\right)+\gamma Q^{\prime}\left(s_{t+1}, a_{t+1} \mid \theta^{Q^{\prime}}\right)-Q\left(s_{t}, a_{t} \mid \theta^{Q}\right)\right)^{2} 
\end{equation}

Then, actor-value network is used to train actor network with policy gradient:

\begin {equation} 
\label{THETA-MU-UPDATE}
\begin{aligned}
\nabla_{\theta^{\mu}} J =\left.\left.\nabla_{a} Q\left(s, a \mid \theta^{Q}\right)\right|_{s=s_{t}, a=\mu\left(s_{t}\right)} \nabla_{\theta_{\mu}} \mu\left(s \mid \theta^{\mu}\right)\right|_{s=s_{t}}
\end{aligned}
\end{equation}

\begin{figure}
  \centering
  \includegraphics[width=\textwidth]{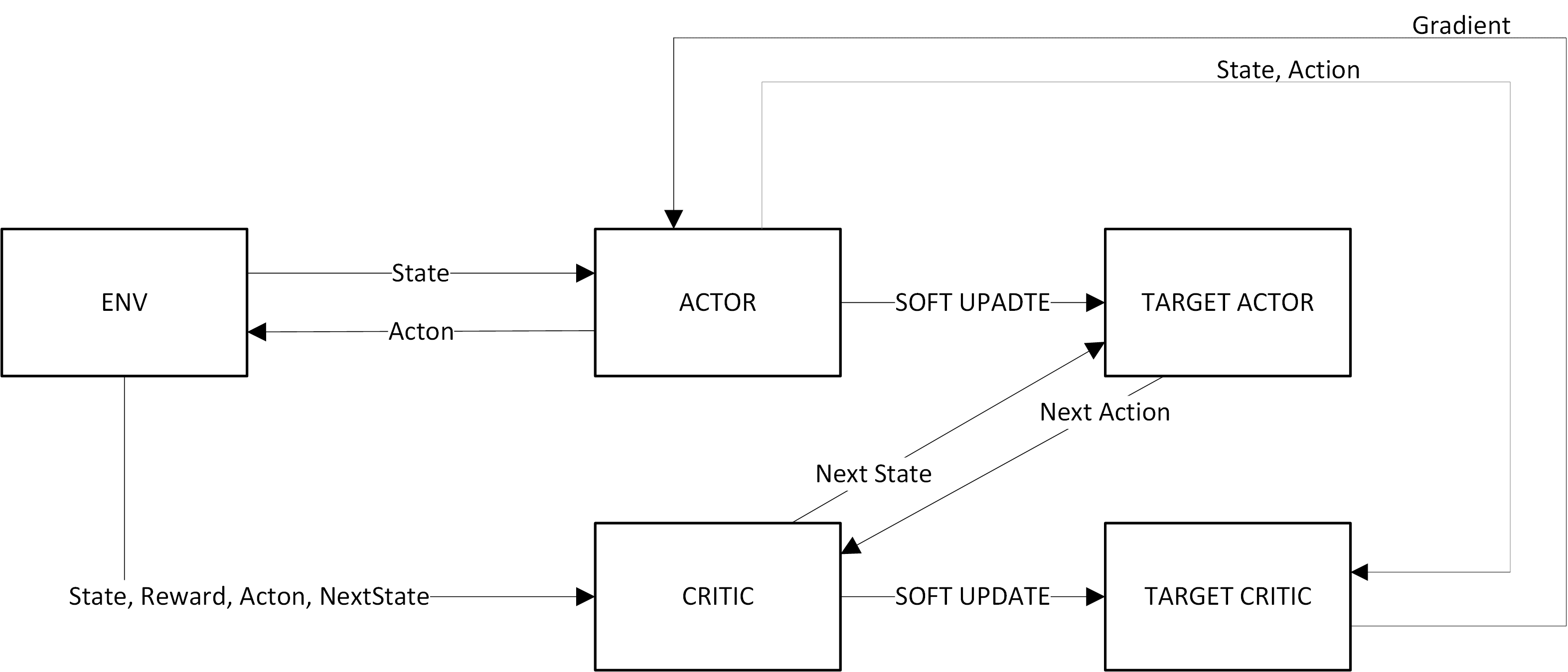}
  \caption{The training process of DDPG}
  \label{The training process of DDPG}
\end{figure}

The data flow of training process is shown in Fig.\ref{The training process of DDPG}.

\subsection{Comprehensive learning particle swarm optimizer (CLPSO)}

For more detail, you can see \cite{Liang2006}.

The velocity updating equation used in CLPSO is as follows:

\begin {equation}\label{CLPSO-V-UPDATE}
V_i^d = wV_i^d + c * rand_i^d * (pbest_{fi(d)}^d - X_i^d) 
\end{equation}

where, $ f_i(d)=[f_i(1),f_i(2),....,f_i(D)] $ defines which particle`s pbest the $i_{th}$ particle should follow. The $d_{th}$ dimension of the $i_{the}$ particle follow the ${f_i(d)}_{th}$ particle`s pbest in $d_{th}$ dimension.

To determine the  $ f_i(d)$, every particle have their own learning $Pc_i$. The $Pc_i$ value for each particle is generated by the following equation:
\begin {equation} 
Pc_i = a + b * \frac{(exp(\frac{10(i-1)}{ps-1})-1)}{exp(10)-1}  
\end{equation}

Here, $ps$ is the population size, $a = 0.05$, $b = 0.45$. When a particle updates its velocity for one dimension, there will be a random value in [0,1] generated and compared with $Pc_i$. If the random value is larger than $Pc_i$, the particle of this dimension will follow its own pbest. Otherwise, it will follow another particle`s pbest for that dimension. CLPSO will employ a tournament selection to choose a target particle. What`s more, to avoid wasting function evaluations in the wrong direction, CLPSO defines a certain number of evaluations as refreshing gap $m$.
During the period of a paticle following a target particle, the number of times the particle ceases improving is recorded as $flag_{clpso}$ 
If $flag_{clpso}$ is bigger than $m$,the particle will get his new target particle by employing a tournament selection again.

%\begin {equation}(V_{CLPSO})_i^d = pbest_{fi(d)}^d - X_i^d\label{ClpsoDeltaV}\end{equation}

\section{PROPOSED ALGORITHM}\label{sec:PROPOSED ALGORITHM}

In this section, an efficient parameters adaption method is proposed. A variant of particle swarm algorithm based on the above algorithm will be introduced later.

\subsection{Parameters self-adaption based on DDPG}

In this section, we will first introduce the input and output of the action network applied to the particle swarm algorithm.
Later, We will then describe how the reward function scores based on changes in state.
The third step will introduce how to train the particle swarm algorithm. How to run it with the trained network will be described later.
Finally, the proposed new particle swarm algorithm is introduced.

\subsubsection{State of PSO (input of actor network)}

The running state designed in this paper is divided into three parts: the current iteration progress, the current particle diversity, and the current duration of the particle no longer growing.
In order for the neural network to work optimally, all operating states will eventually be mapped to the interval -1 to 1, and finally input to the neural network.

The iteration input is considered as a percentage of iterations during the execution of the PSO, 
the values for this input are in the range from 0 to 1. In the start of the algorithm, the iteration is 
considered as 0\% and is increased until it reaches 100\% at the end of the execution. The values are 
obtained from Equation \ref{STATE-ITERATION}.

\begin {equation}\label{STATE-ITERATION}
Iteration = Fe\_num / Fe\_max
\end{equation}

Here, $Fe\_num$ is the number of function evaluations that have been performed, and $Fe\_max$ is the maximum number of function evaluation executions set in the algorithm run.

The diversity input is defined by Equation \ref{STATE-DIVERSITY}, in which the degree of the dispersion of the
swarm is calculated. This means that for less diversity, the particles are closer together, and for high
diversity, the particles are more separated. The diversity equation can be taken as the average of the
Euclidean distances amongst each particle and the best particle:

\begin {equation}\label{STATE-DIVERSITY}
\operatorname{Diversity}(t)=\frac{1}{n_{s}} \sum_{i=1}^{n_{\mathrm{s}}} \sqrt{\sum_{j=1}^{n_{x}}\left[x_{i j}(t)-\overline{x_{j}}(t)\right]}
\end{equation}

Here, $x_{ij}(t)$ is the $j^{th}$ postion of $i^{th}$ particle in the iteration $t$. $\overline{x_{j}}(t)$ is the average $j^{th}$ of all the particles in the iteration $t$. 

The stagnant growth duration input is used to indicate whether the current particle swarm is running efficiently. It is defined as follows:

\begin{equation}\label{STATE-NO-IMPROVEMENT}
Iteration_{no-improvement} = (Fe\_num-Fe\_num_{last-imporve}) / Fe\_max
\end{equation}

Here, $Fe\_num_{last-imporve}$ represents the number of evaluations at the last global optimal update.

In order to map all parameters to 0-1 and make the information more salient, all the above information will be encoded based on the sin-encode method in the literature \cite{vaswani2017attention}.
Assuming that an input is originally x, x can represent $Iteration$, $diversity$ or $Iteration_{no-improvement}$, the output is as follows:

\begin{equation}\label{STATE-SIN-ENCODE}
state_i = sin(x*2^{i})
\end{equation}

Where $state_i$ is the $i^{th}$ parameter newly generated from $x$, in this paper i takes 0, 1, 2, 3, 4.
An x will eventually generate 5 new parameters.
The reason for this operation is that some of the above parameters are very small, and their changes are even smaller, which will cause the action network to fail to capture their changes and fail to perform effective action output.

\subsubsection{Action of PSO (output of actor network)}
\label{ACTION-GENERATE-SECTION}
The actions designed in this paper are used to control the operating parameters in the PSO, such as $w$, $c1$, $c2$. The parameters to be controlled can be set as required. 
The action output by the action network does not control the above parameters of each particle separately, but divides the particles into 5 subgroups, and generates 5 sets of parameters to control different groups.

For the traditional PSO algorithm.
The obtained action vector $a_t$ is 20-dimensional, divided into 5 groups, each of which is aimed at a sub-swarm. For a sub-swarm, the action vector is 4-dimensional: a[0] to a[3]. The $w$, $c1$ and $c2$ required for each round of the optimization algorithm will be generated according to a[0] to a[3]. The generating formula is as follows:

\begin {equation}\label{ACTION-PARAMETTERS}
\begin{aligned}
w = a[0] * 0.8 + 0.1	\\
scale = 1 / (a[1] +a[2] + 0.00001) * a[3] * 8	\\
c1 = scale * a[1]	\\
c2 = scale * a[2]	\\
\end{aligned}
\end{equation}

Here, $scale$ is a parameter helping us normalize $c1$ and $c2$ and it is optional. 

For some PSO variants. Since the parameters have been studied in some algorithms, the original parameter set is already quite good. In order to take advantage of the performance of the original parameter set, the new w, c1, c2 will be configured according to the following formulas.

\begin {equation}\label{ACTION-PARAMETTERS2}
\begin{aligned}
w = a[0] * 0.5 + w_{origin}	\\
c1 = a[1] * 0.5 + c1_{origin}	\\
c2 = a[2] * 0.5 + c2_{origin}	\\
\end{aligned}
\end{equation}

Where $w_{origin}$, $c1_{origin}$, $c2_{origin}$ represent the original parameters of the algorithm. In all subsequent experiments, each algorithm will choose one of the action parameter configuration methods for experimentation.

If there are more parameters in the algorithm that need to be configured, you can increase the number of output parameters and configure them like $c1$ or $w$.

\subsubsection{Reward function}

Reward function is used to caculate reward after an action is excuting. Its target is to encourage the PSO get better Gbest. Therefore, the reward function is designed as follows:

\begin {equation}
r(t) = \begin{cases}
1,\quad & Gbest(t+1)<Gbest(t)\\
-1 ,\quad & Gbest(t+1)=Gbest(t)
\end{cases}
\end{equation}

Here, Gbest(t) is the best solution in the $t^{th}$ iteration.

\subsubsection{Training}
\label{TRAIN-SECTION}

This section will introduce the training process for the traditional PSO algorithm. Other particle swarm variants are basically the same as this process.

In the training process, each iteration of the particle swarm algorithm is equivalent to the agent performing an action in the environment, that is, an epoch of training. The particle swarm completes the optimization of an objective function, which is equivalent to the interaction between the agent and the environment, that is, an episode of training.

The training process is as follows:
First, the action network and the action value network are randomly initialized, and the target action network and the target action value network are copied respectively.
A replay buffer R is then initialized to save the running state, actions and rewards.
The third step initializes the environment, including the initialization of the PSO and the initialization of the objective function to be optimized.
The fourth step is to obtain the running state parameters from the particle swarm.
The fifth step is to input the running state $s_t$ into the action network, get the action $a_t$, and add a certain random noise. The random noise is normally distributed, with 0 as the mean and 0.5 as the variance. The formula is as follows:
\begin{equation}
\label{TRAIN-NOISE}
a(t)=Actor(S_t)+\mathcal N(0,0.5)
\end{equation}
The sixth step converts the resulting actions into the required $w$, $c1$ and $c2$ using the formula \ref{ACTION-PARAMETTERS}.
The seventh step is to perform an iteration of particle swarm optimization according to the above parameters to obtain a new reward $r_{t+1}$ and a new state $s_{t+1}$.
The eighth step saves the state, actions and rewards to cache R.
The ninth step is to randomly select a batch of experiences from cache R. Update the weights of the action-value network by minimizing the loss function (formula \ref{THETA-MU-UPDATE}). Update the weights of the action-action network by (formula \ref{THETA-Q-UPDATE}).
The tenth step is to update the weights of the target action value network and the target network according to the formula \cite{THETA-MU-UPDATE}.
If the particle swarm has not finished iterating yet, then $t$ is increased by 1 and returns to step 4.
If the training is not completed then go back to step 3.

The pseudocode is proposed in Algorithm \ref{train algorithm}.

If the objective function used in training directly adopts the function to be optimized later, the effect will be better. If you train with a set of test functions, it can work on all objective functions, but not as good as the first method.

After training, a trained action network model will be obtained for subsequent runs.

\begin{algorithm}[htb]
  \caption{train algorithm.	}
  \label{train algorithm}
  \begin{algorithmic}[1]
\State Randomly initialize $\theta^Q $ and $\theta^{\mu}$ in action network $\mu(s|\theta^{\mu})$ and action value network $ Q(s,a|\theta^Q) $.
\State Initialize the target network $Q^{\prime}$ and $\mu^{\prime}$, and its weight value is copied from $Q$ and $\mu$ 
\State Initialize the playback buffer R
\For{$episode=1:EpisodeMax$}

\State Initialization environment (PSO and evaluate function)
\For{t=1, Tmax}
	\State get observation $s_t$ from environment 
	\State Choose actions based on $s_t$, network $\mu$ and explore noise\{Eq.\ref{TRAIN-NOISE}\}
	\State Perform the action $a_t$ in the environment and observe the reward $r_t$ and the new state $s_{t+1}$
	\State Save ($s_t$, $a_t$, $r_t$, $s_{t+1}$) to the cache $R$
	\State Update Action Value Network by minimizing the loss function\{Eq.\ref{THETA-Q-UPDATE}\}
	\State Update the action network through the sampled action policy gradient:\{Eq.\ref{THETA-MU-UPDATE}\}
	\State Update the weights of the target network function\{Eq. \ref{TARGET-UPDATE}\}
\EndFor

\EndFor

\end{algorithmic}
\end{algorithm}

\subsubsection{Running}

In the running process, compared to the training process, many steps will be removed, and the overall process is very simple.

The operation process of the traditional particle swarm algorithm will be introduced below. The implementation process of other particle swarm variant algorithms is basically the same as this process.

\begin{figure}
  \centering
  \includegraphics[height=.9\textheight]{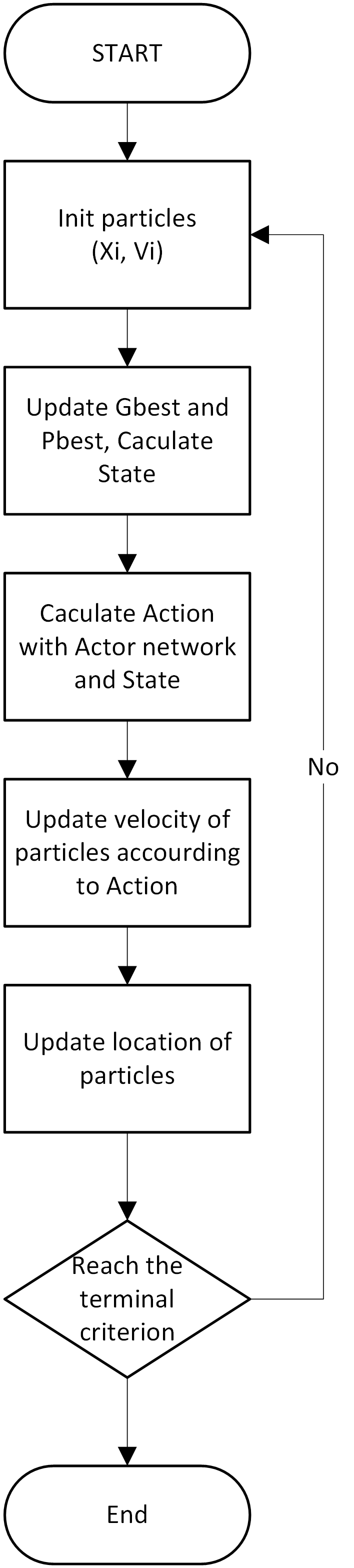}
  \caption{The running process of PSO with RLAM}
  \label{The running process of PSO with RLAM}
\end{figure}

The flow chart of the PSO algorithm combined with RLAM is shown in Figure \ref{The running process of PSO with RLAM}.

n addition to the original process of PSO, the new content is that before the particle velocity is updated, the running state of the particle swarm will be calculated and a new parameter group will be generated to guide the particle update.

\subsubsection{Network structure}

This chapter will introduce the network structure of action network and action value network in detail.

Since the action evaluation network is only used in the pre-training process, and the action network needs to be used in the actual operation, in order to prevent excessive computation during the optimization process, a smaller action network and a larger action evaluation network will be designed.

\begin{figure}
  \centering
  \includegraphics[width=.8\textwidth]{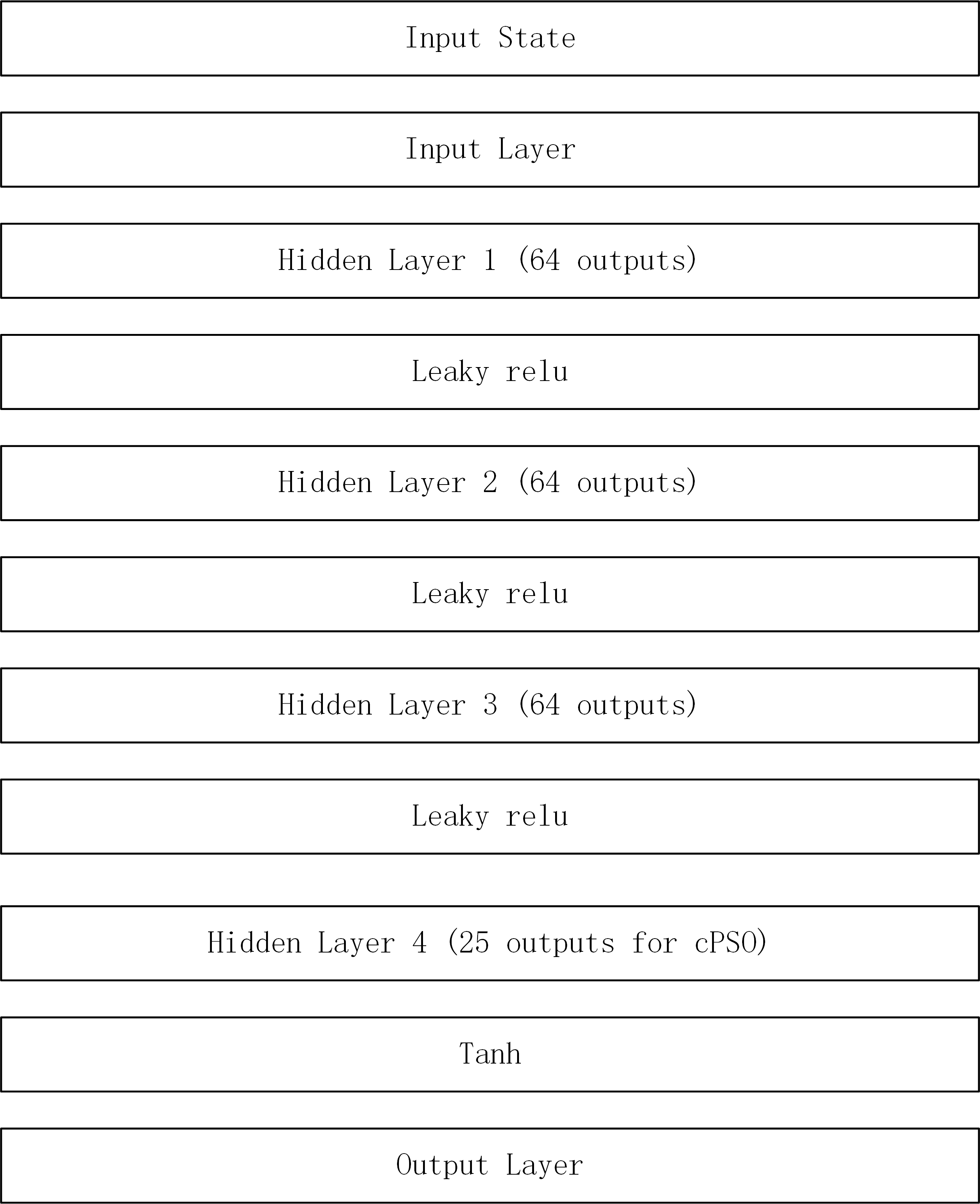}
  \caption{Network structure of actor}
  \label{Network structure of actor}
\end{figure}

\begin{figure}
  \centering
  \includegraphics[width=.8\textwidth]{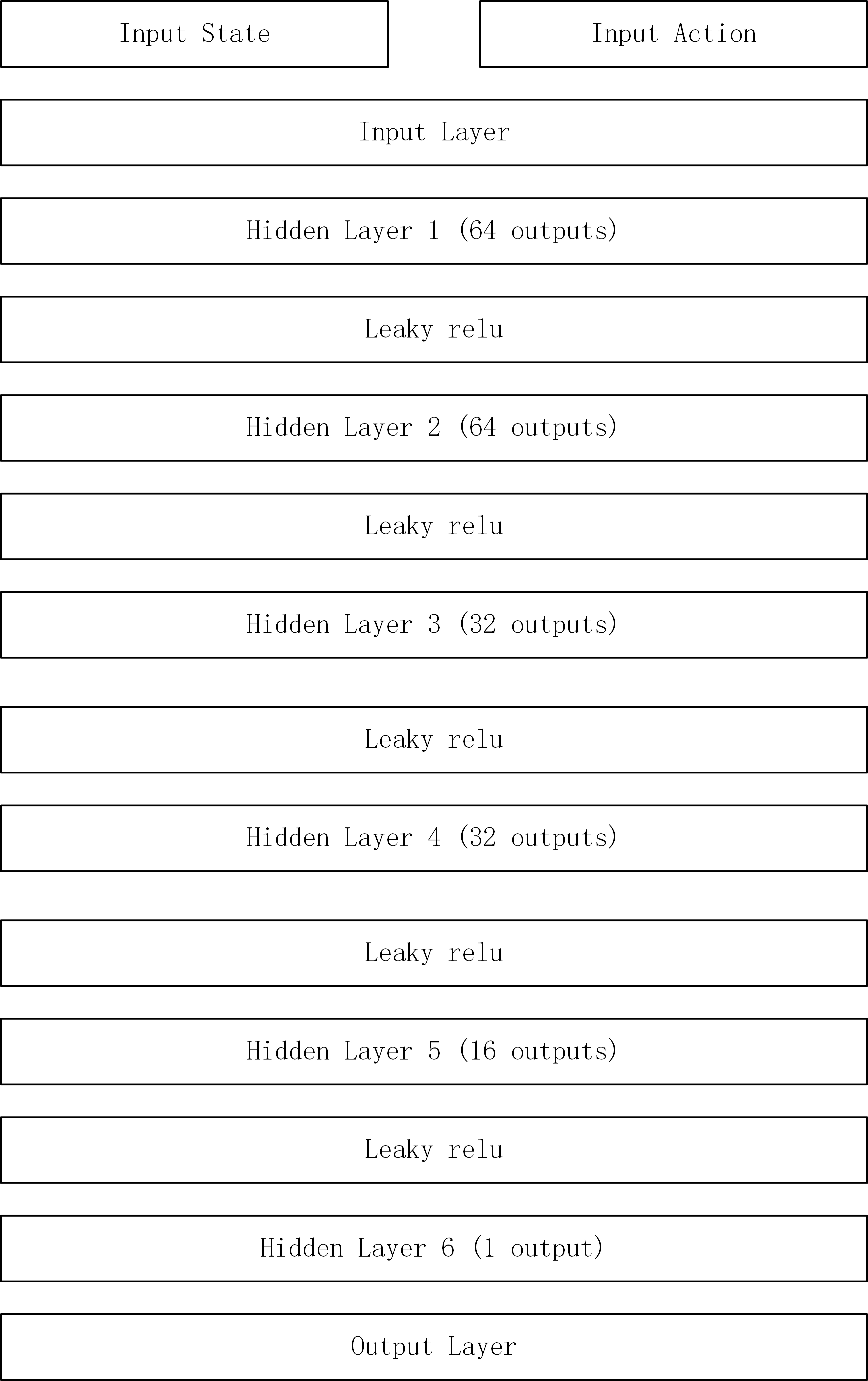}
  \caption{Network structure of critic}
  \label{Network structure of critic}
\end{figure}

The schematic diagrams of the two networks are shown as \ref{Network structure of actor},\ref{Network structure of critic}.

Actor value network adopts the design of a 6-layer fully connected network, and the activation function between the networks is leakyrelu. The action network adopts the design of a 4-layer fully connected network, and the activation function between the networks is also leakyrelu. In the last layer, in order to map the action to the required range of -1 to 1, a layer of tanh activation function is added.

\subsection{RLPSO}

In order to better reflect the parameter adjustment ability of RLAM, this paper designs a new RLPSO algorithm based on RLAM. The speed update equation of this algorithm is as follows:

\begin {equation}
\begin{split}
{V(t+1)}_i^d =& w * {V(t)}_i^d+ c1 * r1 * (pbest_{fi(d)}^d - X_i^d) + \\
&c2 * r2 * (gbest_i^d - X_i^d) + c3 * r3 * (pbest_i^d - X_i^d)
\label{RlepsoVUpdate}\end{split}\end{equation}

Here, $fi(d)$ is introduced in the previous section. pbest is the particle`s own best experience, and gbest is the best experience in this swarm. According to the current running state, $w$, $c1$, $c2$, $c3$, and $c4$ are coefficients generated by the actor network, which has been introduced in Sect \ref{ACTION-GENERATE-SECTION}. $r1$, $r2$ and $r3$ are all uniformly distributed random numbers between 0 and 1. 

To prevent particles from being trapped in the local optimum, there is a mutation stage after the velocity updating. During this stage, first, a random number $r4$ between 0-1 will be generated, and then $r4$ will be compared with $c4 * 0.01 * flag_{clpso}$. If $r4$ is less than it, the mutation will be performed, and the particle position will be reinitialized in solution space.

At the end of one period, particles will move according to their velocity, and then particles` fitness and history best experience will be updated.

The pseudocode is proposed in Algorithm \ref{RLPSO-ALGORITHM}.

\begin{algorithm}[htb]
  \caption{RLPSO algorithm.}
  \label{RLPSO-ALGORITHM}
  \begin{algorithmic}[1]

\State Initialize the particle swarm and parameters
\While{$fe \leq femax$}
	\State Caculate $s_t$  \{Eq. \ref{STATE-ITERATION}, \ref{STATE-DIVERSITY}, \ref{STATE-NO-IMPROVEMENT}, \ref{STATE-SIN-ENCODE}\}
	\State Caculate $a_t$ with actor network $\mu$(generated in Sect. \ref{TRAIN-SECTION}) and $s_t$
\For{k = 1:n}
	\State Convert actions into operating parameters (c1, c2, c3 and c4) \{Eq. \ref{ACTION-PARAMETTERS}\}
	\State Calculate the new speed  \{Eq.\ref{RlepsoVUpdate}\}
	\If{$randomvalue < c4 * 0.01 * flag_{clpso}$}
		\State Reinitialize position
	\Else
		\State Update position $ x_{t+1} = x_t + v_t $
	\EndIf
	\State Calculate the evaluation value of all particles
 	\State Update the parameters in particle operation
	
\EndFor

\EndWhile
\end{algorithmic}
\end{algorithm}

\section{EXPERIMENTS}\label{sec:EXPERIMENTS}

To verify the performance of the proposed algorithm, 3 sets of experiments will be conducted.
In the first experiment, RLAM was fused with various PSO variants, and the obtained results were compared with the original PSO variants.
The second experiment will compare the performance of RLAM with other online adaption methods based on the classical particle swarm algorithm.
Finally, the RLPSO designed based on RLAM is compared with other current state-of-the-art algorithms.

In the following experiments, in order to test the algorithm performance more comprehensively, we selected the CEC2013 test set \cite{liang2013problem} for testing.
The test set includes 28 test functions that can simulate a wide variety of optimization problems.
The test functions of CEC2013 are shown in the table \ref{TABLE-CEC2013}:
% todo 此处介绍具体函数

\begin{table}[]
\caption{All test functions and their optimal values in CEC2013.}\label{TABLE-CEC2013}
%\resizebox{\textwidth}{!}
\centering
{
\begin{tabular}{lll}
No.  & Function Name                               & best value   \\
1    & Sphere Function                             & -1400 \\
2    & Rotated High Conditioned Elliptic Function  & -1300 \\
3    & Rotated Bent Cigar Function                 & -1200 \\
4    & Rotated Discus Function                     & -1100 \\
5    & Different Powers Function                   & -1000 \\
6    & Rotated Rosenbrock’s Function               & -900  \\
7    & Rotated Schaffers F7 Function               & -800  \\
8    & Rotated Ackley’s Function                   & -700  \\
9    & Rotated Weierstrass Function                & -600  \\
10   & Rotated Griewank’s Function                 & -500  \\
11   & Rastrigin’s Function                        & -400  \\
12   & Rotated Rastrigin’s Function                & -300  \\
13   & Non-Continuous Rotated Rastrigin’s Function & -200  \\
14   & Schwefel's Function                         & -100  \\
15   & Rotated Schwefel's Function                 & 100   \\
16   & Rotated Katsuura Function                   & 200   \\
17   & Lunacek Bi\_Rastrigin Function              & 300   \\
18   & Rotated Lunacek Bi\_Rastrigin Function      & 400   \\
19   & Expanded Griewank’s plus Rosenbrock’s Function 500 \\
	 & 20 Expanded Scaffer’s F6 Function          & 500 \\
20   & Expanded Scaffer’s F6 Function             & 600   \\
21   & Composition Function 1 (n=5,Rotated)        & 700   \\
22   & Composition Function 2 (n=3,Unrotated)      & 800   \\
23   & Composition Function 3 (n=3,Rotated)        & 900   \\
24   & Composition Function 4 (n=3,Rotated)        & 1000  \\
25   & Composition Function 5 (n=3,Rotated)        & 1100  \\
26   & Composition Function 6 (n=5,Rotated)        & 1200  \\
27   & Composition Function 7 (n=5,Rotated)        & 1300  \\
28   & Composition Function 8 (n=5,Rotated)        & 1400 
\end{tabular}
}
\end{table}

The domain of all functions is -100 to 100.
Due to the relatively large overall testing volume, in all runs, the end condition for all algorithms is to complete 10,000 evaluations.

The proposed algorithm has been impolemented using Python 3.9 under 64-bit Ubuntu 16.04.7 LTS operating system. 
Experiments are conducted on a server with Intel Xeon Silver 4116, 
2.1 GHz CPU and 128 GB of RAM. 

\subsection{Improvement of PSO variants after combining with RLAM}

In this experiment, we compare the performance of various PSO variants incorporating RLAM with the original version.
The combination of PSO and RLAM is as described above.
For each problem, each algorithm will be run 50 times, the test dimension is 30 dimensions, and the final result will be shown in the table \ref{EXP1-TAB1}.

\begin{figure}
  \centering
  \includegraphics[width=\textwidth]{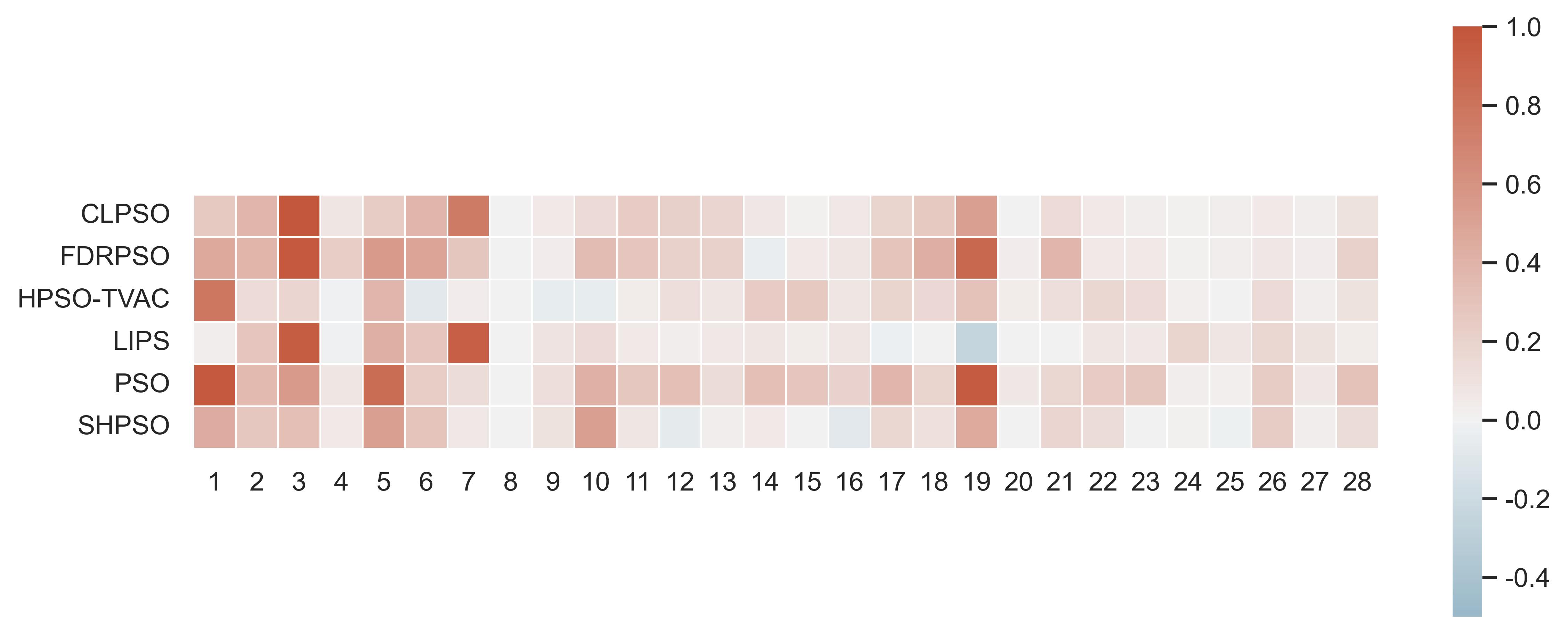}
  \caption{heatmap of improvement}
  \label{rlma-heatmap}
\end{figure}

\begin{table}[]
\caption{Improvement of PSO variants after combining with RLAM.}\label{EXP1-TAB1}
%\resizebox{\textwidth}{!}
\centering
{
\begin{tabular}{lllllll}
\hline 
        & CLPSO & FDRPSO & HPSO-TVAC & LIPS & PSO & SHPSO \\
\hline
1   & 26.0\% & 46.2\% & 78.32\%   & 3.19\%   & 97.62\% & 44.7\% \\
2   & 37.6\% & 38.8\% & 14.26\%   & 28.35\%  & 35.37\% & 26.8\% \\
3   & 98.7\% & 97.3\% & 18.47\%   & 94.85\%  & 55.70\% & 32.2\% \\
4   & 7.8\%  & 22.8\% & -1.01\%   & -1.09\%  & 7.06\%  & 5.5\%  \\
5   & 23.9\% & 54.9\% & 38.45\%   & 42.49\%  & 84.27\% & 52.5\% \\
6   & 37.4\% & 49.6\% & -8.33\%   & 27.94\%  & 22.82\% & 29.3\% \\
7   & 75.0\% & 27.8\% & 3.67\%    & 93.30\%  & 13.61\% & 6.2\%  \\
8   & 0.3\%  & 0.2\%  & 0.17\%    & 0.12\%   & 0.22\%  & 0.1\%  \\
9   & 5.5\%  & 3.5\%  & -4.42\%   & 9.17\%   & 11.59\% & 10.1\% \\
10  & 15.5\% & 34.0\% & -4.43\%   & 15.59\%  & 41.57\% & 52.2\% \\
11  & 25.0\% & 28.5\% & 4.27\%    & 4.52\%   & 26.94\% & 7.8\%  \\
12  & 21.7\% & 20.8\% & 12.58\%   & 2.73\%   & 32.11\% & -6.4\% \\
13  & 18.3\% & 20.7\% & 8.57\%    & 5.77\%   & 13.62\% & 3.0\%  \\
14  & 6.5\%  & -3.9\% & 24.55\%   & 8.28\%   & 32.44\% & 4.9\%  \\
15  & 1.4\%  & 5.5\%  & 26.54\%   & 3.99\%   & 28.49\% & 0.9\%  \\
16  & 6.0\%  & 8.3\%  & 7.06\%    & 7.90\%   & 19.88\% & -8.0\% \\
17  & 18.8\% & 29.7\% & 19.61\%   & -1.97\%  & 37.90\% & 16.3\% \\
18  & 26.1\% & 43.2\% & 15.95\%   & 0.92\%   & 19.59\% & 11.2\% \\
19  & 52.9\% & 87.4\% & 31.36\%   & -24.77\% & 95.46\% & 45.7\% \\
20  & 0.0\%  & 3.5\%  & 4.08\%    & 0.00\%   & 6.32\%  & 0.8\%  \\
21  & 14.3\% & 37.4\% & 12.33\%   & 0.74\%   & 16.82\% & 18.0\% \\
22  & 5.5\%  & 5.0\%  & 17.34\%   & 7.54\%   & 24.48\% & 13.5\% \\
23  & 2.9\%  & 5.2\%  & 14.87\%   & 5.68\%   & 27.27\% & 0.2\%  \\
24  & 1.5\%  & 1.2\%  & 1.60\%    & 19.45\%  & 3.26\%  & 1.0\%  \\
25  & 3.2\%  & 2.6\%  & 0.34\%    & 6.86\%   & 1.57\%  & -1.7\% \\
26  & 5.5\%  & 6.7\%  & 15.05\%   & 17.34\%  & 24.05\% & 23.4\% \\
27  & 2.2\%  & 3.7\%  & 2.66\%    & 9.70\%   & 6.45\%  & 2.2\%  \\
28  & 10.1\% & 20.7\% & 9.99\%    & 3.99\%   & 30.77\% & 13.6\% \\
\hline
AVERAGE & 19.6\% & 25.0\% & 13.0\%    & 14.0\%   & 29.2\%  & 14.5\%   \\  
\hline
\end{tabular}
}
\end{table}

\begin{table}[]
\caption{Pairwise comparison between PSO and PSO with RLAM}\label{EXP1-WILCOXON}
\centering
\begin{tabular}{lllllll}
         & CLPSO    & FDRPSO   & HPSO-TVAC & LIPS     & PSO      & SHPSO    \\
improve  & 28       & 27       & 24        & 25       & 28       & 25       \\
decrease & 0        & 1        & 4         & 3        & 0        & 3        \\
p-value  & 3.79E-06 & 8.07E-06 & 1.19E-04  & 1.19E-04 & 3.79E-06 & 9.86E-05
\end{tabular}
\end{table}

In the table \ref{EXP1-TAB1}, the result is calculated as follows:

\begin{equation}
improve = (Gbest_{origin}-Gbest_{train})/(Gbest_{origin}-benchmark_{best})
\end{equation}

Where $improve$ is the percentage of improvement, which is the data in the table. $Gbest_{origin}$ is the optimal value obtained by the original version of the PSO variant. $Gbest_{train}$ is the optimal value obtained by the RLMA version of the PSO variant.
Each row represents the result on a different evaluation function.
$benchmark_{best}$ represents the optimal value of the current test function.

In order to show the improvement more intuitively, Figure \ref{rlma-heatmap} shows the heatmap of the improvement of various PSO algorithms after adding RLMA.

From table \ref{EXP1-TAB1} and figure \ref{rlma-heatmap} we can see that almost all algorithms have significant improvement after incorporating RLAM on all test functions. The average improvement ranged from 13\% to 29.2\%, and the overall average improvement was 19.2\%.

In Unimodal's test function (F01-F05), most of the PSOs have a relatively large improvement after combining with RLAM. In the Multimodal test function (F06-F20), although the overall improvement is not as good as that of Unimodal, most of them are also significantly improved, and a few are significantly improved. In the Composition test function (F21-F28), the overall improvement is smaller than the first two categories, but it still has a certain effect. From these conclusions, it can be seen that RLAM can generally achieve a very good improvement effect on simple problems, and the improvement rate has decreased on complex problems, but it still has a significant improvement effect.

In addition, for a more comprehensive statistical analysis, this paper carries out nonparametric statistical tests. Table \ref{EXP1-WILCOXON} presents the results of the pairwise comparison including the number of increases, and decreases and p-value of Wilcoxon test\cite{xu2020reinforcement,Derrac2011}. we have normalized the results on every function to be in $[0,1]$ according to the best and worst results obtained by all the algorithms\cite{Garcia-Martinez2010}. 
As can be seen from the table, out of all 6 tested PSO methods, two algorithms improve in all test functions after combining RLAM. 
The worst among the other algorithms decrease in 4 of the 28 test functions. 
Overall, among the 168 combinations of test algorithms and test functions, a total of 11 combinations have decreased, and the rest have improved, and the proportion of effective results is as high as 93.5%.
According to the results of Wilcoxon test, RLAM is significantly effective with a level of significance $\alpha = 0.05$.

\subsection{campare to other online adaption methods}

To test the pros and cons of RLAM with other online adaption methods, RLAM is compared with 4 other online adaption methods in this section.
The other four algorithms are: adaptation algorithm based on type 1 fuzzy logic\cite{Melin2013}, adaptation algorithm based on type 2 fuzzy logic\cite{Olivas2016}, adaptation algorithm based on success rate history\cite{Tanabe2013} and adaptation algorithm based on Qlearning\cite{Liu2019}.

In the result, PSO with RLAM is labeled with PSOtrain, PSO with type 1 fuzzy logic is labeled with TF1PSO, PSO with type 2 fuzzy logic is labeled with TF2PSO, PSO with adaptation algorithm based on success rate history is labeled with SuccessHistoryPSO, PSO with adaptation algorithm based on Qlearning is labeled with QLPSO.
\begin{table}[]
\caption{Convergence accuracy comparison between RLAM and other online adaption method.}\label{online-adaption-test-table1}
%\resizebox{\textwidth}{!}
\centering
{
\begin{tabular}{llllll}
No & FT1PSO    & FT2PSO    & SuccessHistoryPSO & QLPSO     & PSOtrain  \\
1        & 3.47E+02  & 6.92E+02  & 3.16E+02          & 2.78E+02  & \textbf{-1.33E+03} \\
2        & 3.38E+07  & 3.57E+07  & 5.10E+07          & 2.68E+07  & \textbf{2.05E+07}  \\
3        & 1.85E+10  & 2.56E+10  & 2.57E+10          & 2.05E+10  & \textbf{8.66E+09}  \\
4        & 6.76E+04  & 7.02E+04  & 6.92E+04          & \textbf{5.57E+04}  & 5.92E+04  \\
5        & -4.27E+02 & -1.70E+02 & 1.17E+02          & \textbf{-8.42E+02} & -8.20E+02 \\
6        & -6.67E+02 & -6.75E+02 & -6.23E+02         & -7.82E+02 & \textbf{-8.16E+02} \\
7        & \textbf{-6.77E+02} & -6.50E+02 & -6.40E+02         & -6.76E+02 & -6.70E+02 \\
8        & -6.79E+02 & -6.79E+02 & -6.79E+02         & -6.79E+02 & \textbf{-6.79E+02} \\
9        & -5.72E+02 & -5.72E+02 & -5.71E+02         & -5.71E+02 & \textbf{-5.75E+02} \\
10       & -2.31E+01 & -2.11E+01 & 4.84E+01          & -2.18E+02 & \textbf{-3.22E+02} \\
11       & -2.23E+02 & -2.13E+02 & -2.21E+02         & -2.82E+02 & \textbf{-2.83E+02} \\
12       & -1.01E+02 & -8.81E+01 & -4.00E+01         & -1.30E+02 & \textbf{-1.40E+02} \\
13       & 1.03E+02  & 1.09E+02  & 1.21E+02          & 3.37E+01  & \textbf{1.71E+01}  \\
14       & 3.44E+03  & 3.56E+03  & 3.26E+03          & 5.34E+03  & \textbf{2.79E+03}  \\
15       & 5.48E+03  & 4.84E+03  & 4.93E+03          & 7.11E+03  & \textbf{4.26E+03}  \\
16       & 2.03E+02  & 2.03E+02  & \textbf{2.03E+02}          & 2.03E+02  & 2.03E+02  \\
17       & 4.96E+02  & 5.21E+02  & 5.41E+02          & 4.94E+02  & \textbf{4.61E+02}  \\
18       & 6.49E+02  & 6.75E+02  & 6.52E+02          & \textbf{6.47E+02}  & 6.72E+02  \\
19       & 2.23E+03  & 8.66E+02  & 1.25E+03          & 9.46E+02  & \textbf{5.15E+02}  \\
20       & 6.14E+02  & 6.14E+02  & 6.14E+02          & \textbf{6.14E+02}  & 6.14E+02  \\
21       & 1.74E+03  & 1.52E+03  & 1.96E+03          & 1.22E+03  & \textbf{9.96E+02}  \\
22       & 5.23E+03  & 5.12E+03  & 5.49E+03          & 5.63E+03  & \textbf{4.37E+03}  \\
23       & 6.21E+03  & 6.19E+03  & \textbf{5.80E+03}          & 8.29E+03  & 6.30E+03  \\
24       & 1.29E+03  & 1.28E+03  & 1.30E+03          & 1.28E+03  & \textbf{1.28E+03}  \\
25       & 1.41E+03  & 1.41E+03  & 1.41E+03          & \textbf{1.39E+03}  & 1.40E+03  \\
26       & 1.45E+03  & 1.54E+03  & 1.52E+03          & \textbf{1.44E+03}  & 1.44E+03  \\
27       & 2.32E+03  & 2.34E+03  & 2.35E+03          & 2.38E+03  & \textbf{2.28E+03}  \\
28       & 3.52E+03  & 3.51E+03  & 3.53E+03          & 2.93E+03  & \textbf{2.16E+03} 
\end{tabular}
}
\end{table}

\begin{figure}
  \centering
  \includegraphics[width=\textwidth]{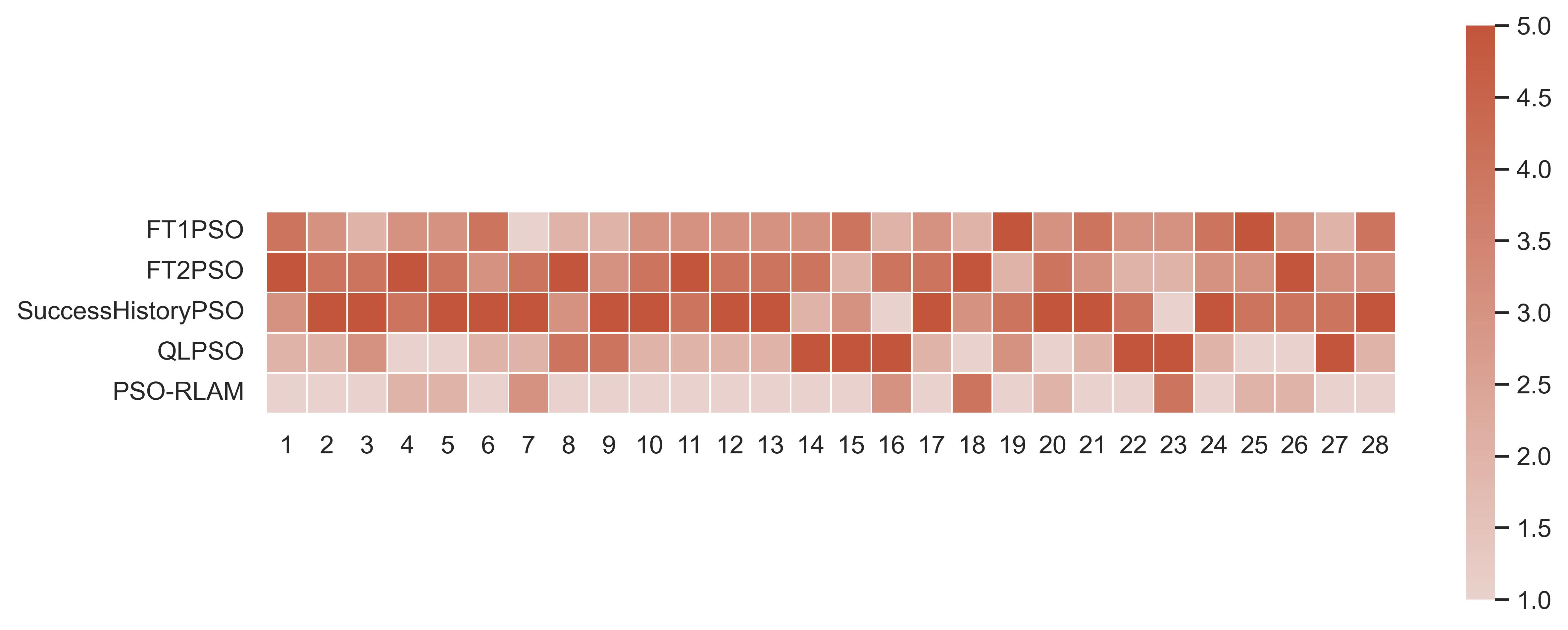}
  \caption{Multiple comparison between RLMA and other online tuning}
  \label{ONLINE-RANK-HEATMAP}
\end{figure}

\begin{table}[]
\caption{Pairwise comparison
between PSO with RLAM and other
online tuning method}\label{ONLINETUNE-EXP-WINLOSS}
\resizebox{\textwidth}{!}
{
\begin{tabular}{lllll}
PSO-TRAIN VS. & FT1PSO      & FT2PSO      & SuccessHistoryPso & QLPSO       \\
WIN           & 24          & 27          & 25                & 21          \\
LOSS          & 4           & 1           & 3                 & 7           \\
p-value       & 6.74804E-05 & 4.21632E-06 & 1.45882E-05       & 0.001319008
\end{tabular}
}
\end{table}

The best results obtained by PSO with RLAM and PSOs with
other online tuning methods are listed in Table \ref{online-adaption-test-table1}, and the best
results are also shown in bold. 
In this table some data are the same but only one of the cases marked in bold is due to insufficient numerical resolution, and the finer results will show their differences.
In this table, the fuzzy logic based adaptation performed the worst, where the type 1 fuzzy logic based adaptation won the first place in F7, while the type 2 fuzzy logic based adaptation did not get any first place.
And the adaptation based on successful history won the first place in F16, F23, while the adaptation based on reinforcement learning performed the best in the test. Among them, the adaptation based on Q-learning won the first place in F4, F5, F18, F20, F25, F26, and RLAM won the first place in a total of 19 other functions.

Fig. 11 describes the ranking of these algorithms for a multiple comparison.
From the figure \ref{ONLINE-RANK-HEATMAP} we can see that RLAM has the lightest overall color and the highest average ranking on the entire function test set.
The adaptation based on Q-learning also works well, but it achieved the last place in the 6 test functions of F14, F15, F16, F22, F23 and F27.
The effect of other algorithms is not much different. Among them, the adaptation algorithm based on type 1 fuzzy logic ranks better overall, with almost no worst results, indicating that its robustness is better.

Tables \ref{ONLINETUNE-EXP-WINLOSS} are pairwise comparisons
between PSOs with online tuning methods. It is clear that
PSO with RLAM is significantly better than FT1PSO, FT2PSO, success-history-PSO, QLPSO with a level of significance $\alpha = 0.05$.

Overall, RLAM has an excellent performance on all test
functions when comparing with other online tuning methods.

\subsection{Comparison of RLPSO with other state-of-the-art PSO variants}

Since most of the current state-of-the-art particle swarm algorithms combine many other optimization methods, adjusting parameters alone cannot make a good comparison.
In order to test the strength of RLAM applied in particle swarm optimization, here we will compare the performance of RLPSO based on RLAM design and some other improved particle swarm optimization algorithms.
The algorithms compared include some of the widely used particle swarm algorithms:CLPSO\cite{Liang2006},FDR-PSO\cite{Peram2003},LIPS\cite{Qu2013}, PSO\cite{kennedy1995particle} and SHPSO\cite{Engelbrecht2010}. And several state-of-the-art
variants of PSO including EPSO\cite{Lynn2017}, AWPSO\cite{Liu2021} and PPPSO\cite{Zhang2018}. 
All algorithms are executed 50 times for each problem and the results are the average.

\begin{table}[]
\caption{Comparison between RLPSO and other PSO variants}\label{RLPSO-EXP-TABLE1}
\resizebox{\textwidth}{!}
{
\begin{tabular}{llllllllll}
\hline 
No & CLPSO     & FDRPSO    & LIPS      & PSO       & SHPSO     & AWPSO     & PPPSO     & EPSO      & RLPSO     \\
\hline

1  & 5.19E+04  & 2.75E+04  & 5.06E+04  & -8.12E+02 & -1.18E+03 & 7.39E+02  & -2.16E+02 & -1.37E+03 & \textbf{-1.40E+03} \\
2  & 8.98E+08  & 3.46E+08  & 1.49E+09  & 3.10E+07  & 6.42E+07  & 2.59E+07  & 3.33E+07  & 3.57E+07  & \textbf{1.72E+07}  \\
3  & 9.09E+16  & 2.75E+12  & 7.56E+18  & 1.47E+10  & 4.47E+10  & \textbf{9.83E+09}  & 3.05E+10  & 1.34E+10  & 1.14E+10  \\
4  & 1.03E+05  & 1.21E+05  & 6.71E+04  & 6.06E+04  & 5.88E+04  & 6.18E+04  & 6.18E+04  & 5.74E+04  & \textbf{5.71E+04}  \\
5  & 2.03E+04  & 8.23E+03  & 3.01E+04  & -3.13E+02 & -7.76E+02 & -2.85E+02 & -2.64E+02 & -9.54E+02 & \textbf{-9.55E+02} \\
6  & 8.86E+03  & 9.39E+02  & 1.32E+04  & -8.02E+02 & -7.37E+02 & -7.88E+02 & -7.73E+02 & -7.59E+02 & \textbf{-8.07E+02} \\
7  & 6.78E+04  & -4.79E+02 & 4.53E+06  & -6.52E+02 & -6.03E+02 & -6.10E+02 & -6.29E+02 & \textbf{-6.83E+02} & -6.79E+02 \\
8  & -6.79E+02 & -6.79E+02 & -6.79E+02 & -6.79E+02 & -6.79E+02 & \textbf{-6.79E+02} & -6.79E+02 & -6.79E+02 & -6.79E+02 \\
9  & -5.58E+02 & -5.58E+02 & -5.53E+02 & -5.73E+02 & -5.62E+02 & -5.75E+02 & -5.73E+02 & -5.67E+02 & \textbf{-5.75E+02} \\
10 & 6.87E+03  & 2.81E+03  & 8.24E+03  & -3.58E+02 & -5.43E+01 & -1.11E+02 & -2.14E+02 & -4.00E+02 & \textbf{-4.38E+02}] \\
11 & 5.24E+02  & 2.26E+02  & 4.78E+02  & -2.58E+02 & -2.59E+02 & \textbf{-2.95E+02} & -2.60E+02 & -1.45E+02 & -2.88E+02 \\
12 & 6.04E+02  & 2.72E+02  & 5.76E+02  & -8.19E+01 & -3.41E+01 & -7.22E+01 & -1.06E+02 & -9.19E+00 & \textbf{-1.40E+02} \\
13 & 7.30E+02  & 3.58E+02  & 6.81E+02  & 4.64E+01  & 1.16E+02  & 5.73E+01  & \textbf{2.58E+01}  & 1.13E+02  & 5.62E+01  \\
14 & 7.19E+03  & 7.45E+03  & 8.80E+03  & 4.19E+03  & 3.86E+03  & 4.47E+03  & 4.43E+03  & 7.10E+03  & \textbf{2.76E+03}  \\
15 & 8.03E+03  & 8.54E+03  & 8.48E+03  & 6.72E+03  & 6.21E+03  & 7.48E+03  & 7.27E+03  & 7.98E+03  & \textbf{5.00E+03}  \\
16 & 2.04E+02  & 2.04E+02  & 2.04E+02  & 2.03E+02  & \textbf{2.02E+02}  & 2.04E+02  & 2.03E+02  & 2.04E+02  & 2.03E+02  \\
17 & 1.68E+03  & 1.60E+03  & 1.22E+03  & 5.42E+02  & 5.94E+02  & 5.51E+02  & 5.41E+02  & 6.20E+02  & \textbf{4.78E+02}  \\
18 & 2.00E+03  & 1.88E+03  & 1.33E+03  & 7.01E+02  & 7.57E+02  & 7.04E+02  & 6.95E+02  & 7.33E+02  & \textbf{6.79E+02}  \\
19 & 1.54E+06  & 9.37E+04  & 5.14E+05  & 9.59E+02  & 5.88E+02  & 1.37E+03  & 5.45E+02  & 5.36E+02  & \textbf{5.23E+02}  \\
20 & 6.15E+02  & 6.15E+02  & 6.15E+02  & 6.14E+02  & 6.14E+02  & 6.14E+02  & 6.14E+02  & 6.15E+02  & \textbf{6.14E+02}  \\
21 & 4.58E+03  & 4.27E+03  & 3.25E+03  & 1.10E+03  & 1.38E+03  & 1.21E+03  & 1.19E+03  & 1.11E+03  & \textbf{9.90E+02}  \\
22 & 9.00E+03  & 9.75E+03  & 1.04E+04  & 5.38E+03  & 5.42E+03  & 5.14E+03  & 4.92E+03  & 8.46E+03  & \textbf{4.07E+03}  \\
23 & 9.48E+03  & 9.48E+03  & 1.02E+04  & 7.95E+03  & \textbf{7.28E+03}  & 8.56E+03  & 8.18E+03  & 9.46E+03  & 7.64E+03  \\
24 & 1.37E+03  & 1.30E+03  & 1.54E+03  & 1.28E+03  & 1.30E+03  & \textbf{1.28E+03}  & 1.28E+03  & 1.30E+03  & 1.28E+03  \\
25 & 1.49E+03  & 1.41E+03  & 1.54E+03  & 1.40E+03  & 1.40E+03  & 1.39E+03  & \textbf{1.39E+03}  & 1.42E+03  & 1.40E+03  \\
26 & 1.51E+03  & 1.53E+03  & 1.64E+03  & 1.52E+03  & 1.49E+03  & 1.53E+03  & 1.56E+03  & 1.51E+03  & \textbf{1.44E+03}  \\
27 & 2.86E+03  & 2.62E+03  & 3.16E+03  & 2.31E+03  & 2.51E+03  & \textbf{2.21E+03}  & 2.28E+03  & 2.45E+03  & 2.27E+03  \\
28 & 7.79E+03  & 5.56E+03  & 7.89E+03  & 2.61E+03  & 3.04E+03  & 2.94E+03  & 2.42E+03  & 3.68E+03  & \textbf{2.20E+03} 
\end{tabular}
}
\end{table}

The best result obtained by RLPSO and other PSO variants are listed in Table \ref{RLPSO-EXP-TABLE1}, and the best results are also shown in bold.
In the results we can see that RLPSO took the first place in 18 of the 28 test functions, far exceeding several other PSO algorithms. SHPSO took first place on two functions. AWPSO took first place on 5 functions. PPPSO took first place on 2 functions. EPSO took first place on a function.

\begin{figure}
  \centering
  \includegraphics[width=\textwidth]{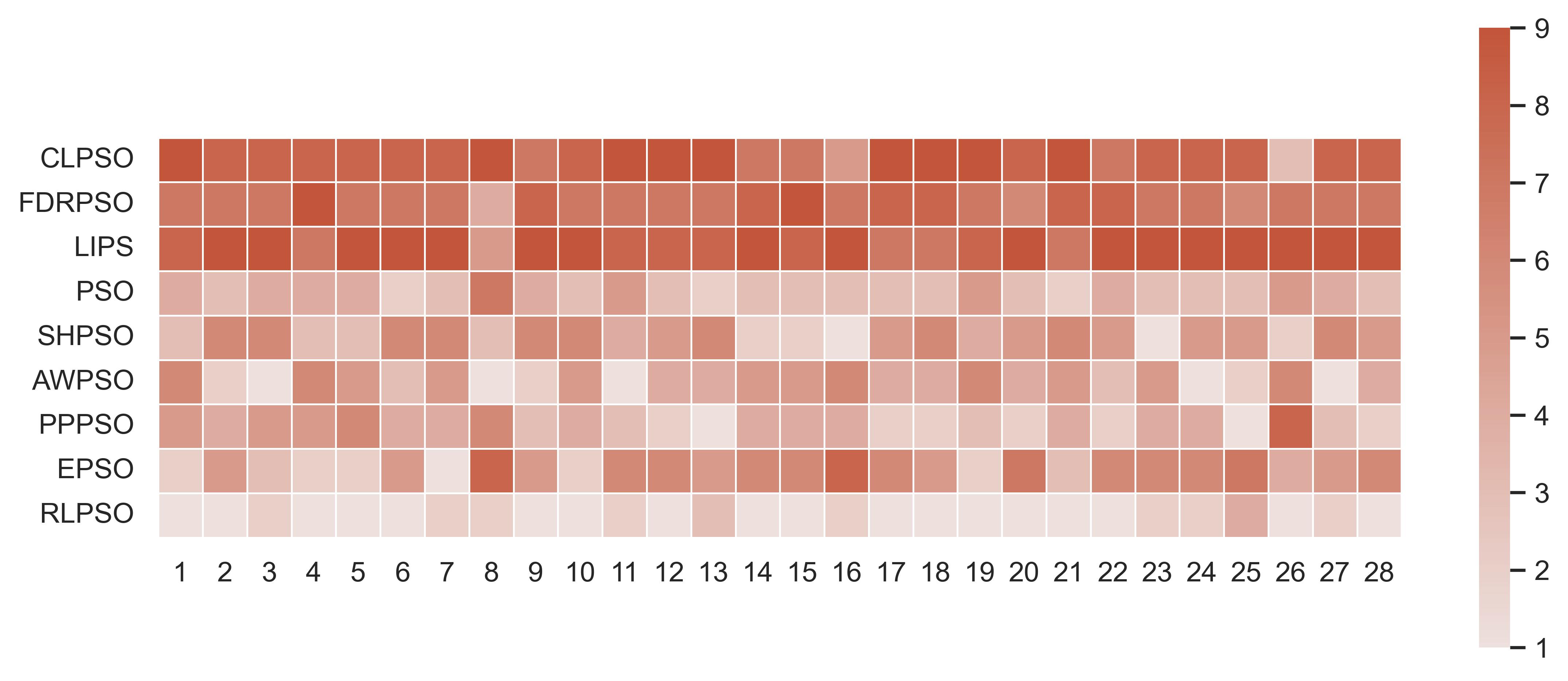}
  \caption{Multiple comparison between RLPSO and other PSO}
  \label{RLPSO-RANK-HEATMAP}
\end{figure}

Figure \ref{RLPSO-RANK-HEATMAP} shows the specific ranking of each algorithm on different test problems. It can be seen that RLPSO ranks very high on almost all problems, reflecting the excellent performance of RLPSO.

To show the stability of the proposed algorithm, convergence curves are depicted in Figs. \ref{RLPSO-RESULT}.

Tables \ref{RLPSO-EXP-TABLE3} are pairwise comparisons
between RLPSO. It is clear that
RLPSO is significantly better than other PSOs with a level of significance $\alpha = 0.05$.

In general, RLPSO has outstanding performance among many particle swarm algorithms, which shows that the RLAM method can obtain an excellent particle swarm variant algorithm after a certain design, which also highlights the power of RLAM.

\begin{table}[]
\caption{Pairwise comparison
between RLPSO and other
sota PSO}\label{RLPSO-EXP-TABLE3}
\resizebox{\textwidth}{!}
{
\begin{tabular}{lllllllll}
RLPSO VS. & CLPSO & FDRPSO & LIPS & PSO & SHPSO & AWPSO & PPPSO & EPSO \\
WIN       & 28    & 28     & 28   & 26  & 26    & 22    & 26    & 27   \\
LOSS      & 0     & 0      & 0    & 2   & 2     & 6     & 2     & 1    \\
p-value & 3.69479E-06 & 3.78688E-06 & 3.35953E-06 & 2.28059E-05 & 0.000320816 & 0.002853853 & 0.000118906 & 4.71565E-06
\end{tabular}
}
\end{table}

\begin{figure*}[htbp]
	\centering
	\subfigure[F1]{
		\begin{minipage}[t]{0.5\columnwidth}
			\centering
			\includegraphics[width=\textwidth]{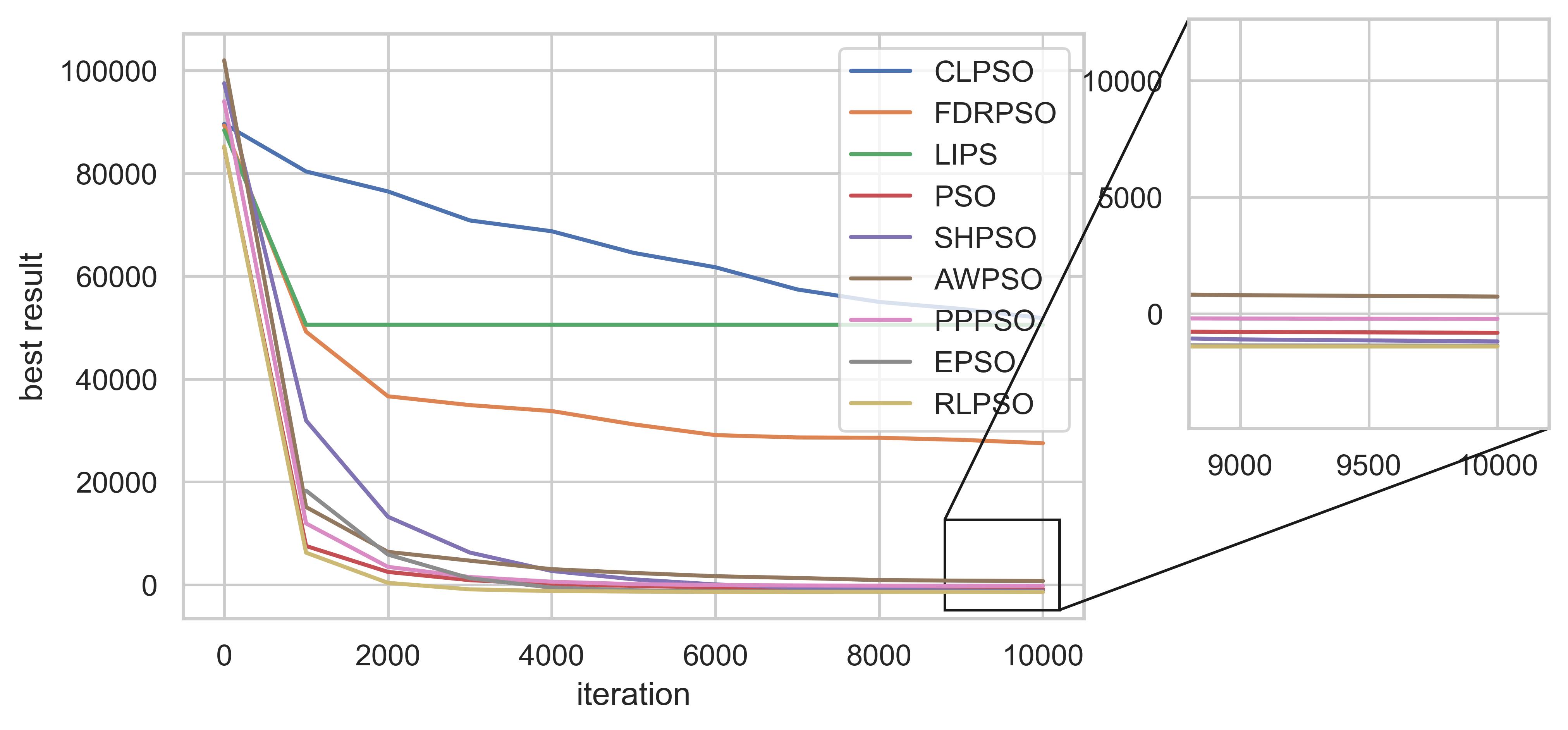}
		\end{minipage}
	}%
	\subfigure[F2]{
		\begin{minipage}[t]{0.5\columnwidth}
			\centering
			\includegraphics[width=\textwidth]{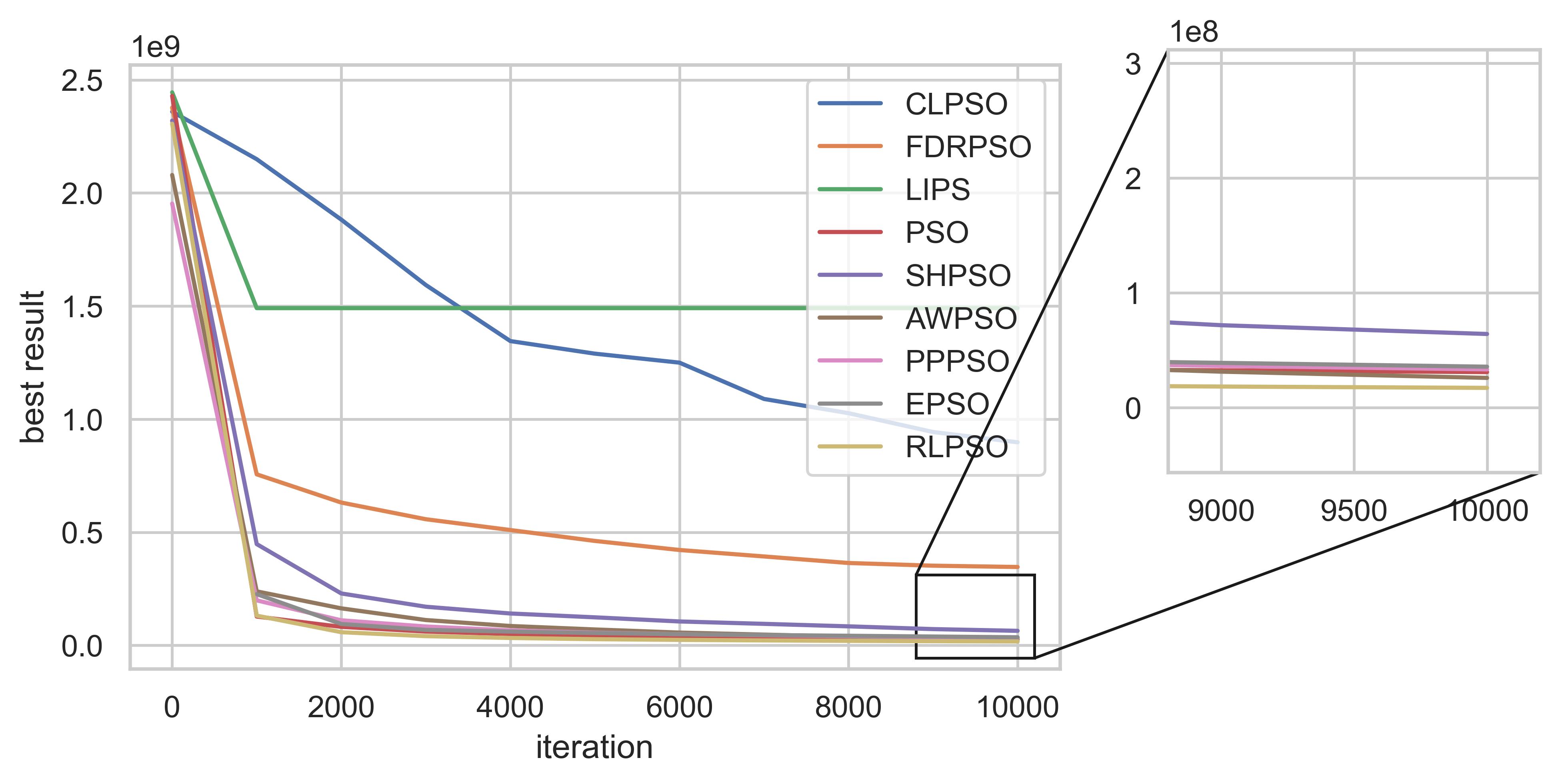}
		\end{minipage}
	}%
	%此处的空行很重要，想让图片在什么地方换行就在代码对应位置空行

	\subfigure[F3]{
		\begin{minipage}[t]{0.5\linewidth}
			\centering
			\includegraphics[width=\textwidth]{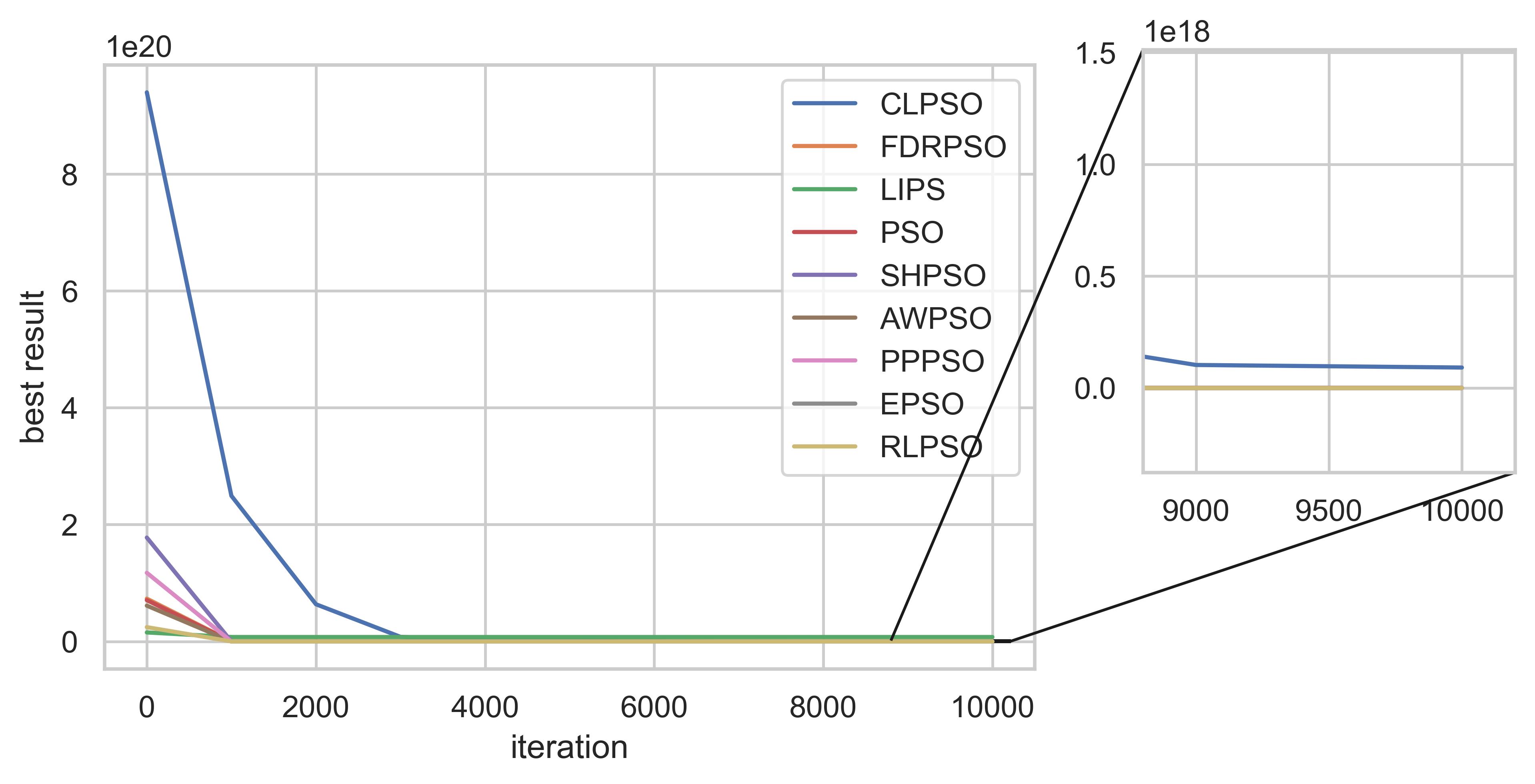}
		\end{minipage}
	}%
	\subfigure[F4]{
		\begin{minipage}[t]{0.5\linewidth}
			\centering
			\includegraphics[width=\textwidth]{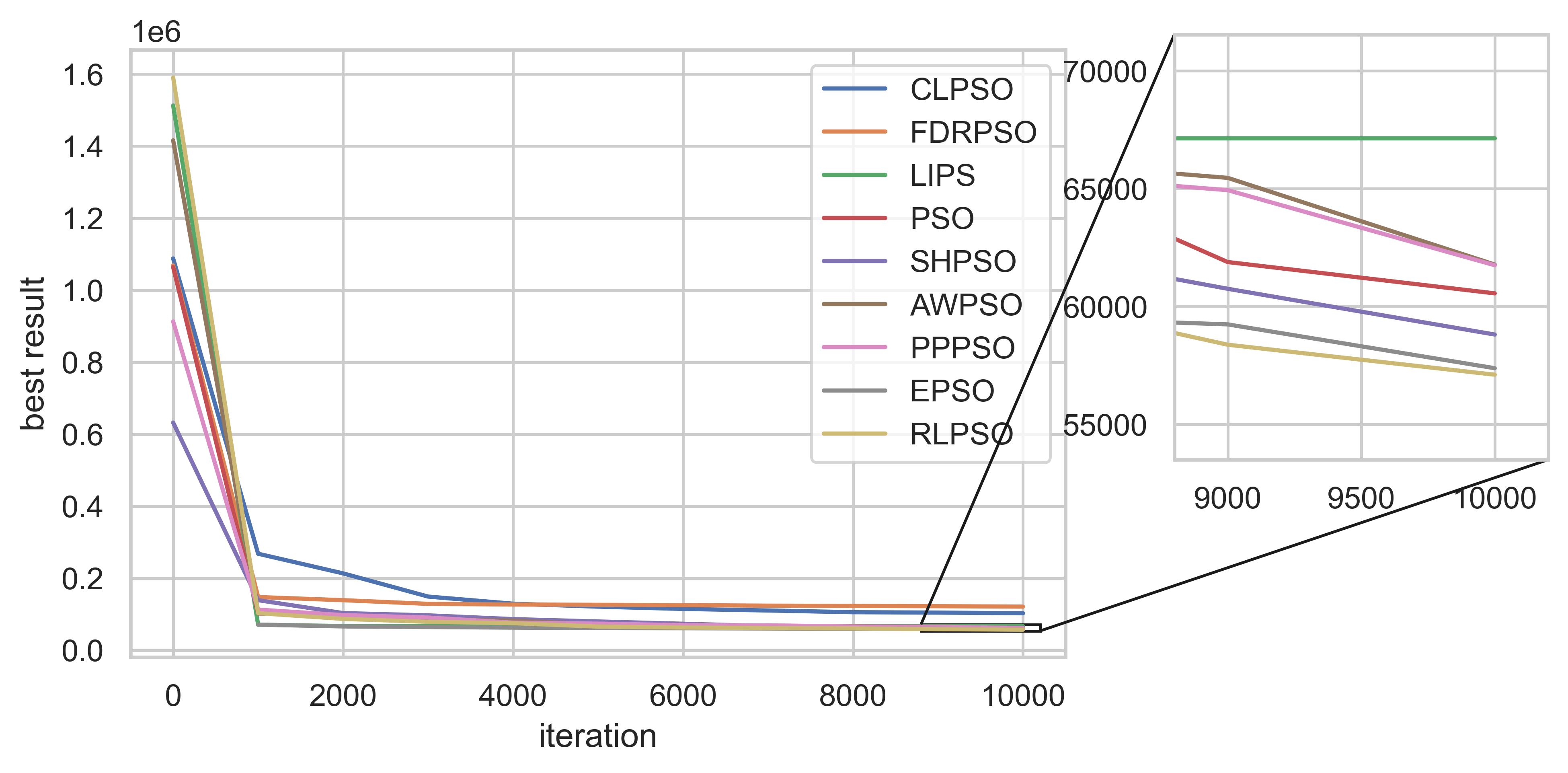}
		\end{minipage}
	}%

	\subfigure[F5]{
		\begin{minipage}[t]{0.5\linewidth}
			\centering
			\includegraphics[width=\textwidth]{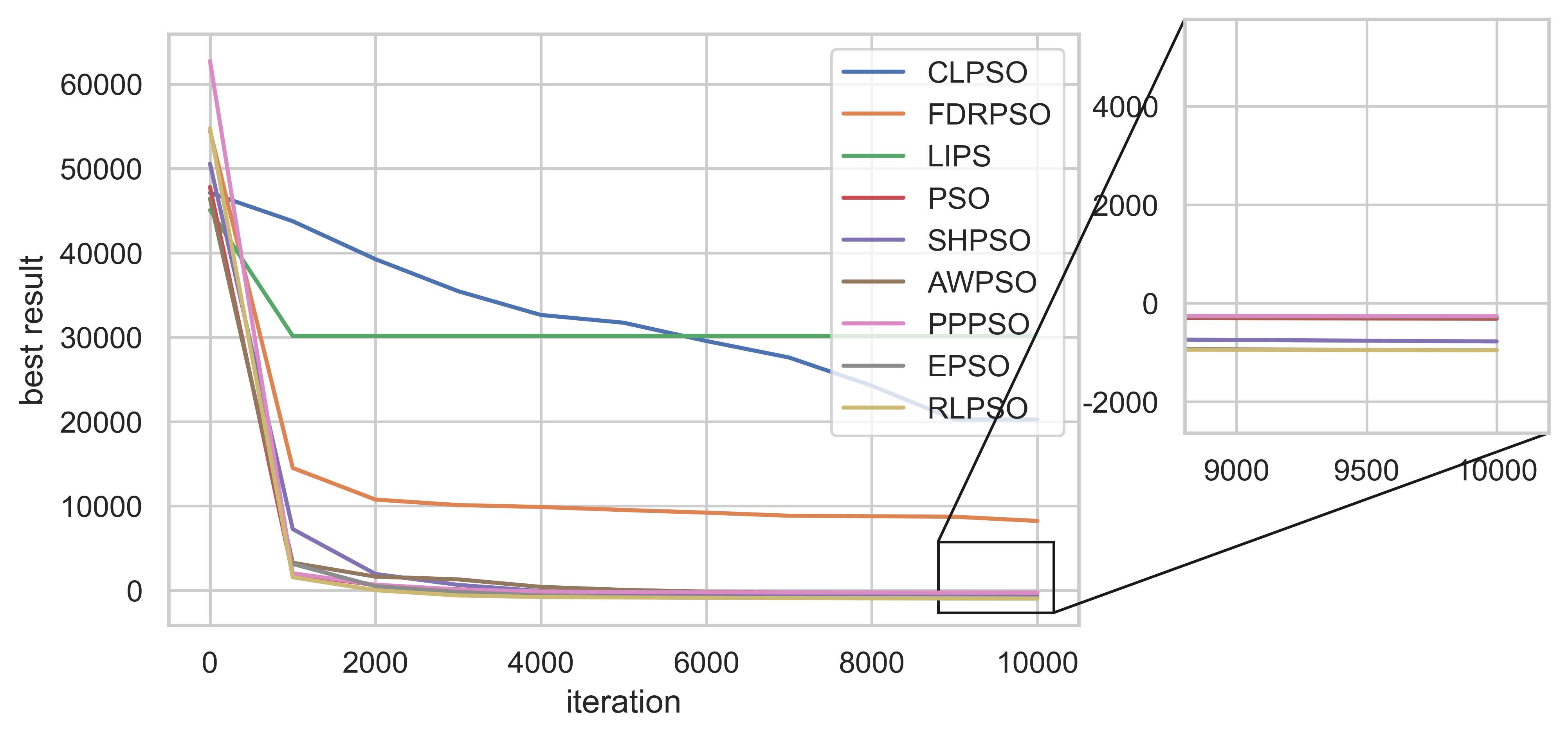}
		\end{minipage}
	}%
	\subfigure[F6]{
		\begin{minipage}[t]{0.5\linewidth}
			\centering
			\includegraphics[width=\textwidth]{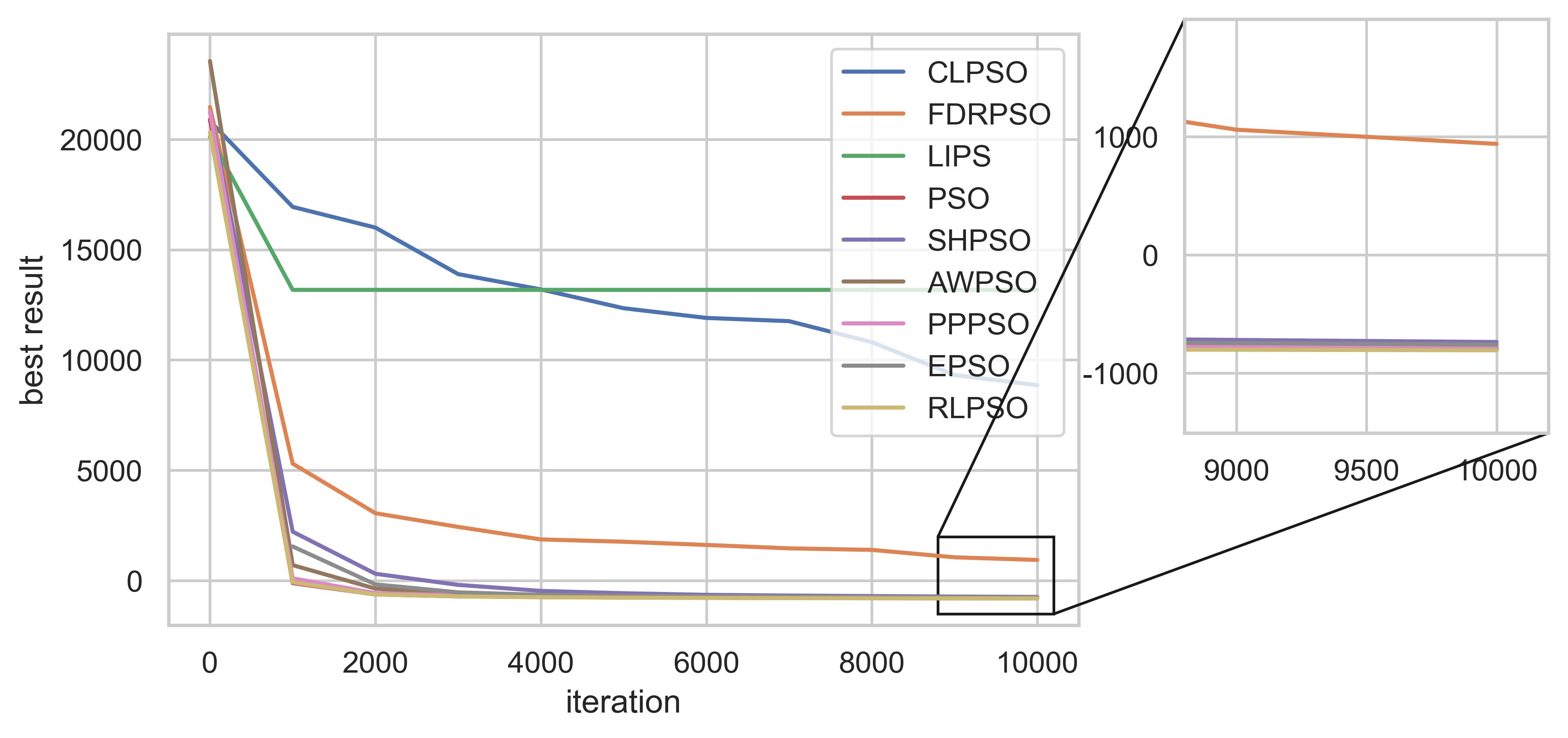}
		\end{minipage}
	}%

	\subfigure[F7]{
		\begin{minipage}[t]{0.5\linewidth}
			\centering
			\includegraphics[width=\textwidth]{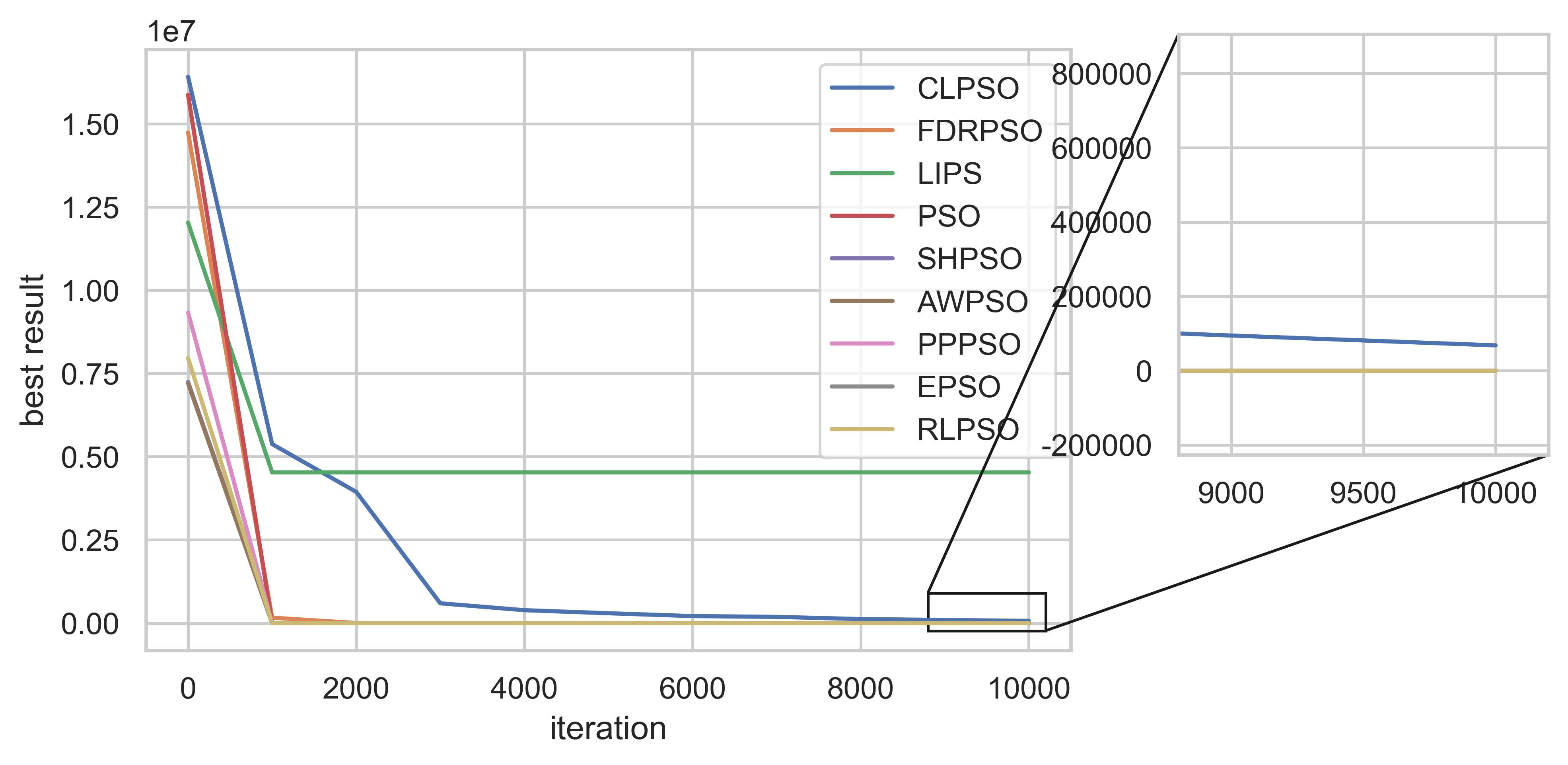}
		\end{minipage}
	}%
	\subfigure[F8]{
		\begin{minipage}[t]{0.5\linewidth}
			\centering
			\includegraphics[width=\textwidth]{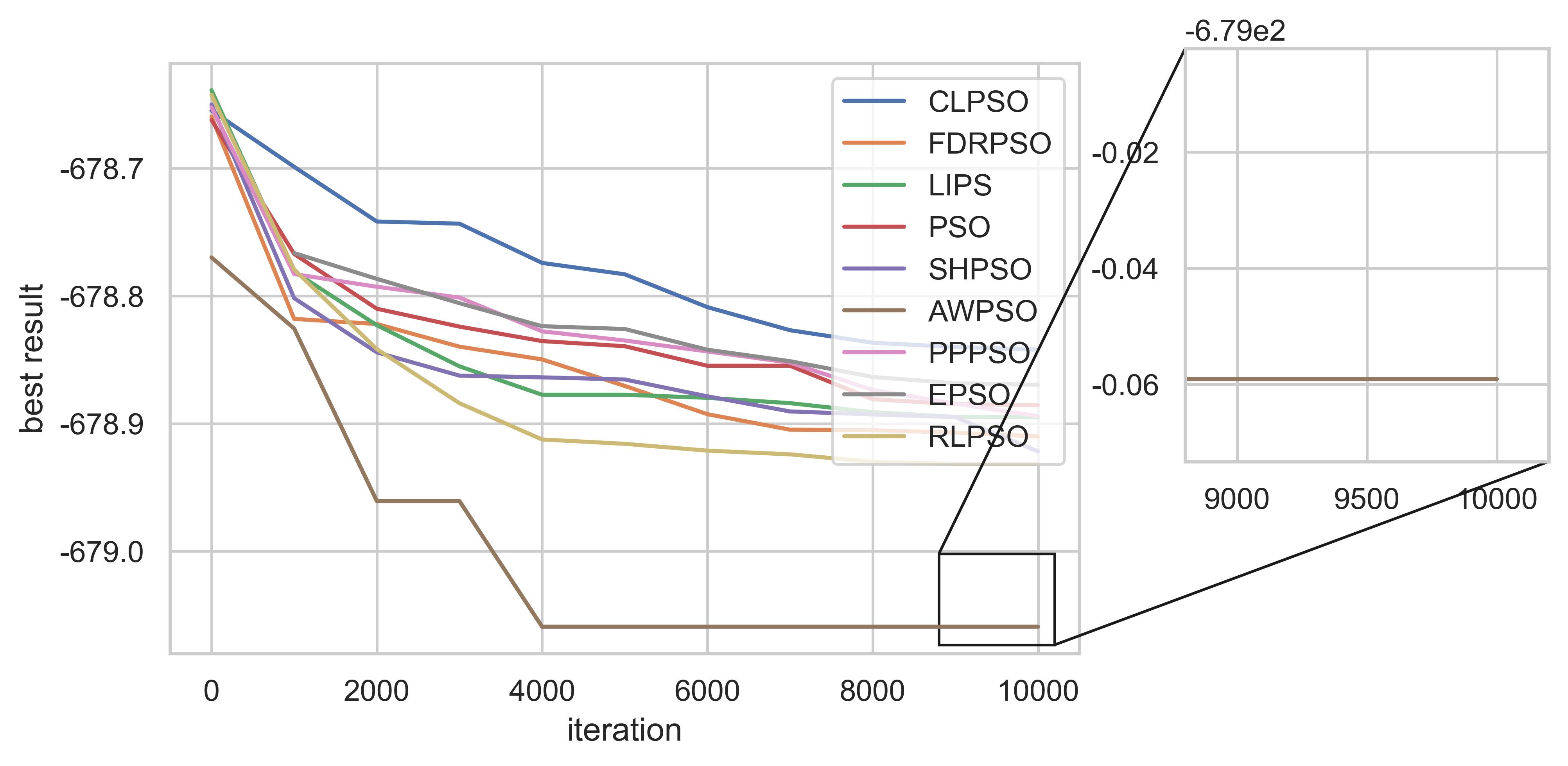}
		\end{minipage}
	}%

	\centering
	\caption{Convergence curves between RLPSO and other PSO variants}
	\label{RLPSO-RESULT}
\end{figure*}

\begin{figure*}
	\ContinuedFloat
	\subfigure[F9]{
		\begin{minipage}[t]{0.5\linewidth}
			\centering
			\includegraphics[width=\textwidth]{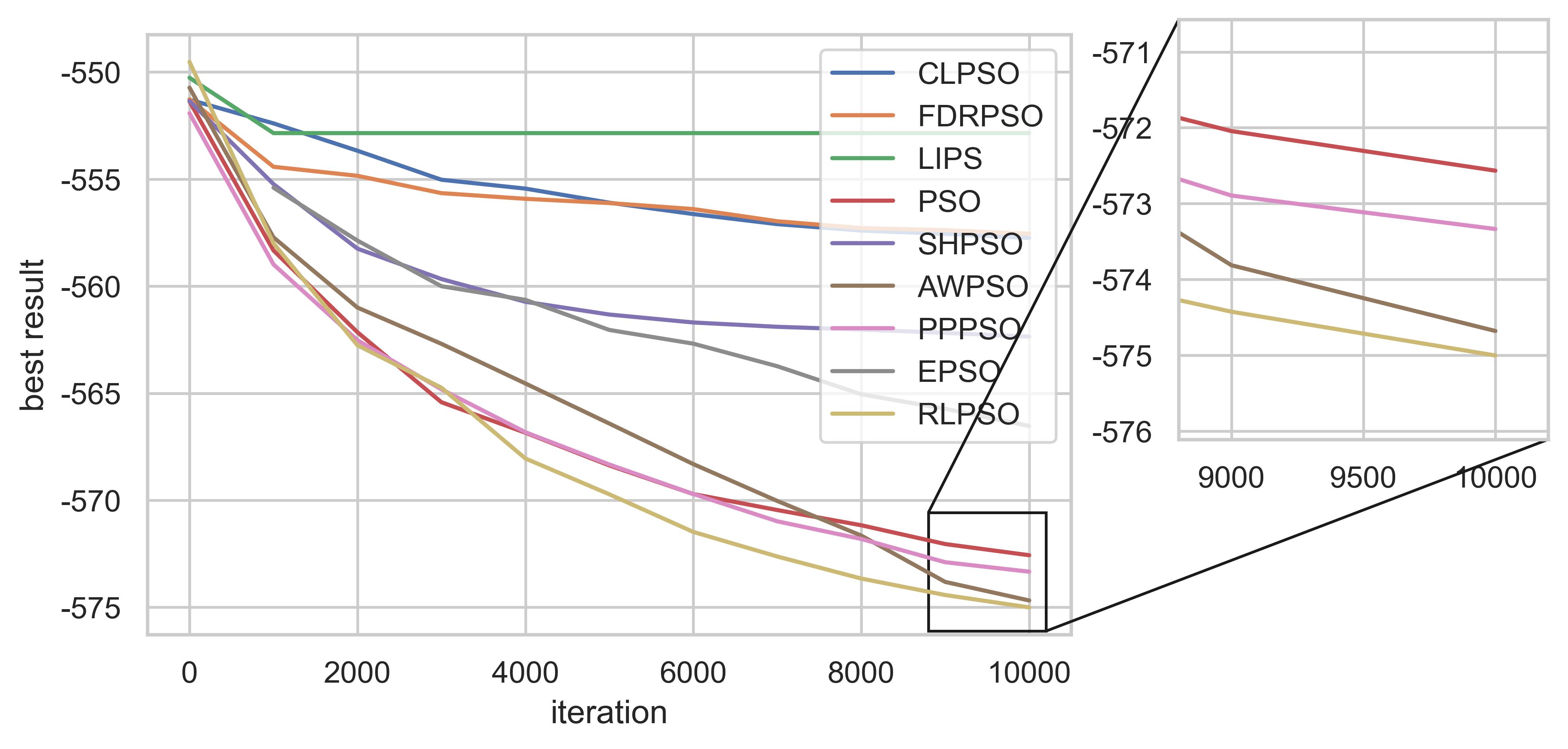}
		\end{minipage}
	}%
	\subfigure[F10]{
		\begin{minipage}[t]{0.5\linewidth}
			\centering
			\includegraphics[width=\textwidth]{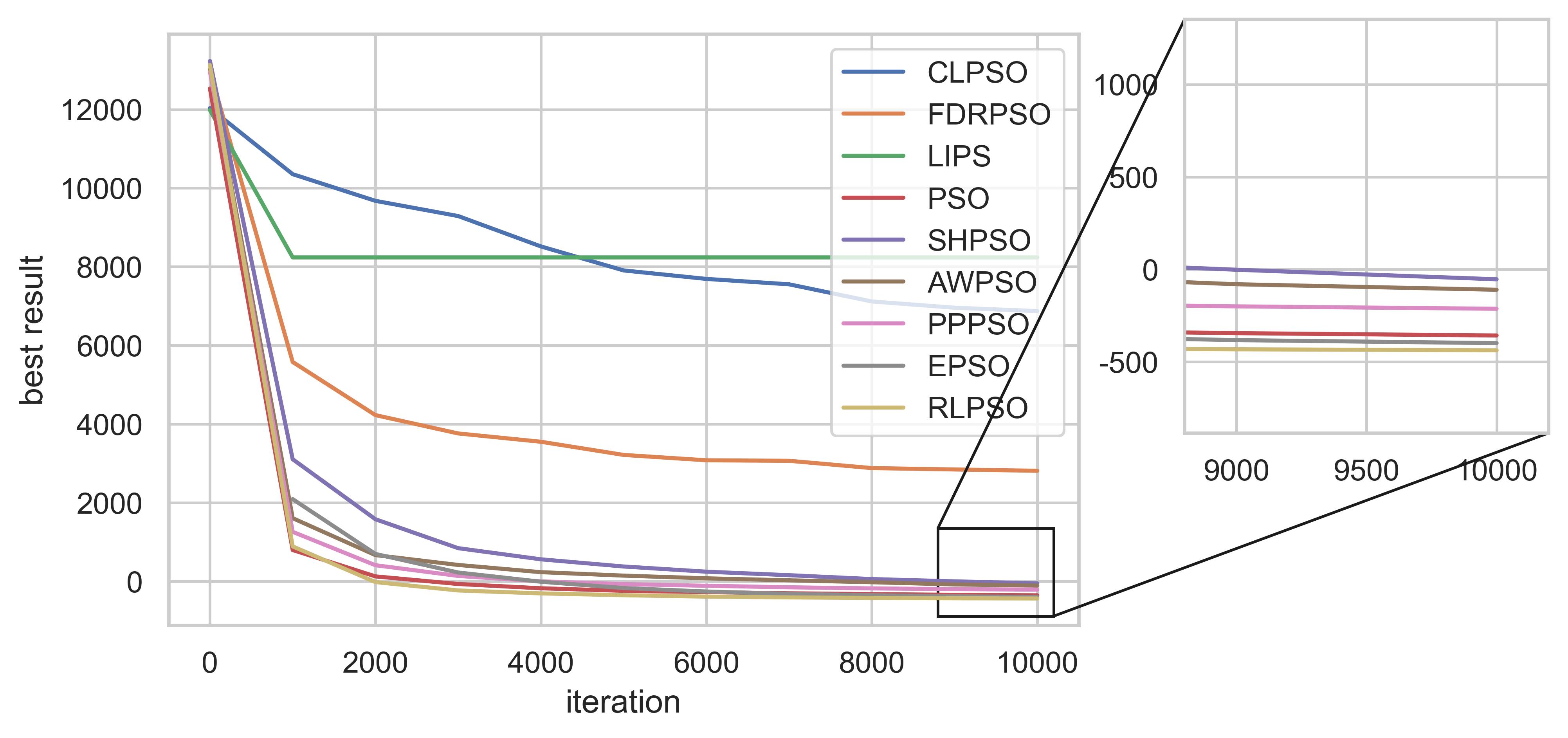}
		\end{minipage}
	}%

	\subfigure[F11]{
		\begin{minipage}[t]{0.5\linewidth}
			\centering
			\includegraphics[width=\textwidth]{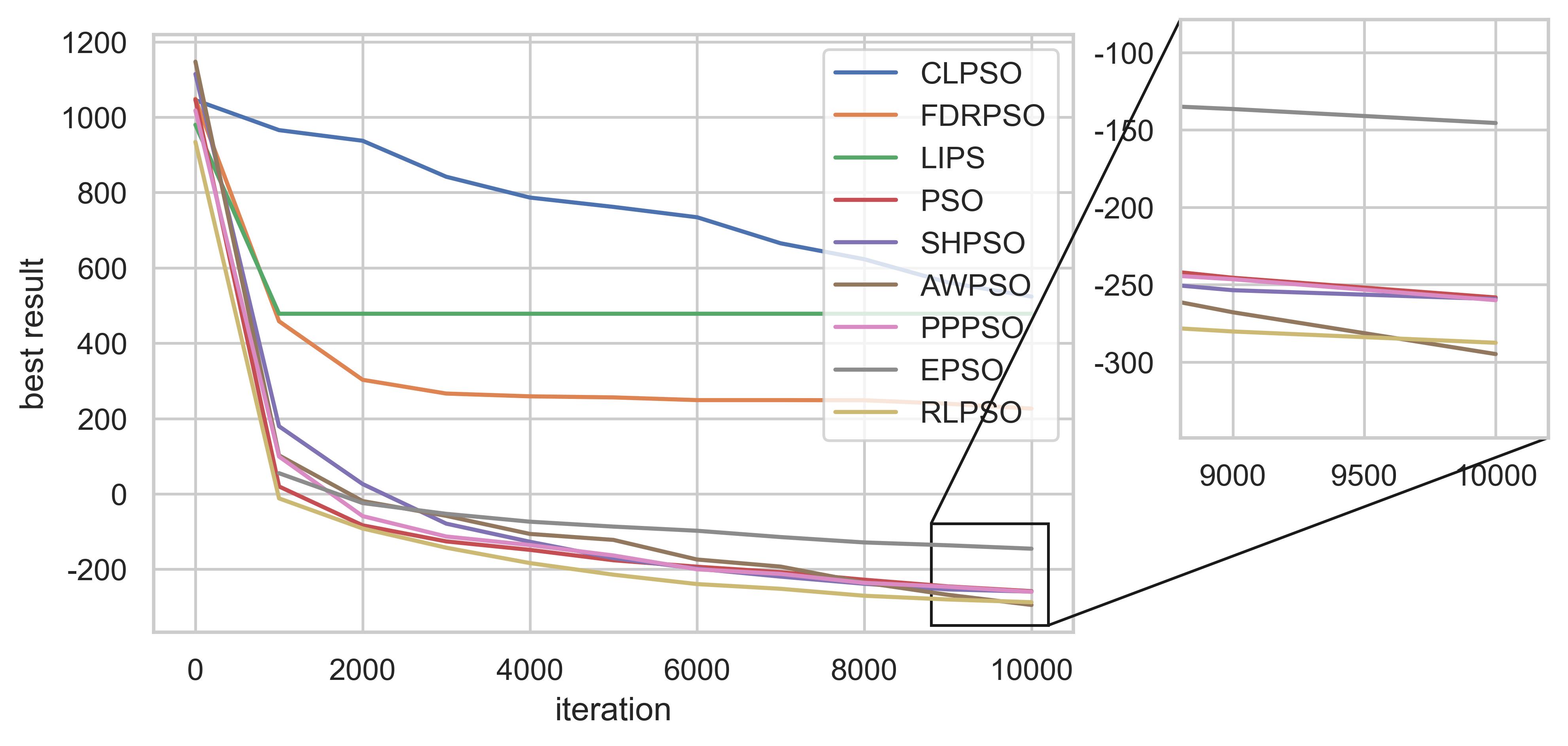}
		\end{minipage}
	}%
	\subfigure[F12]{
		\begin{minipage}[t]{0.5\linewidth}
			\centering
			\includegraphics[width=\textwidth]{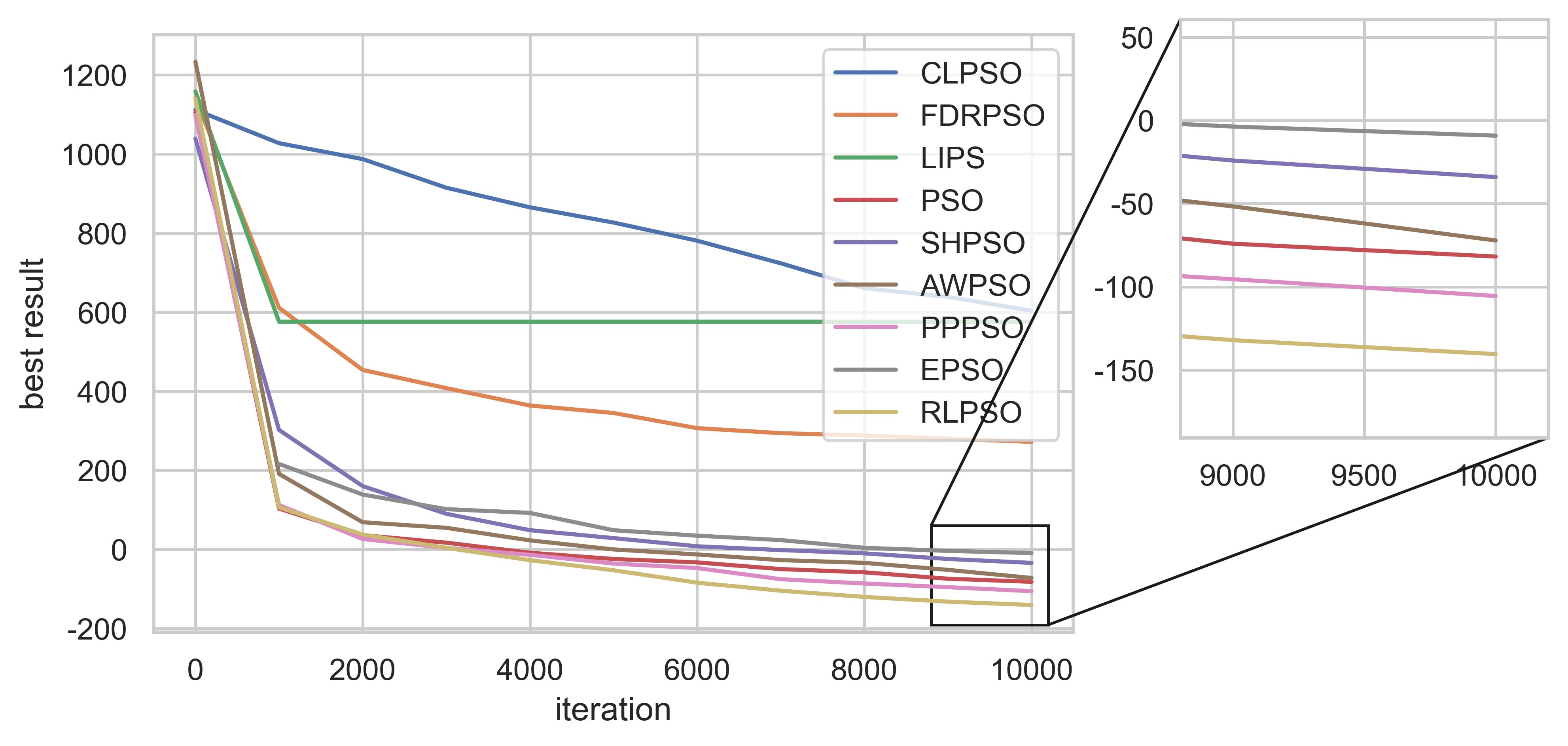}
		\end{minipage}
	}%

	\subfigure[F13]{
		\begin{minipage}[t]{0.5\linewidth}
			\centering
			\includegraphics[width=\textwidth]{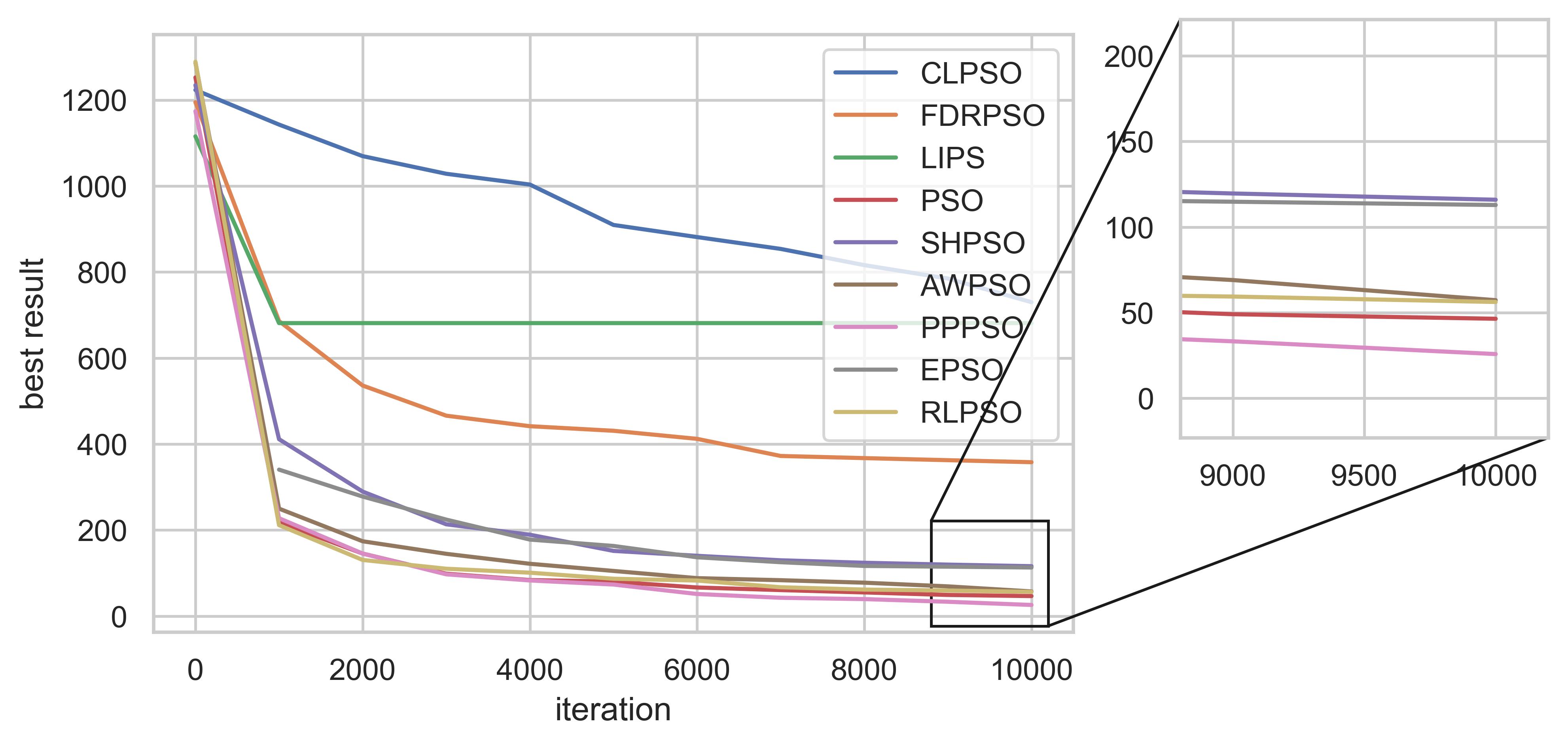}
		\end{minipage}
	}%
	\subfigure[F14]{
		\begin{minipage}[t]{0.5\linewidth}
			\centering
			\includegraphics[width=\textwidth]{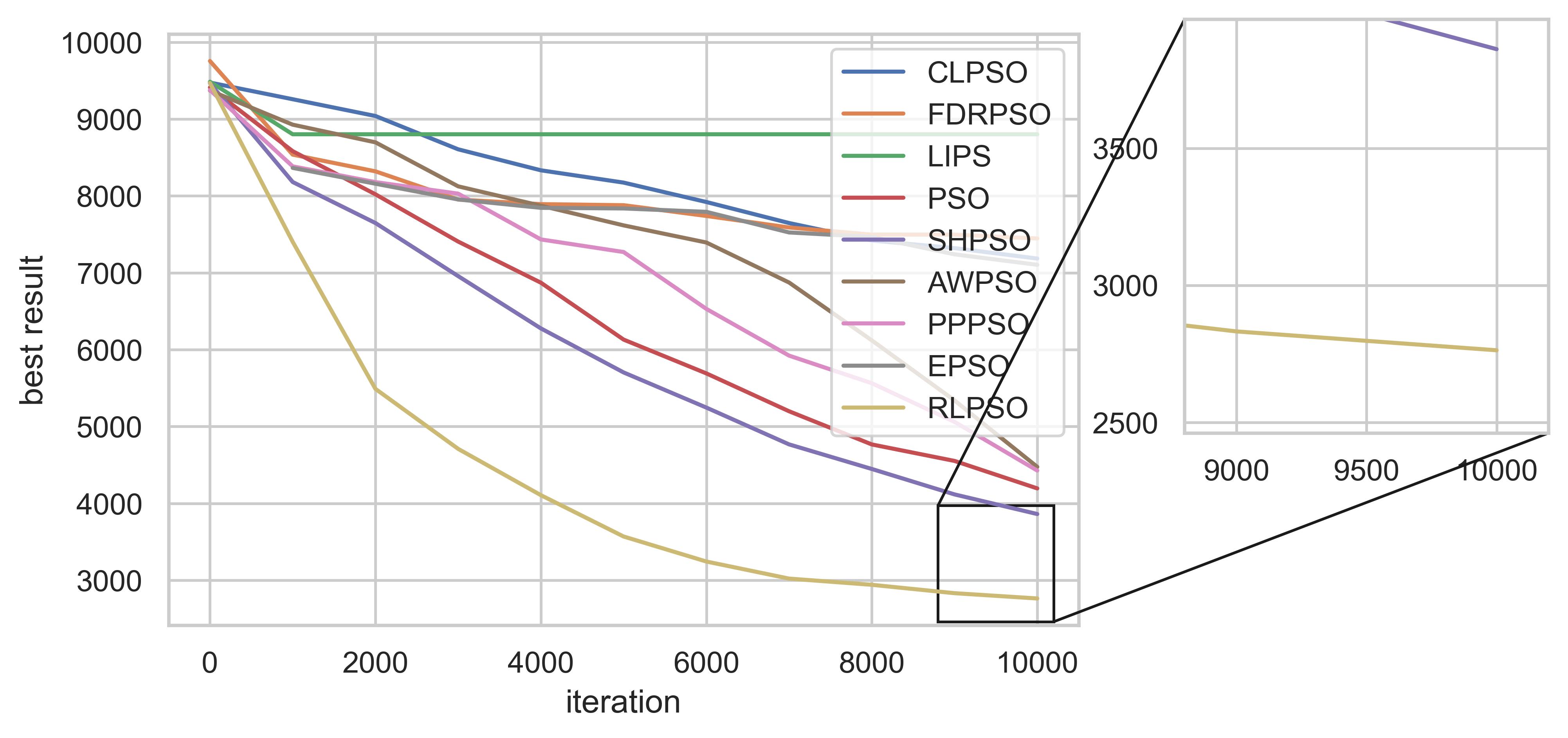}
		\end{minipage}
	}%

	\subfigure[F15]{
		\begin{minipage}[t]{0.5\linewidth}
			\centering
			\includegraphics[width=\textwidth]{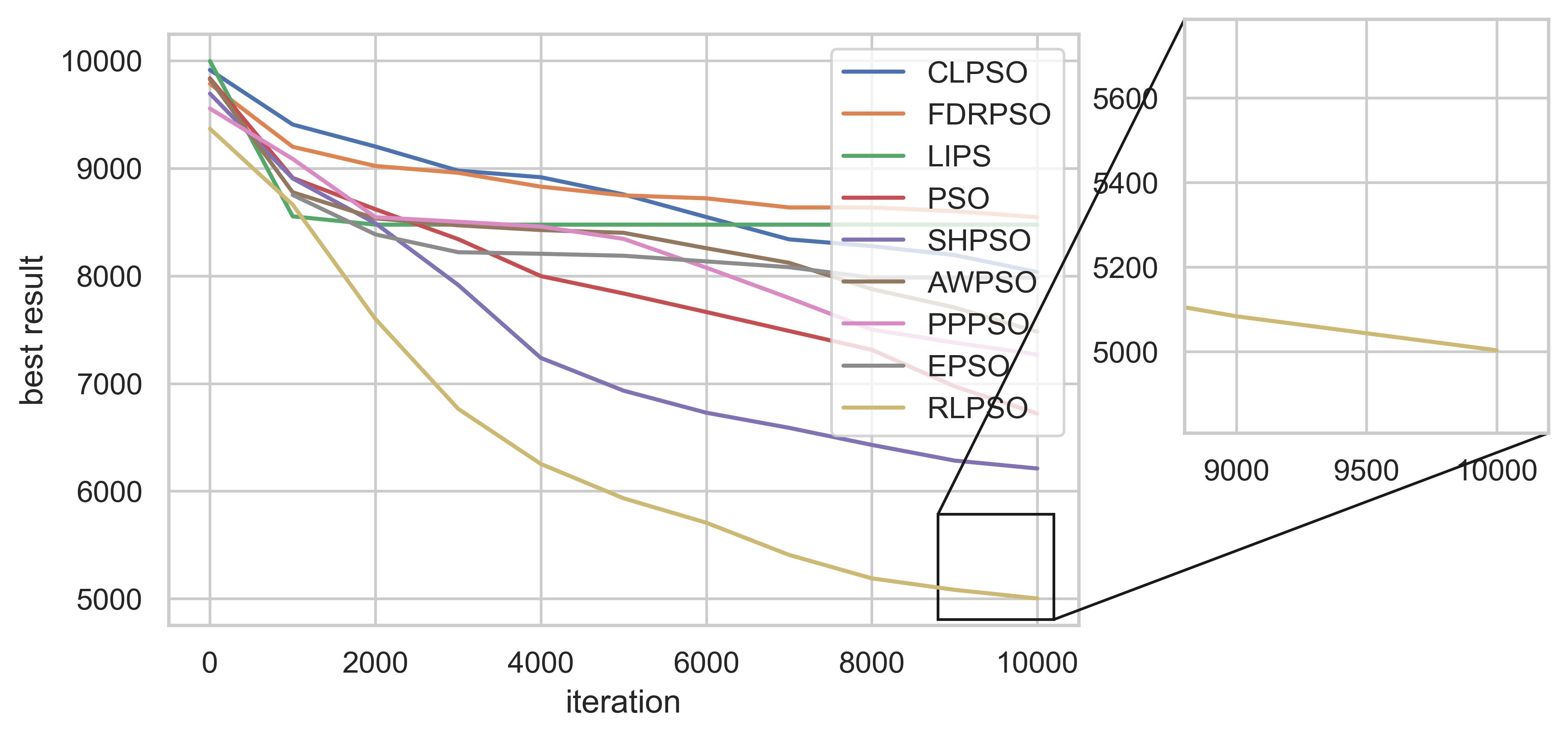}
		\end{minipage}
	}%
	\subfigure[F16]{
		\begin{minipage}[t]{0.5\linewidth}
			\centering
			\includegraphics[width=\textwidth]{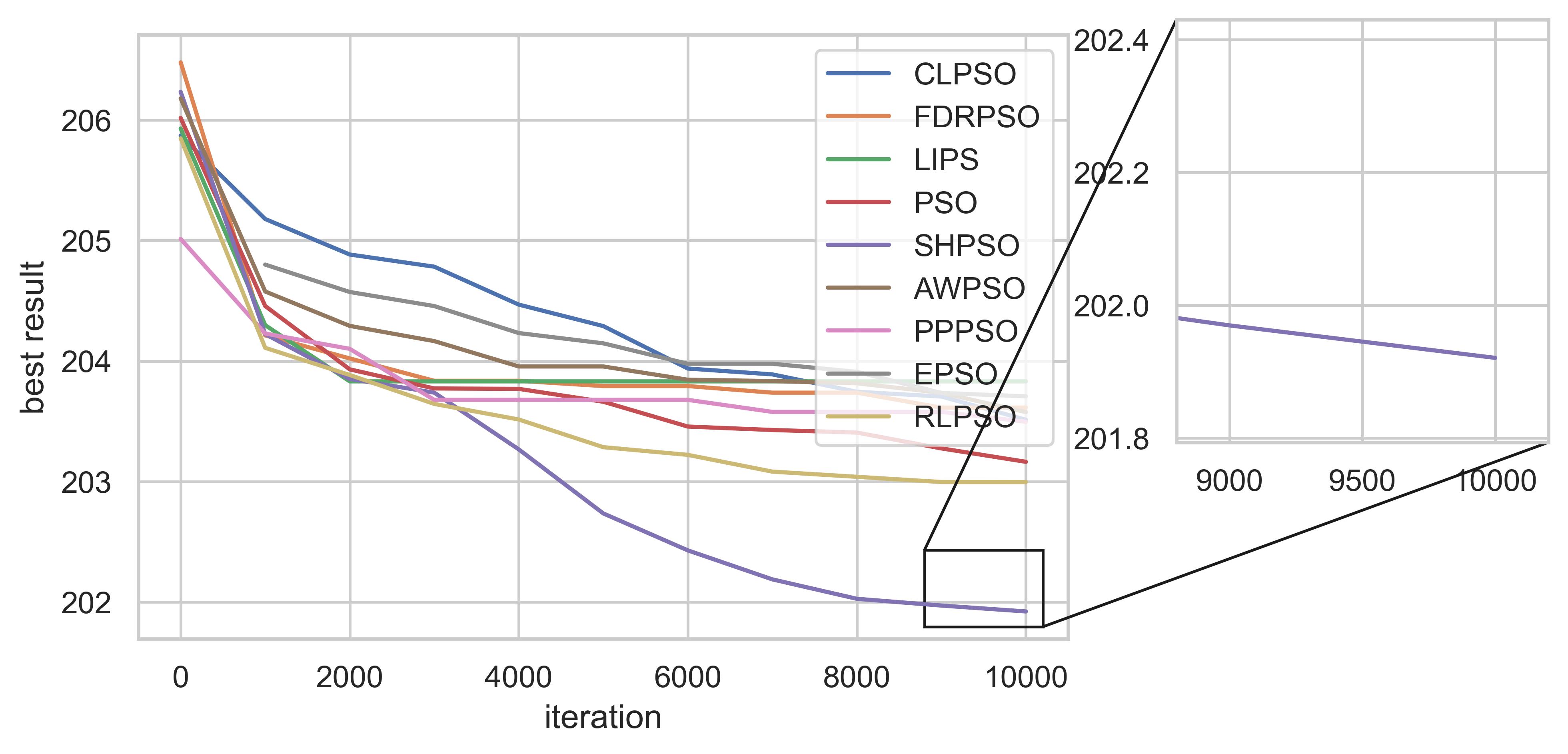}
		\end{minipage}
	}%

	\centering
	\caption{continued}
	\label{RLPSO-RESULT2}
\end{figure*}

\begin{figure*}
	\ContinuedFloat
	\subfigure[F17]{
		\begin{minipage}[t]{0.5\linewidth}
			\centering
			\includegraphics[width=\textwidth]{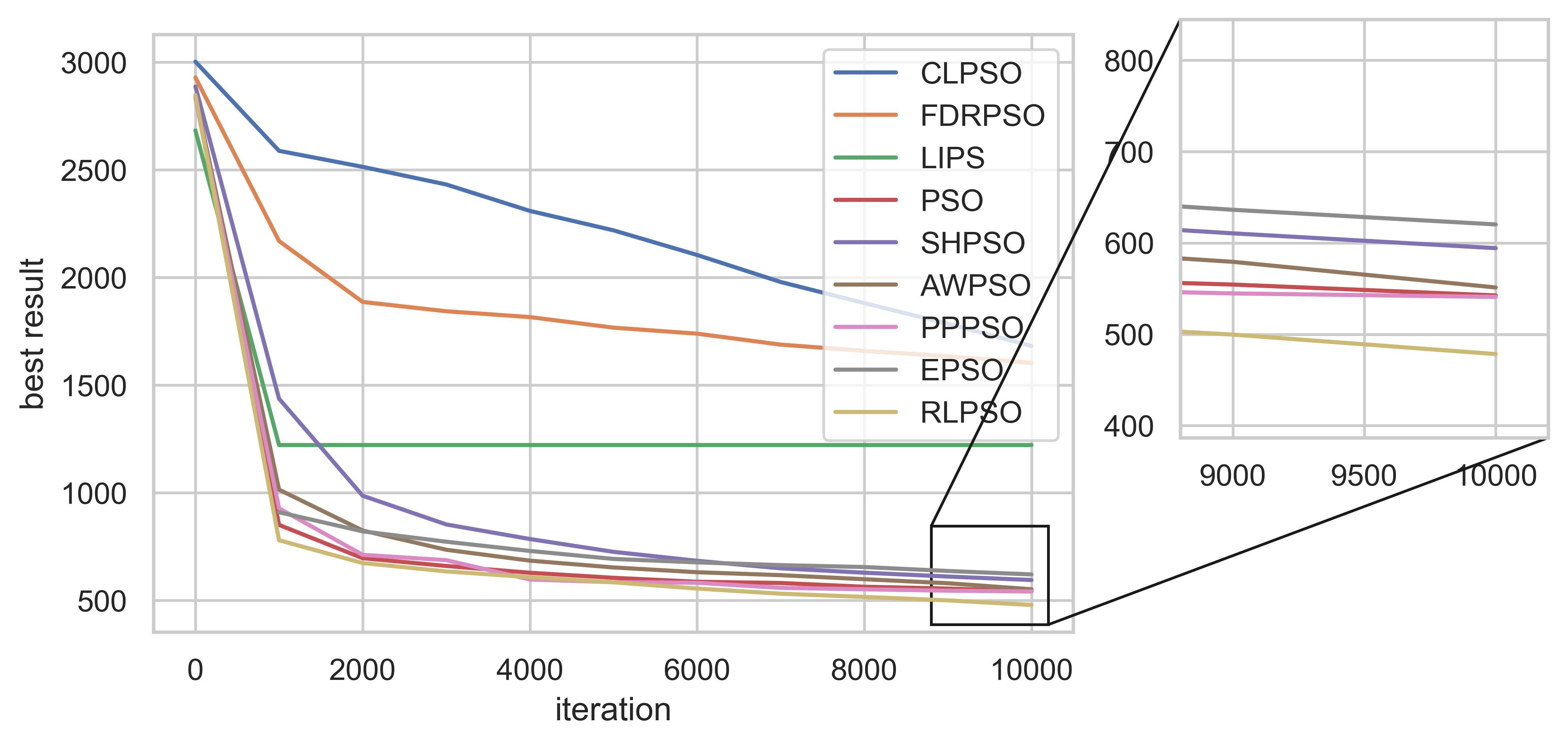}
		\end{minipage}
	}%
	\subfigure[F18]{
		\begin{minipage}[t]{0.5\linewidth}
			\centering
			\includegraphics[width=\textwidth]{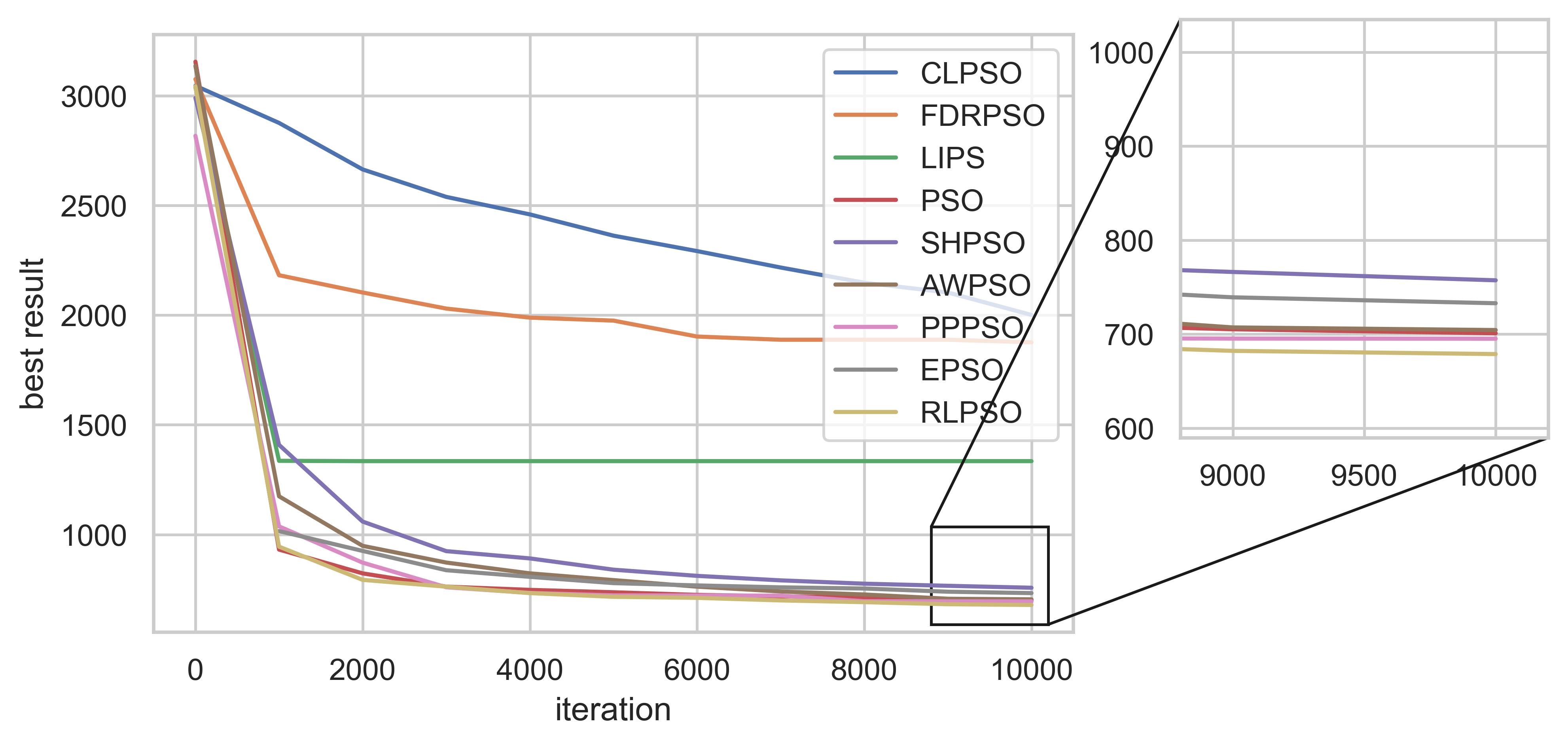}
		\end{minipage}
	}%

	\subfigure[F19]{
		\begin{minipage}[t]{0.5\linewidth}
			\centering
			\includegraphics[width=\textwidth]{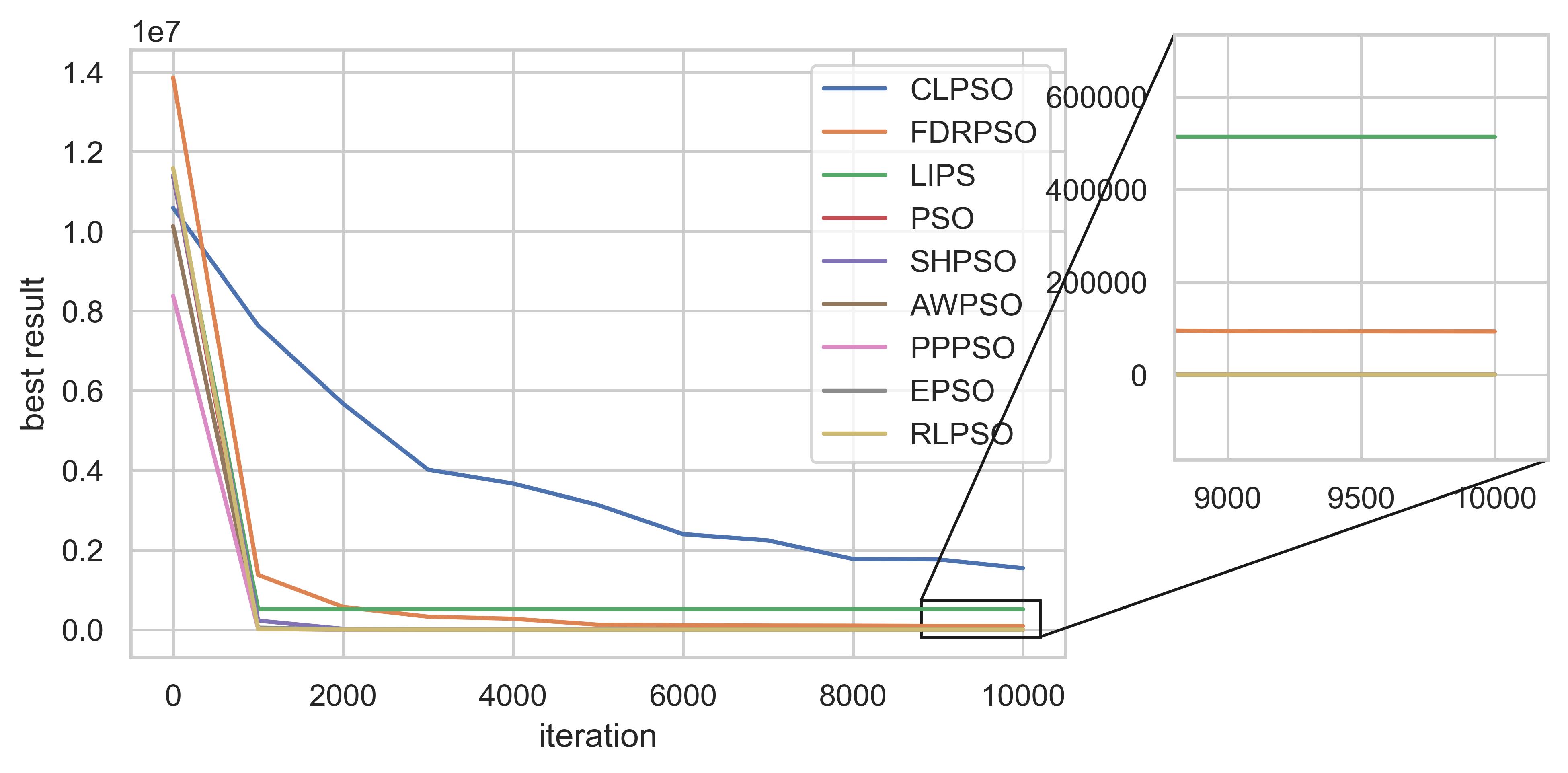}
		\end{minipage}
	}%
	\subfigure[F20]{
		\begin{minipage}[t]{0.5\linewidth}
			\centering
			\includegraphics[width=\textwidth]{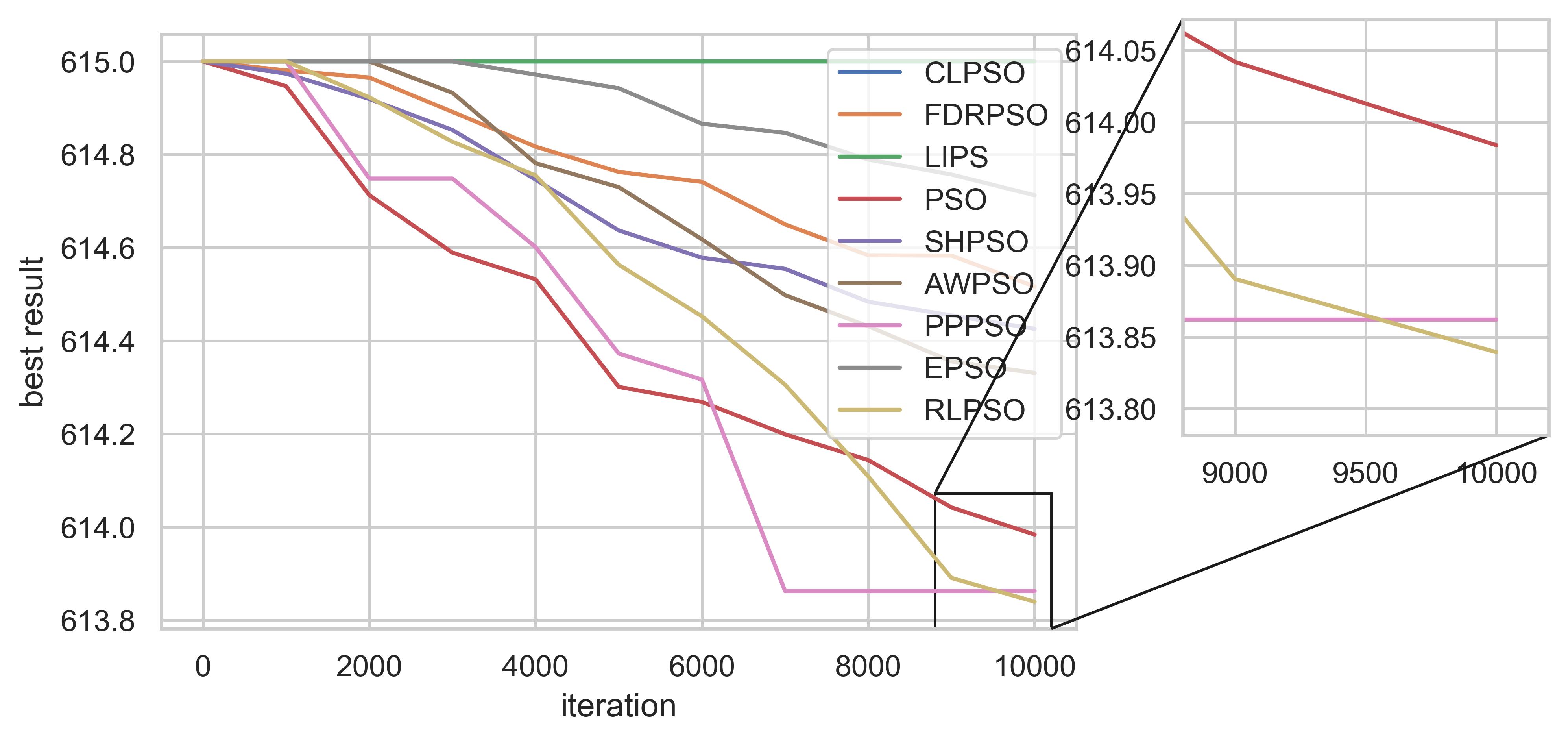}
		\end{minipage}
	}%

	\subfigure[F21]{
		\begin{minipage}[t]{0.5\linewidth}
			\centering
			\includegraphics[width=\textwidth]{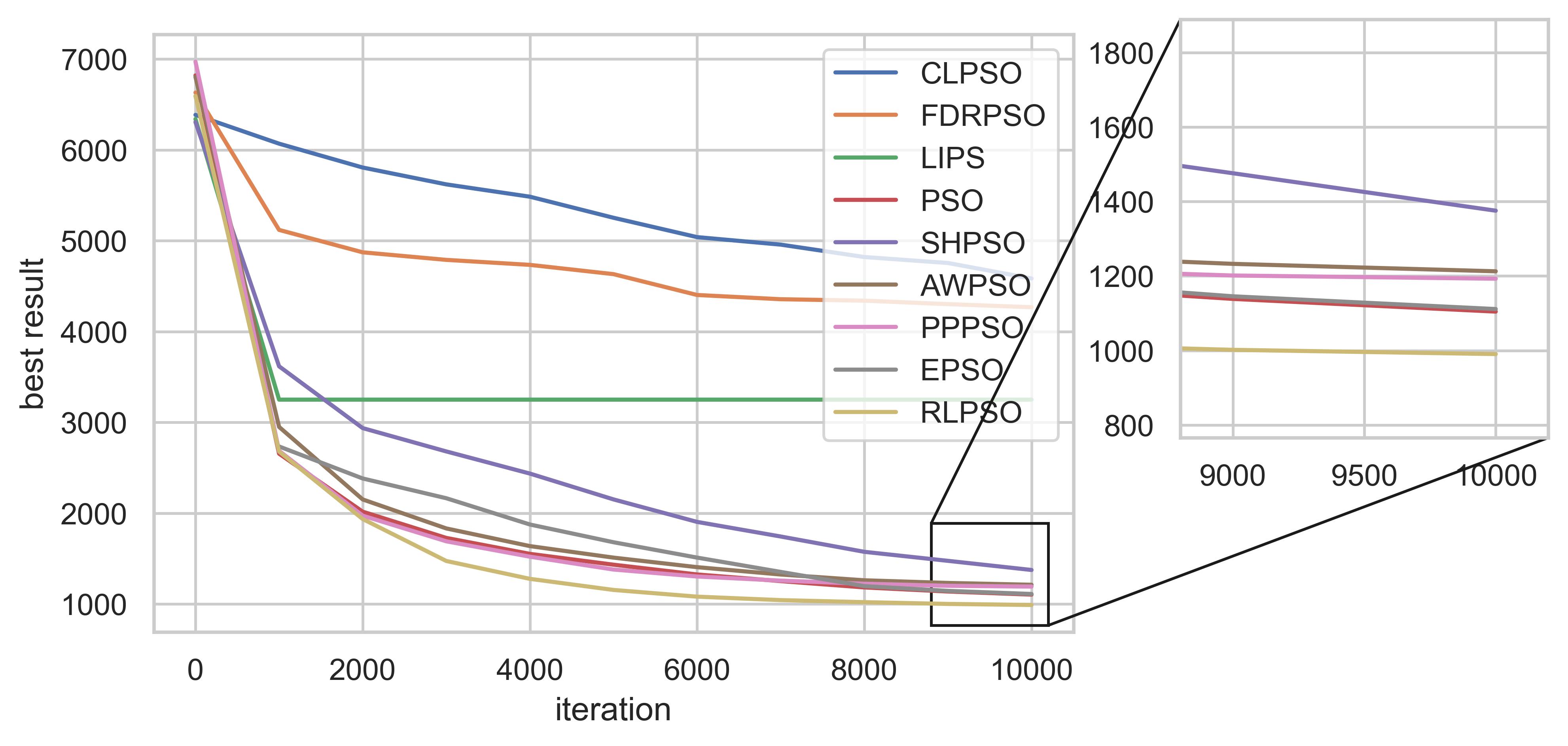}
		\end{minipage}
	}%
	\subfigure[F22]{
		\begin{minipage}[t]{0.5\linewidth}
			\centering
			\includegraphics[width=\textwidth]{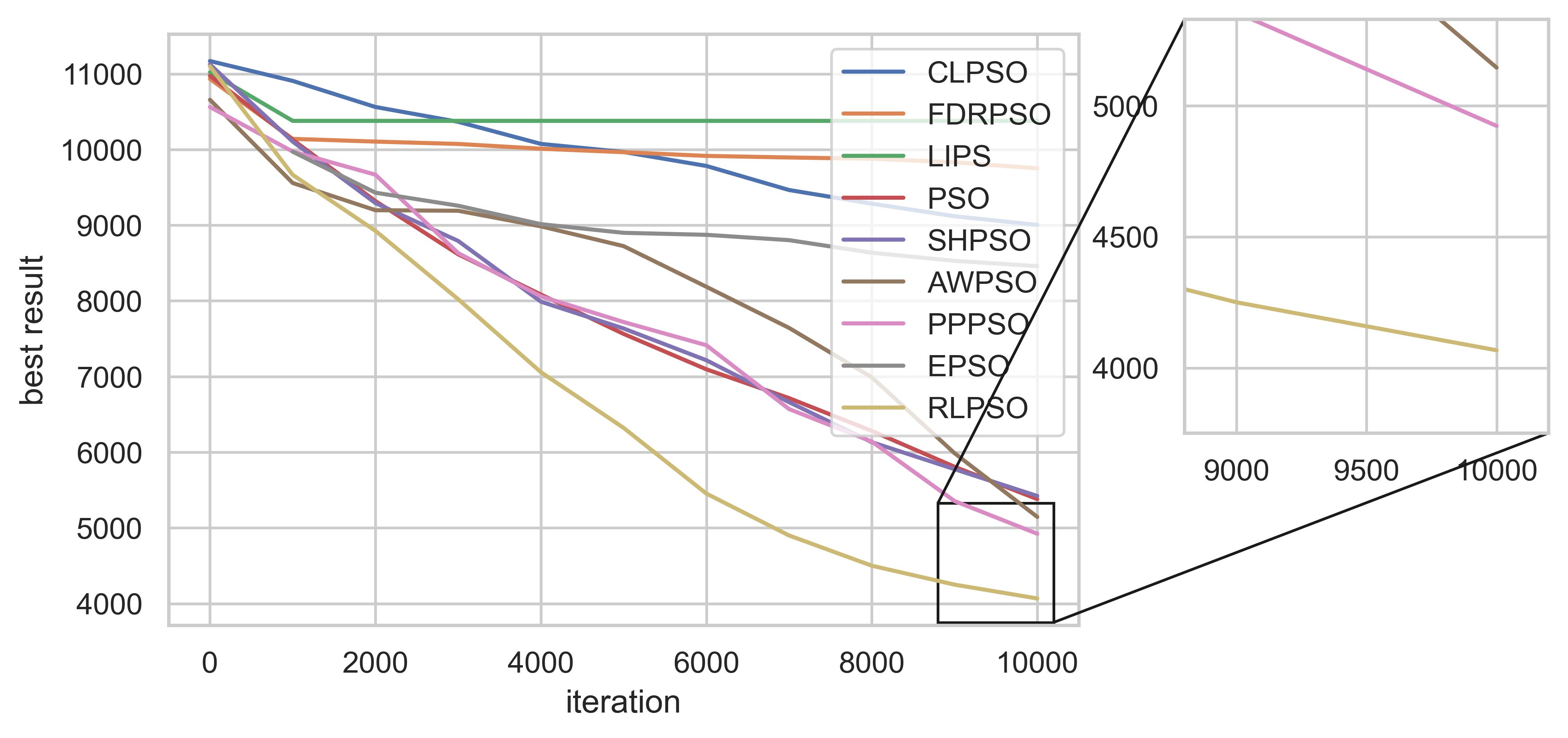}
		\end{minipage}
	}%
	
	\subfigure[F23]{
		\begin{minipage}[t]{0.5\linewidth}
			\centering
			\includegraphics[width=\textwidth]{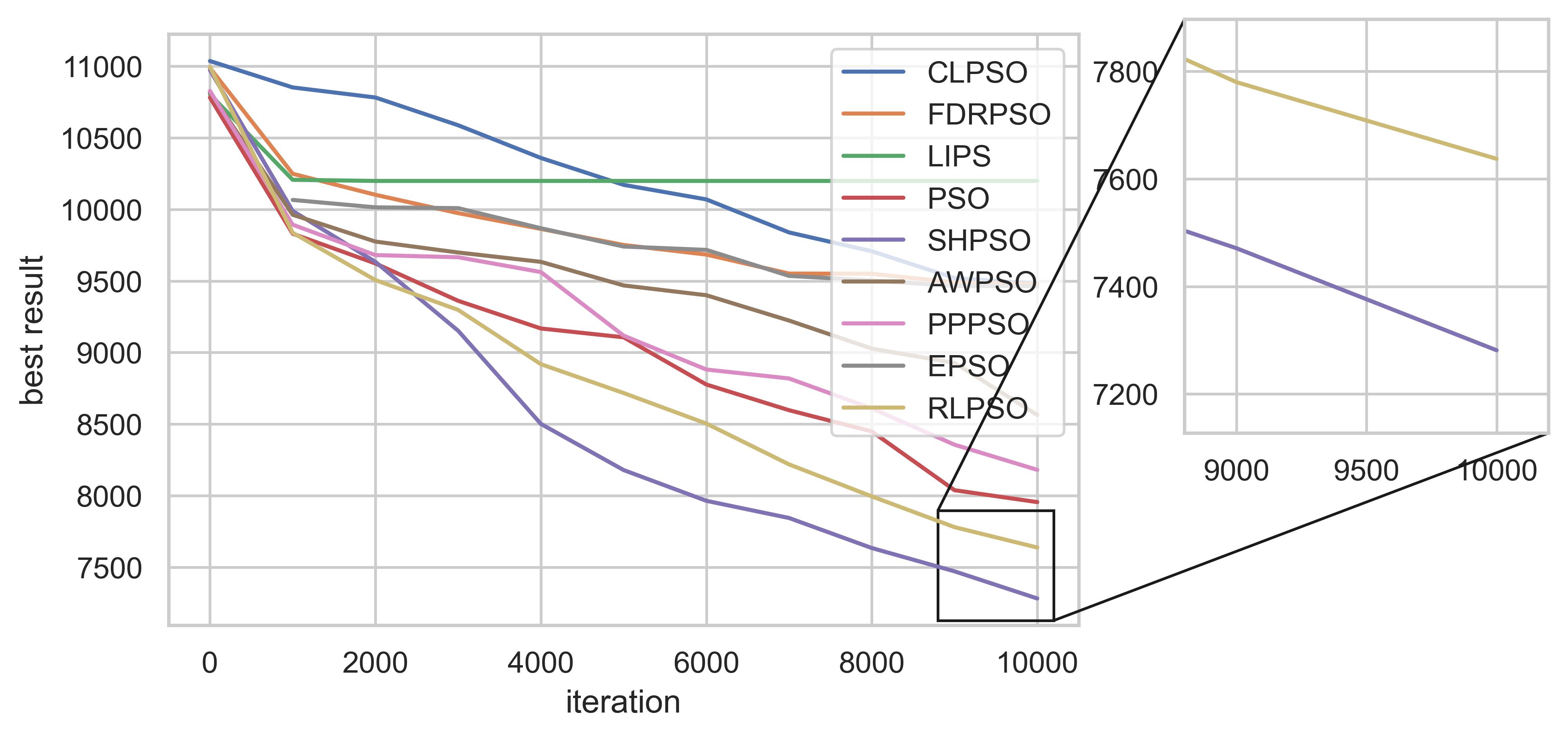}
		\end{minipage}
	}%
	\subfigure[F24]{
		\begin{minipage}[t]{0.5\linewidth}
			\centering
			\includegraphics[width=\textwidth]{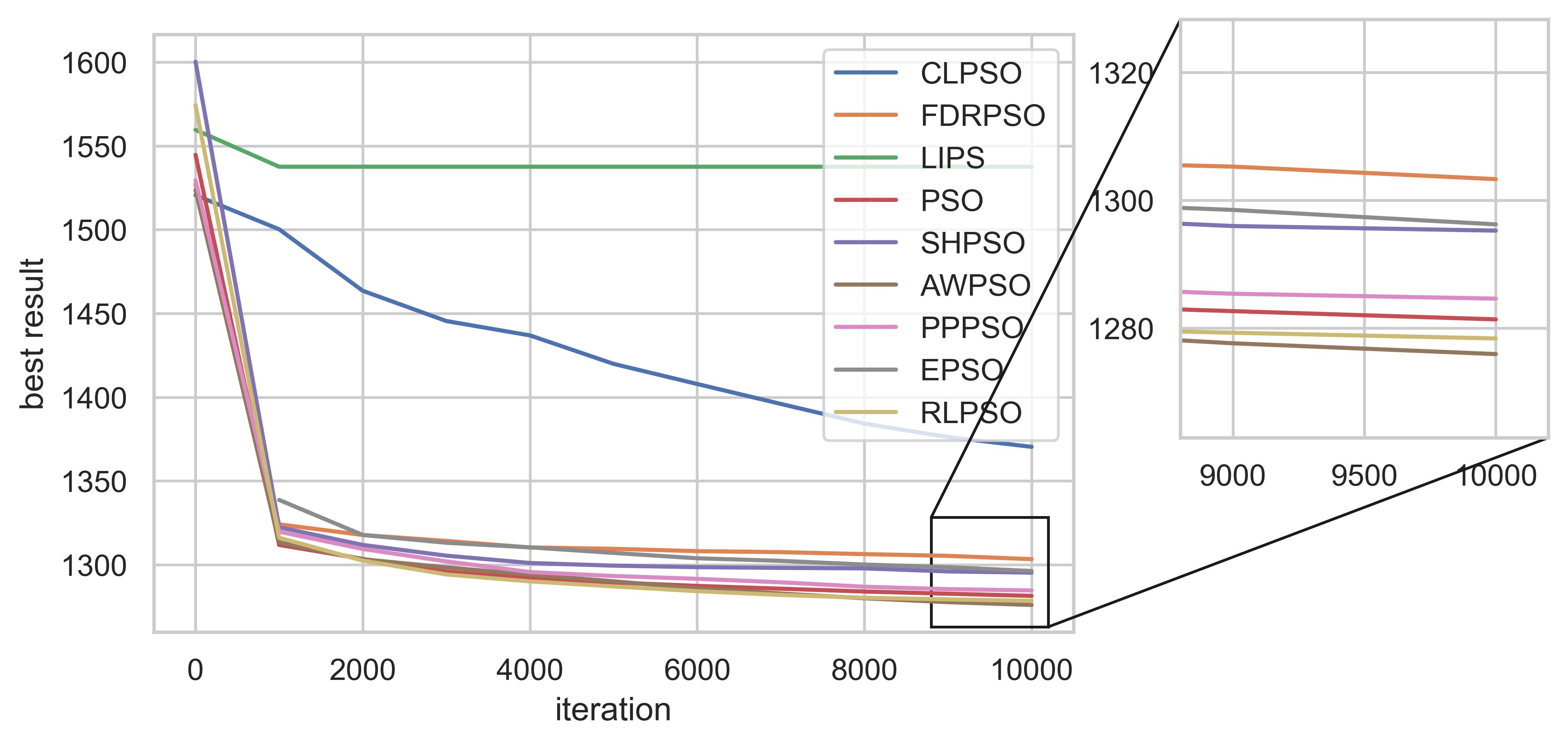}
		\end{minipage}
	}%

	\centering
	\caption{continued}
	\label{RLPSO-RESULT3}
\end{figure*}

\begin{figure*}
	\ContinuedFloat
	\subfigure[F25]{
		\begin{minipage}[t]{0.5\linewidth}
			\centering
			\includegraphics[width=\textwidth]{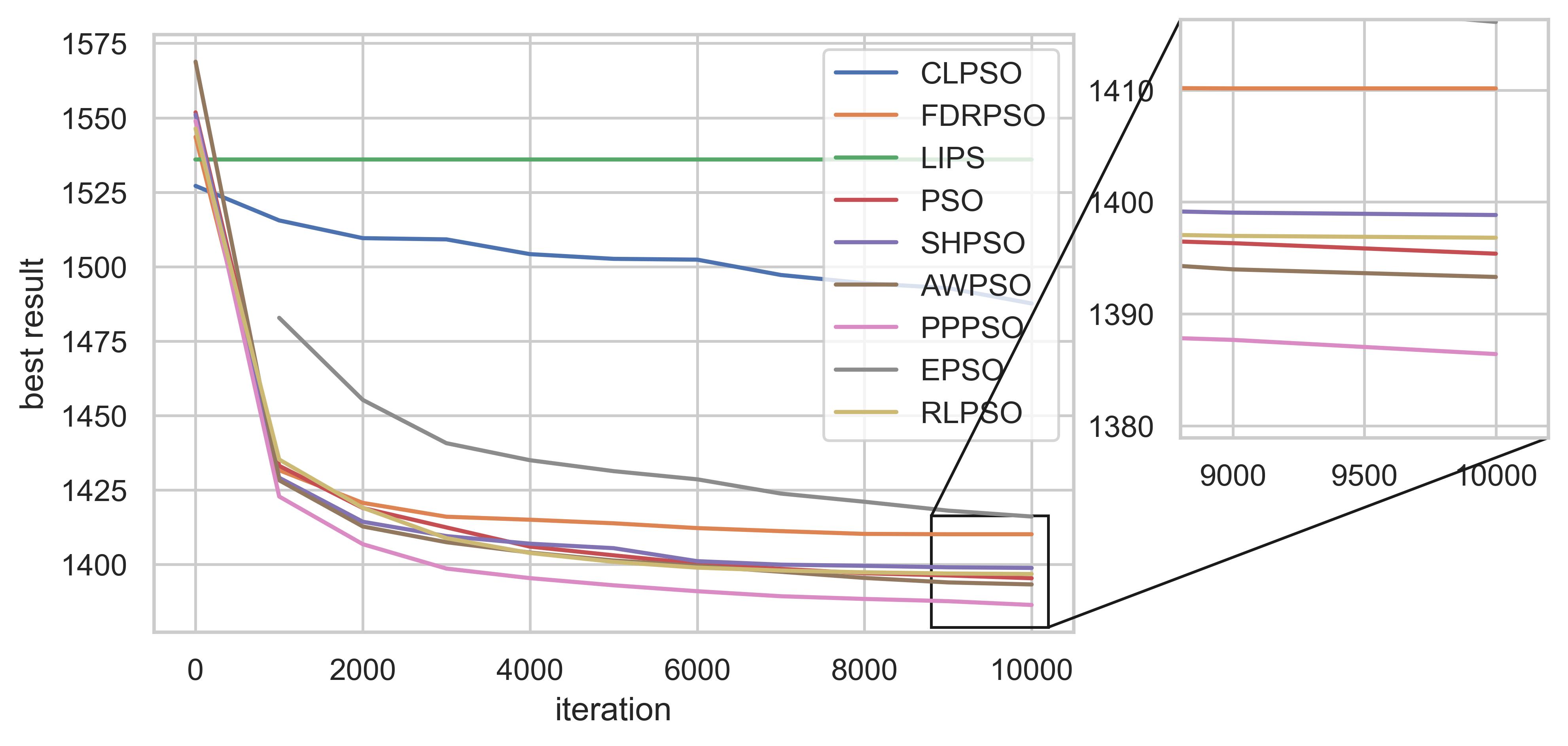}
		\end{minipage}
	}%
	\subfigure[F26]{
		\begin{minipage}[t]{0.5\linewidth}
			\centering
			\includegraphics[width=\textwidth]{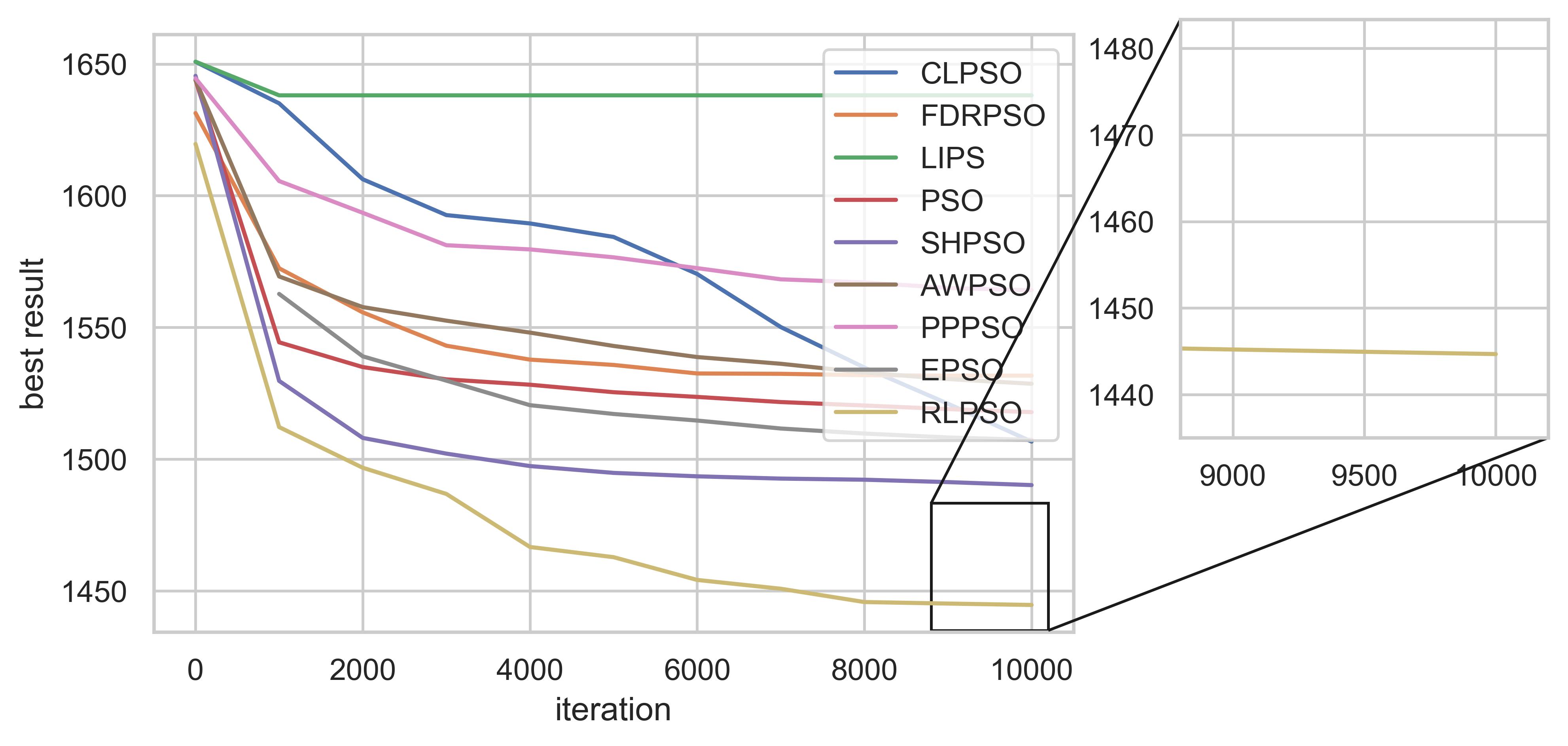}
		\end{minipage}
	}%
	
	\subfigure[F27]{
		\begin{minipage}[t]{0.5\linewidth}
			\centering
			\includegraphics[width=\textwidth]{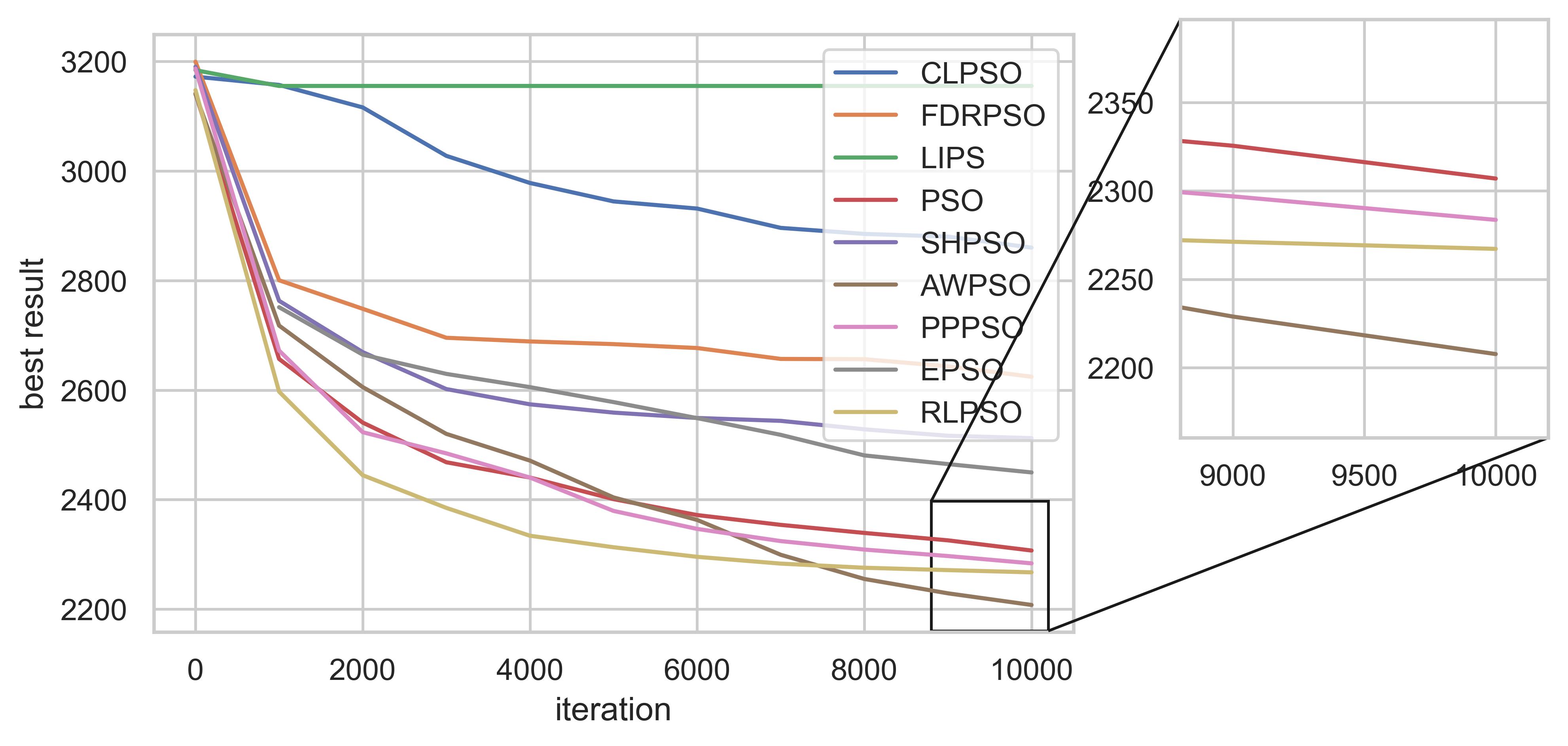}
		\end{minipage}
	}%
	\subfigure[F28]{
		\begin{minipage}[t]{0.5\linewidth}
			\centering
			\includegraphics[width=\textwidth]{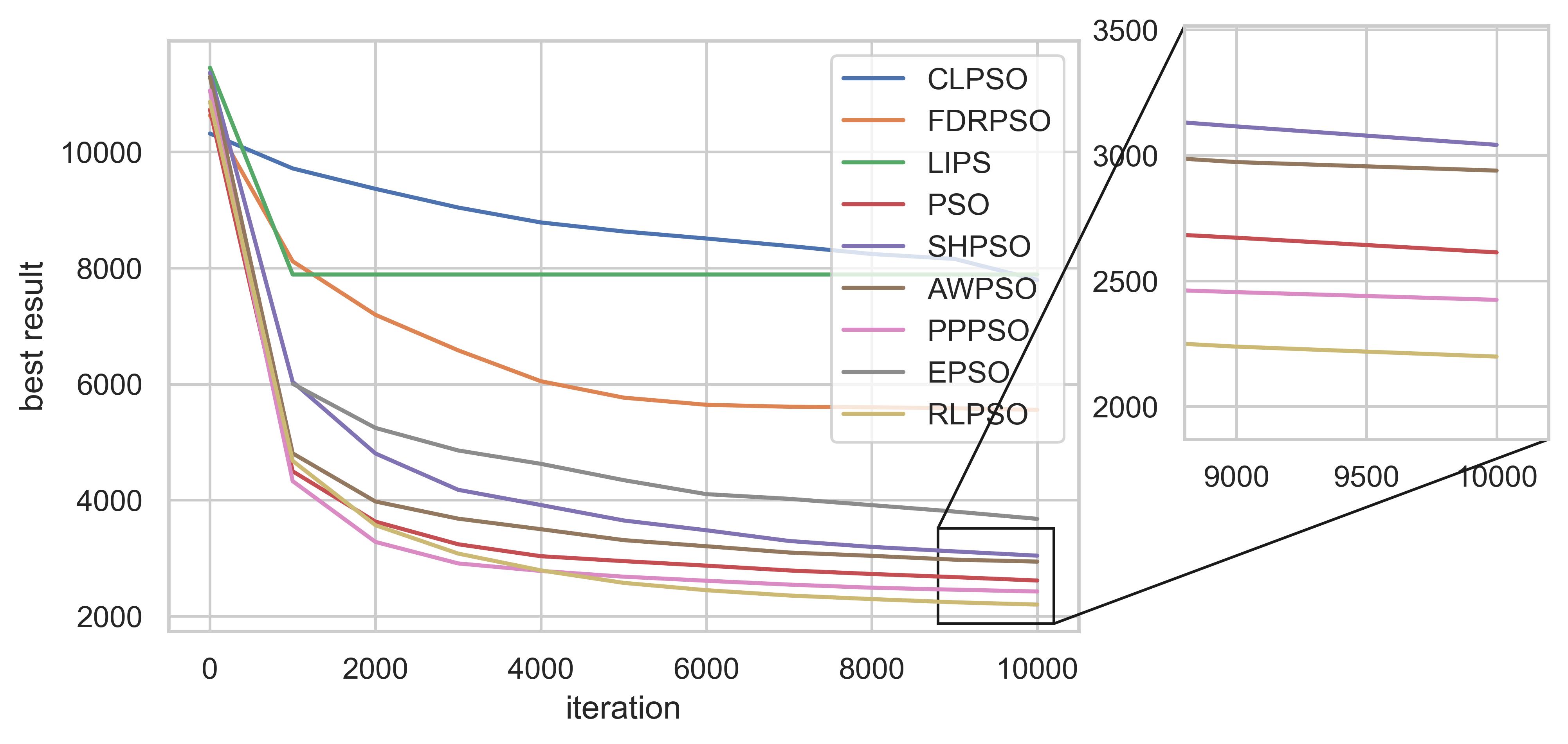}
		\end{minipage}
	}%
	
	\centering
	\caption{continued}
	\label{RLPSO-RESULT4}
\end{figure*}

\section{CONCLUTION}\label{sec:CONCLUTION}

% introduce
In this paper, a reinforcement learning-based parameter adaption method RLAM and an RLAM-based RLPSO are proposed.
In RLAM, Through each generation, the particle running selecting the optimal coefficients returned by actor network.
Also, the actor network will be trained before the running. It can be trained using the target function or just using some test function.
A combination of iteration, no-improving-iteration and diversity is seen as the calculation of the state.
Reward is calculated according to the change of best result after an update. 
In RLPSO, in addition to RLAM, CLPSO and mutation mechanisms are also added to the algorithm.
Furthermore, this paper carries out comprehensive experiments to compare the new algorithm with other online adaption method, investigate the effects of RLAM with different PSO variants and compare RLPSO with other state-of-the-art PSO variants.

The proposed method is incorporated into multiple PSO variants and tested on the CEC2013 test set, and it can be seen that almost all PSO variants have improved final optimization accuracy. This result shows that RLAM is beneficial and harmless in almost all problems and all optimization algorithms. And it solves the problem that manual parameter adjustment is too cumbersome and burdensome.
The algorithm proposed in this paper has also been compared with other online adaption methods, including adaptive particle swarm optimization based on reinforcement learning, adaptive particle swarm optimization based on fuzzy logic, and adaptive particle swarm optimization based on success rate history. In the final results we see that the adaptive algorithm proposed in this paper significantly outperforms other adaptive algorithms.
Since most of the current state-of-the-art particle swarm algorithms combine many other methods, adjusting parameters alone cannot make a good comparison. Therefore, this paper designs RLPSO based on RLAM, which combines CLPSO and some mutation and grouping operations. The final algorithm is compared with a variety of top particle swarm algorithms on the CEC2013 test set, and it can be seen from the results that the proposed algorithm is at the leading level.

% future

In the future, based on RLAM, the selection of states, the control of actions and the applicability of other optimization algorithms will be further studied to further improve the optimization performance.



\bibliographystyle{elsarticle-num} 
\bibliography{cite}

\begin{thebibliography}{10}
\expandafter\ifx\csname url\endcsname\relax
  \def\url#1{\texttt{#1}}\fi
\expandafter\ifx\csname urlprefix\endcsname\relax\def\urlprefix{URL }\fi
\expandafter\ifx\csname href\endcsname\relax
  \def\href#1#2{#2} \def\path#1{#1}\fi

\bibitem{DELSER2019220}
J.~{Del Ser}, E.~Osaba, D.~Molina, X.-S. Yang, S.~Salcedo-Sanz, D.~Camacho,
  S.~Das, P.~N. Suganthan, C.~A. {Coello Coello}, F.~Herrera,
  \href{https://www.sciencedirect.com/science/article/pii/S2210650218310277}{Bio-inspired
  computation: Where we stand and what's next}, Swarm and Evolutionary
  Computation 48 (2019) 220--250.
\newblock \href {https://doi.org/https://doi.org/10.1016/j.swevo.2019.04.008}
  {\path{doi:https://doi.org/10.1016/j.swevo.2019.04.008}}.
\newline\urlprefix\url{https://www.sciencedirect.com/science/article/pii/S2210650218310277}

\bibitem{XU2019100562}
Y.~Xu, D.~Pi,
  \href{https://www.sciencedirect.com/science/article/pii/S221065021831006X}{A
  hybrid enhanced bat algorithm for the generalized redundancy allocation
  problem}, Swarm and Evolutionary Computation 50 (2019) 100562.
\newblock \href {https://doi.org/https://doi.org/10.1016/j.swevo.2019.100562}
  {\path{doi:https://doi.org/10.1016/j.swevo.2019.100562}}.
\newline\urlprefix\url{https://www.sciencedirect.com/science/article/pii/S221065021831006X}

\bibitem{6031913}
Z.~Zhu, J.~Zhou, Z.~Ji, Y.-H. Shi, Dna sequence compression using adaptive
  particle swarm optimization-based memetic algorithm, IEEE Transactions on
  Evolutionary Computation 15~(5) (2011) 643--658.
\newblock \href {https://doi.org/10.1109/TEVC.2011.2160399}
  {\path{doi:10.1109/TEVC.2011.2160399}}.

\bibitem{arqub2014numerical}
O.~A. Arqub, Z.~Abo-Hammour, Numerical solution of systems of second-order
  boundary value problems using continuous genetic algorithm, Information
  sciences 279 (2014) 396--415.

\bibitem{abu2017adaptation}
O.~Abu~Arqub, Adaptation of reproducing kernel algorithm for solving fuzzy
  fredholm--volterra integrodifferential equations, Neural Computing and
  Applications 28~(7) (2017) 1591--1610.

\bibitem{abu2018solutions}
O.~Abu~Arqub, B.~Maayah, Solutions of bagley--torvik and painlev{\'e} equations
  of fractional order using iterative reproducing kernel algorithm with error
  estimates, Neural Computing and Applications 29~(5) (2018) 1465--1479.

\bibitem{zhang2016object}
H.~Zhang, X.~Cao, J.~K. Ho, T.~W. Chow, Object-level video advertising: an
  optimization framework, IEEE Transactions on industrial informatics 13~(2)
  (2016) 520--531.

\bibitem{milner2012nature}
S.~Milner, C.~Davis, H.~Zhang, J.~Llorca, Nature-inspired self-organization,
  control, and optimization in heterogeneous wireless networks, IEEE
  transactions on mobile computing 11~(7) (2012) 1207--1222.

\bibitem{kennedy1995particle}
J.~Kennedy, R.~Eberhart, Particle swarm optimization, in: Proceedings of
  ICNN'95-international conference on neural networks, Vol.~4, IEEE, 1995, pp.
  1942--1948.

\bibitem{Xu2020}
Y.~Xu, D.~Pi, \href{https://doi.org/10.1007/s00521-019-04527-9}{{A
  reinforcement learning-based communication topology in particle swarm
  optimization}}, Neural Computing and Applications 32~(14) (2020)
  10007--10032.
\newblock \href {https://doi.org/10.1007/s00521-019-04527-9}
  {\path{doi:10.1007/s00521-019-04527-9}}.
\newline\urlprefix\url{https://doi.org/10.1007/s00521-019-04527-9}

\bibitem{Liu2019}
Y.~Liu, H.~Lu, S.~Cheng, Y.~Shi, {An Adaptive Online Parameter Control
  Algorithm for Particle Swarm Optimization Based on Reinforcement Learning},
  2019 IEEE Congress on Evolutionary Computation, CEC 2019 - Proceedings (2019)
  815--822\href {https://doi.org/10.1109/CEC.2019.8790035}
  {\path{doi:10.1109/CEC.2019.8790035}}.

\bibitem{liu2021swin}
Z.~Liu, Y.~Lin, Y.~Cao, H.~Hu, Y.~Wei, Z.~Zhang, S.~Lin, B.~Guo, Swin
  transformer: Hierarchical vision transformer using shifted windows, in:
  Proceedings of the IEEE/CVF International Conference on Computer Vision,
  2021, pp. 10012--10022.

\bibitem{Devlin2019}
J.~Devlin, M.~W. Chang, K.~Lee, K.~Toutanova, {BERT: Pre-training of deep
  bidirectional transformers for language understanding}, NAACL HLT 2019 - 2019
  Conference of the North American Chapter of the Association for Computational
  Linguistics: Human Language Technologies - Proceedings of the Conference
  1~(Mlm) (2019) 4171--4186.
\newblock \href {http://arxiv.org/abs/1810.04805} {\path{arXiv:1810.04805}}.

\bibitem{Lillicrap2016}
T.~P. Lillicrap, J.~J. Hunt, A.~Pritzel, N.~Heess, T.~Erez, Y.~Tassa,
  D.~Silver, D.~Wierstra, {Continuous control with deep reinforcement
  learning}, 4th International Conference on Learning Representations, ICLR
  2016 - Conference Track Proceedings (2016).
\newblock \href {http://arxiv.org/abs/1509.02971} {\path{arXiv:1509.02971}}.

\bibitem{bartz2007experimental}
T.~Bartz-Beielstein, M.~Preuss, Experimental research in evolutionary
  computation, in: Proceedings of the 9th annual conference companion on
  genetic and evolutionary computation, 2007, pp. 3001--3020.

\bibitem{birattari2009tuning}
M.~Birattari, J.~Kacprzyk, Tuning metaheuristics: a machine learning
  perspective, Vol. 197, Springer, 2009.

\bibitem{hoos2011automated}
H.~H. Hoos, Automated algorithm configuration and parameter tuning, in:
  Autonomous search, Springer, 2011, pp. 37--71.

\bibitem{eberhart2000comparing}
R.~C. Eberhart, Y.~Shi, Comparing inertia weights and constriction factors in
  particle swarm optimization, in: Proceedings of the 2000 congress on
  evolutionary computation. CEC00 (Cat. No. 00TH8512), Vol.~1, IEEE, 2000, pp.
  84--88.

\bibitem{Ratnaweera2004}
A.~Ratnaweera, S.~K. Halgamuge, H.~C. Watson, {Self-organizing hierarchical
  particle swarm optimizer with time-varying acceleration coefficients}, IEEE
  Transactions on Evolutionary Computation 8~(3) (2004) 240--255.
\newblock \href {https://doi.org/10.1109/TEVC.2004.826071}
  {\path{doi:10.1109/TEVC.2004.826071}}.

\bibitem{Zheng20031802}
Y.-L. Zheng, L.-H. Ma, L.-Y. Zhang, J.-X. Qian, On the convergence analysis and
  parameter selection in particle swarm optimization, Vol.~3, 2003, pp.
  1802--1807, cited By 218.

\bibitem{Cui200789}
H.-M. Cui, Q.-B. Zhu, Convergence analysis and parameter selection in particle
  swarm optimization, Computer Engineering and Applications 43~(23) (2007)
  89--91, cited By 25.

\bibitem{Yu20051286}
H.-J. Yu, L.-P. Zhang, D.-Z. Chen, S.-X. Hu,
  \href{https://www.scopus.com/inward/record.uri?eid=2-s2.0-27744459346\&partnerID=40\&md5=66be80f05b33b02c32efd51c64b806a6}{Adaptive
  particle swarm optimization algorithm based on feedback mechanism}, Zhejiang
  Daxue Xuebao (Gongxue Ban)/Journal of Zhejiang University (Engineering
  Science) 39~(9) (2005) 1286--1291, cited By 24.
\newline\urlprefix\url{https://www.scopus.com/inward/record.uri?eid=2-s2.0-27744459346\&partnerID=40\&md5=66be80f05b33b02c32efd51c64b806a6}

\bibitem{chen2006natural}
G.~Chen, X.~Huang, J.~Jia, Z.~Min, Natural exponential inertia weight strategy
  in particle swarm optimization, in: 2006 6th world congress on intelligent
  control and automation, Vol.~1, IEEE, 2006, pp. 3672--3675.

\bibitem{guimin2006study}
C.~Guimin, J.~Jianyuan, H.~Qi, Study on the strategy of decreasing inertia
  weight in particle swarm optimization algorithm, Journal of xi'an jiaotong
  university 40~(1) (2006) 53--56.

\bibitem{malik2007new}
R.~F. Malik, T.~A. Rahman, S.~Z.~M. Hashim, R.~Ngah, New particle swarm
  optimizer with sigmoid increasing inertia weight, International Journal of
  Computer Science and Security 1~(2) (2007) 35--44.

\bibitem{feng2007comparing}
Y.~Feng, Y.-M. Yao, A.-X. Wang, Comparing with chaotic inertia weights in
  particle swarm optimization, in: 2007 international conference on machine
  learning and cybernetics, Vol.~1, IEEE, 2007, pp. 329--333.

\bibitem{chen2018chaotic}
K.~Chen, F.~Zhou, A.~Liu, Chaotic dynamic weight particle swarm optimization
  for numerical function optimization, Knowledge-Based Systems 139 (2018)
  23--40.

\bibitem{eberhart2001tracking}
R.~C. Eberhart, Y.~Shi, Tracking and optimizing dynamic systems with particle
  swarms, in: Proceedings of the 2001 congress on evolutionary computation
  (IEEE Cat. No. 01TH8546), Vol.~1, IEEE, 2001, pp. 94--100.

\bibitem{tanabe2013success}
R.~Tanabe, A.~Fukunaga, Success-history based parameter adaptation for
  differential evolution, in: 2013 IEEE congress on evolutionary computation,
  IEEE, 2013, pp. 71--78.

\bibitem{Lynn2017}
N.~Lynn, P.~N. Suganthan,
  \href{http://dx.doi.org/10.1016/j.asoc.2017.02.007}{{Ensemble particle swarm
  optimizer}}, Applied Soft Computing Journal 55 (2017) 533--548.
\newblock \href {https://doi.org/10.1016/j.asoc.2017.02.007}
  {\path{doi:10.1016/j.asoc.2017.02.007}}.
\newline\urlprefix\url{http://dx.doi.org/10.1016/j.asoc.2017.02.007}

\bibitem{liu2020multipopulation}
Z.~Liu, T.~Nishi, Multipopulation ensemble particle swarm optimizer for
  engineering design problems, Mathematical Problems in Engineering 2020
  (2020).

\bibitem{tatsis2017grid}
V.~A. Tatsis, K.~E. Parsopoulos, Grid-based parameter adaptation in particle
  swarm optimization, in: 12th Metaheuristics International Conference (MIC
  2017), 2017, pp. 217--226.

\bibitem{Olivas2016}
F.~Olivas, F.~Valdez, O.~Castillo, P.~Melin,
  \href{http://dx.doi.org/10.1007/s00500-014-1567-3}{{Dynamic parameter
  adaptation in particle swarm optimization using interval type-2 fuzzy
  logic}}, Soft Computing 20~(3) (2016) 1057--1070.
\newblock \href {https://doi.org/10.1007/s00500-014-1567-3}
  {\path{doi:10.1007/s00500-014-1567-3}}.
\newline\urlprefix\url{http://dx.doi.org/10.1007/s00500-014-1567-3}

\bibitem{MELIN20133196}
P.~Melin, F.~Olivas, O.~Castillo, F.~Valdez, J.~Soria, M.~Valdez,
  \href{https://www.sciencedirect.com/science/article/pii/S0957417412012742}{Optimal
  design of fuzzy classification systems using pso with dynamic parameter
  adaptation through fuzzy logic}, Expert Systems with Applications 40~(8)
  (2013) 3196--3206.
\newblock \href {https://doi.org/https://doi.org/10.1016/j.eswa.2012.12.033}
  {\path{doi:https://doi.org/10.1016/j.eswa.2012.12.033}}.
\newline\urlprefix\url{https://www.sciencedirect.com/science/article/pii/S0957417412012742}

\bibitem{xu2020reinforcement}
Y.~Xu, D.~Pi, A reinforcement learning-based communication topology in particle
  swarm optimization, Neural Computing and Applications 32~(14) (2020)
  10007--10032.

\bibitem{liu2019adaptive}
Y.~Liu, H.~Lu, S.~Cheng, Y.~Shi, An adaptive online parameter control algorithm
  for particle swarm optimization based on reinforcement learning, in: 2019
  IEEE congress on evolutionary computation (CEC), IEEE, 2019, pp. 815--822.

\bibitem{samma2016new}
H.~Samma, C.~P. Lim, J.~M. Saleh, A new reinforcement learning-based memetic
  particle swarm optimizer, Applied Soft Computing 43 (2016) 276--297.

\bibitem{lu2021reinforcement}
L.~Lu, H.~Zheng, J.~Jie, M.~Zhang, R.~Dai, Reinforcement learning-based
  particle swarm optimization for sewage treatment control, Complex \&
  Intelligent Systems 7~(5) (2021) 2199--2210.

\bibitem{hsieh2016q}
Y.-Z. Hsieh, M.-C. Su, A q-learning-based swarm optimization algorithm for
  economic dispatch problem, Neural Computing and Applications 27~(8) (2016)
  2333--2350.

\bibitem{wu2022employing}
D.~Wu, G.~G. Wang, Employing reinforcement learning to enhance particle swarm
  optimization methods, Engineering Optimization 54~(2) (2022) 329--348.

\bibitem{WANG2018162}
F.~Wang, H.~Zhang, K.~Li, Z.~Lin, J.~Yang, X.-L. Shen,
  \href{https://www.sciencedirect.com/science/article/pii/S0020025518300380}{A
  hybrid particle swarm optimization algorithm using adaptive learning
  strategy}, Information Sciences 436-437 (2018) 162--177.
\newblock \href {https://doi.org/https://doi.org/10.1016/j.ins.2018.01.027}
  {\path{doi:https://doi.org/10.1016/j.ins.2018.01.027}}.
\newline\urlprefix\url{https://www.sciencedirect.com/science/article/pii/S0020025518300380}

\bibitem{Liang2006}
J.~J. Liang, A.~K. Qin, P.~N. Suganthan, S.~Baskar, {Comprehensive learning
  particle swarm optimizer for global optimization of multimodal functions},
  IEEE Transactions on Evolutionary Computation 10~(3) (2006) 281--295.
\newblock \href {https://doi.org/10.1109/TEVC.2005.857610}
  {\path{doi:10.1109/TEVC.2005.857610}}.

\bibitem{vaswani2017attention}
A.~Vaswani, N.~Shazeer, N.~Parmar, J.~Uszkoreit, L.~Jones, A.~N. Gomez,
  {\L}.~Kaiser, I.~Polosukhin, Attention is all you need, Advances in neural
  information processing systems 30 (2017).

\bibitem{liang2013problem}
J.~Liang, B.~Qu, P.~Suganthan, A.~G. Hern{\'a}ndez-D{\'\i}az, Problem
  definitions and evaluation criteria for the cec 2013 special session on
  real-parameter optimization, Computational Intelligence Laboratory, Zhengzhou
  University, Zhengzhou, China and Nanyang Technological University, Singapore,
  Technical Report 201212~(34) (2013) 281--295.

\bibitem{Derrac2011}
J.~Derrac, S.~Garc{\'{i}}a, D.~Molina, F.~Herrera,
  \href{http://dx.doi.org/10.1016/j.swevo.2011.02.002}{{A practical tutorial on
  the use of nonparametric statistical tests as a methodology for comparing
  evolutionary and swarm intelligence algorithms}}, Swarm and Evolutionary
  Computation 1~(1) (2011) 3--18.
\newblock \href {https://doi.org/10.1016/j.swevo.2011.02.002}
  {\path{doi:10.1016/j.swevo.2011.02.002}}.
\newline\urlprefix\url{http://dx.doi.org/10.1016/j.swevo.2011.02.002}

\bibitem{Garcia-Martinez2010}
C.~Garc{\'{i}}a-Mart{\'{i}}nez, M.~Lozano, {Evaluating a local genetic
  algorithm as context-independent local search operator for metaheuristics},
  Soft Computing 14~(10) (2010) 1117--1139.
\newblock \href {https://doi.org/10.1007/s00500-009-0506-1}
  {\path{doi:10.1007/s00500-009-0506-1}}.

\bibitem{Melin2013}
P.~Melin, F.~Olivas, O.~Castillo, F.~Valdez, J.~Soria, M.~Valdez,
  \href{http://dx.doi.org/10.1016/j.eswa.2012.12.033}{{Optimal design of fuzzy
  classification systems using PSO with dynamic parameter adaptation through
  fuzzy logic}}, Expert Systems with Applications 40~(8) (2013) 3196--3206.
\newblock \href {https://doi.org/10.1016/j.eswa.2012.12.033}
  {\path{doi:10.1016/j.eswa.2012.12.033}}.
\newline\urlprefix\url{http://dx.doi.org/10.1016/j.eswa.2012.12.033}

\bibitem{Tanabe2013}
R.~Tanabe, A.~Fukunaga, {Success-history based parameter adaptation for
  Differential Evolution}, 2013 IEEE Congress on Evolutionary Computation, CEC
  2013~(3) (2013) 71--78.
\newblock \href {https://doi.org/10.1109/CEC.2013.6557555}
  {\path{doi:10.1109/CEC.2013.6557555}}.

\bibitem{Peram2003}
T.~Peram, K.~Veeramachaneni, C.~K. Mohan, {Fitness-distance-ratio based
  particle swarm optimization}, 2003 IEEE Swarm Intelligence Symposium, SIS
  2003 - Proceedings~(2) (2003) 174--181.
\newblock \href {https://doi.org/10.1109/SIS.2003.1202264}
  {\path{doi:10.1109/SIS.2003.1202264}}.

\bibitem{Qu2013}
B.~Y. Qu, P.~N. Suganthan, S.~Das, {A distance-based locally informed particle
  swarm model for multimodal optimization}, IEEE Transactions on Evolutionary
  Computation 17~(3) (2013) 387--402.
\newblock \href {https://doi.org/10.1109/TEVC.2012.2203138}
  {\path{doi:10.1109/TEVC.2012.2203138}}.

\bibitem{Engelbrecht2010}
A.~P. Engelbrecht, {Heterogeneous particle swarm optimization}, Lecture Notes
  in Computer Science (including subseries Lecture Notes in Artificial
  Intelligence and Lecture Notes in Bioinformatics) 6234 LNCS (2010) 191--202.
\newblock \href {https://doi.org/10.1007/978-3-642-15461-4\_17}
  {\path{doi:10.1007/978-3-642-15461-4\_17}}.

\bibitem{Liu2021}
W.~Liu, Z.~Wang, Y.~Yuan, N.~Zeng, K.~Hone, X.~Liu, {A Novel
  Sigmoid-Function-Based Adaptive Weighted Particle Swarm Optimizer}, IEEE
  Transactions on Cybernetics 51~(2) (2021) 1085--1093.
\newblock \href {https://doi.org/10.1109/TCYB.2019.2925015}
  {\path{doi:10.1109/TCYB.2019.2925015}}.

\bibitem{Zhang2018}
H.~Zhang, M.~Yuan, Y.~Liang, Q.~Liao,
  \href{https://doi.org/10.1016/j.asoc.2018.04.008}{{A novel particle swarm
  optimization based on prey–predator relationship}}, Applied Soft Computing
  Journal 68 (2018) 202--218.
\newblock \href {https://doi.org/10.1016/j.asoc.2018.04.008}
  {\path{doi:10.1016/j.asoc.2018.04.008}}.
\newline\urlprefix\url{https://doi.org/10.1016/j.asoc.2018.04.008}

\end{thebibliography}


%\begin{thebibliography}{00}


%\bibitem{}

%\end{thebibliography}
\end{document}